\documentclass[10pt,twocolumn,letterpaper]{article}

\PassOptionsToPackage{table}{xcolor}
\usepackage[pagenumbers]{cvpr}      

\usepackage{algorithm}
\usepackage{algpseudocode}
\usepackage{amsthm}
\usepackage{multirow}
\usepackage{tcolorbox}
\usepackage[table]{xcolor}

\definecolor{methodcolor}{HTML}{85B6A5}
\definecolor{tableblue}{HTML}{6D92B8}
\definecolor{tableapricot}{HTML}{DAA07E}
\definecolor{tablegreen}{HTML}{85B6A5}
\definecolor{tablepurple}{HTML}{ABA2C7}
\definecolor{tablemistblue}{HTML}{8FAAC7}
\definecolor{tableperiwinkleblue}{HTML}{899BC2}
\definecolor{tableslate}{HTML}{B9C2CC}
\definecolor{tablerose}{HTML}{CB96A2}
\definecolor{tablecyanblue}{HTML}{5F9698}
\definecolor{tableteal}{HTML}{66A8A3}
\definecolor{prompttitlebg}{HTML}{E1EBE5}
\definecolor{promptframe}{HTML}{A5BBAF}
\definecolor{promptbg}{HTML}{F5F9F6}
\definecolor{prompttitletext}{HTML}{34483D}

\newcommand{\method}{\textcolor{methodcolor}{\textsc{\textbf{Antiphony}}}\xspace}
\newtheorem{remark}{Remark}
\newtheorem{proposition}{Proposition}

\tcbuselibrary{listings}
\newtcblisting{prompt}[2][]{
  listing only,
  colbacktitle=prompttitlebg,
  colframe=promptframe,
  colback=promptbg,
  coltitle=prompttitletext,
  fonttitle=\bfseries\small,
  title={#2},
  boxrule=0.6pt,
  arc=1.5mm,
  top=2mm,
  bottom=2mm,
  left=2mm,
  right=2mm,
  toptitle=0.8mm,
  bottomtitle=0.8mm,
  listing options={
    language={},
    numbers=none,
    basicstyle=\ttfamily\footnotesize,
    breaklines=true,
    breakatwhitespace=true,
    breakindent=0pt,
    breakautoindent=false,
    columns=fullflexible,
    keepspaces=true,
    showstringspaces=false
  },
  #1
}

\definecolor{cvprblue}{rgb}{0.21,0.49,0.74}
\usepackage[pagebackref,breaklinks,colorlinks,allcolors=cvprblue]{hyperref}

\def\confName{CVPR}
\def\confYear{2027}

\title{Antiphony: A Dual-Process Agent for Deepfake Video Detection\\with Real-Video Calibration}

\author{
Xinyu Yan\textsuperscript{1,2}\quad
Yurong Hao\textsuperscript{1}\quad
Zherui Li\textsuperscript{1}\quad
Yebo Wu\textsuperscript{1,3}\quad
Tiantong Wu\textsuperscript{1,2}\quad
Jiaming Zhang\textsuperscript{1}\\
Zihao Zeng\textsuperscript{1,2}\quad
Zheng Lin\textsuperscript{1,4}\quad
Zijie Wang\textsuperscript{1}\quad
Pengpeng Qiao\textsuperscript{5}\quad
Yang Cao\textsuperscript{5}\quad
Wei Yang Bryan Lim\textsuperscript{1}\\[5pt]
\textsuperscript{1}Nanyang Technological University\qquad
\textsuperscript{2}Alibaba-NTU Global e-Sustainability CorpLab (ANGEL)\\
\textsuperscript{3}University of Macau\qquad
\textsuperscript{4}Xidian University\qquad
\textsuperscript{5}Institute of Science Tokyo\\[5pt]
{\ttfamily\small XINYU039@e.ntu.edu.sg}
}

\begin{document}
\maketitle
\begin{abstract}
Reliable video deepfake detection requires generalization to unseen generators while limiting false alarms on authentic footage. Although existing detectors capture diverse visual cues, they may produce overconfident errors under generator shift, limiting the reliability of confidence-based fusion.
Threshold calibration is further complicated by shifts in authentic capture domains and the lack of future-generator examples.
To address these challenges, we present \method, a framework that combines calibrated fusion of four visual experts with selective investigation. Authentic videos serve as a shared reference for score fusion and threshold calibration, enabling threshold updates without target deepfakes.
For this expert-fusion stage, we establish finite-sample false-alarm control when the scoring procedure is fixed independently of calibration and authentic samples are exchangeable.
To refine decisions near the calibrated threshold, a tool-using multimodal investigator examines these videos and contributes an additional fusion score.
We train the investigator on generator-held-out predictions and counterfactual expert outputs, to improve robustness to unreliable expert evidence.
Across six benchmark protocols, \method achieves a mean benchmark score of 88.3\%, exceeding the strongest evaluated MLLM and conventional detector baselines by 14.7 and 11.5 points, respectively.
It also outperforms ARGUS by 20.2 points on average and by 3.6 balanced-accuracy points on FaceVid's 20 unseen generators.
\end{abstract}    
\vspace{-0.5cm}
\section{Introduction}
\label{sec:intro}

\begin{figure}[t]
  \centering
  \includegraphics[width=\linewidth]{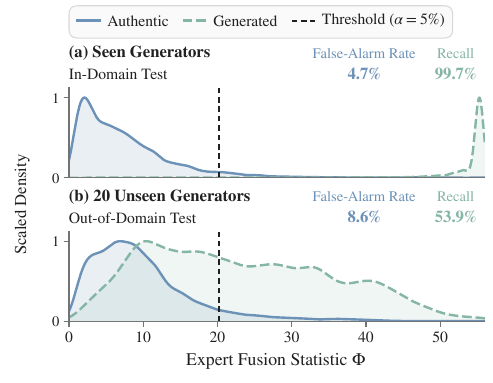}
  \vspace{-0.8cm}
  \caption{\textbf{Authentic scores as a calibration reference.} At a fixed FaceVid threshold, recall drops sharply on unseen generators, while false-alarm rates change less. Curves are peak-normalized.}
  \label{fig:motivation}
  \vspace{-0.6cm}
\end{figure}

Recent video generators such as Seedance~2.5~\cite{seedance25} and Wan~3.0~\cite{wan30} can synthesize and edit videos from multimodal inputs, expanding the ways in which realistic visual content can be created and altered. For applications such as content moderation and media verification, these capabilities heighten the need for reliable video authenticity assessment. Missed deepfakes can allow generated or manipulated content to be accepted as authentic, while false alarms can undermine the credibility of authentic footage. Supporting these applications requires detectors to remain effective as new generators emerge and authentic capture domains change. Crucially, such decisions may be required before representative deepfakes from a new generator are available for training or validation.

This setting calls for a decision rule that retains deepfake recall under generator shift while limiting its false-alarm rate (FAR) to a prescribed level. Specialist detectors provide complementary appearance and temporal cues~\cite{stil,univfd,effort}, making expert fusion a promising way to exploit evidence across different generators. Under generator shift, however, an expert can remain highly confident even when its prediction is wrong. Confidence-based selection or direct score aggregation can consequently favor misleading evidence, even when another expert provides useful cues. Fusion and threshold calibration therefore face a coupled challenge of interpreting heterogeneous expert scores and setting the operating point of the fused statistic when expert reliability on unseen generators is unknown.

Our reproductions on FaceVid-Forensics-100K~\cite{argus} illustrate this difficulty. On the shared evaluation subset of Effort and ARGUS, Effort assigns a deepfake probability below 0.1 to 28.1\% of the deepfakes, indicating high confidence in incorrect authentic predictions. An oracle that chooses correctly whenever either detector is right reaches 82.6\% balanced accuracy (BAcc), whereas the tested simple fusion rules reach at most 68.2\%. Complementary evidence is therefore available, yet confidence-based decisions can leave much of its potential unrealized. A separate FaceVid diagnostic at fixed thresholds shows smaller changes in FAR than in fake recall across the in-domain and unseen-generator splits (Figure~\ref{fig:motivation}). These observations motivate using authentic videos as a common reference for interpreting expert scores and calibrating the fused decision threshold when target deepfakes are unavailable.

Building on this idea, we present \method, a framework that combines calibrated expert fusion with selective investigation. We use authentic-reference calibration to give heterogeneous expert scores a common interpretation, then calibrate the fused statistic on a separate authentic subset to set its operating point. This separates evidence aggregation from the choice of a false-alarm level and allows the score reference and decision threshold to be updated using authentic target videos while keeping the experts fixed. For videos near the calibrated threshold, selective investigation acquires additional visual evidence under a limited computational budget.
A tool-using multimodal large language model (MLLM) performs this investigation using video observations and expert reports. Accurate expert predictions on familiar generators can encourage the investigator to follow these reports without inspecting the video, leaving it vulnerable to expert errors under generator shift. We therefore use generator-held-out predictions to expose it to errors on unfamiliar generators, and introduce counterfactual expert evidence to create additional conflicts between expert reports and unchanged visual observations. At inference, the investigator's score undergoes authentic-reference calibration before being fused with the original expert scores. The augmented statistic receives a separately calibrated threshold, allowing additional evidence to confirm or revise the initial verdict.
Across six benchmark protocols, \method achieves a mean benchmark score of 88.3\%, outperforming the strongest conventional detectors, X-CLIP and DeMamba, by 11.5 points. It further surpasses Qwen3.8-Max, the strongest evaluated MLLM, by 14.7 points and ARGUS by 20.2 points.
Our key contributions are summarized as follows:
\begin{itemize}
\item We propose \method, a deepfake video detection framework that combines complementary visual experts through authentic-reference calibration, allowing its threshold to be updated without target deepfakes.
\item We equip \method with a selective investigator trained on generator-held-out predictions and counterfactual expert evidence. Its score augments the calibrated fusion, linking tool-based inspection to expert evidence.
\item We evaluate \method across six benchmark protocols; it achieves a mean benchmark score of 88.3\%. On ViF, its estimated amortized batch throughput is $93\times$ that of released ARGUS.
\end{itemize}
\section{Related Work}
\label{sec:related}

\paragraph{Visual Deepfake Detectors.}
Facial deepfake detectors exploit image-level artifacts, such as blending boundaries~\cite{facexray} and frequency statistics~\cite{f3net}. Video detectors further capture temporal coherence and spatiotemporal inconsistencies~\cite{ftcn,tfcu,stil}. AltFreezing~\cite{altfreezing} and TALL~\cite{tall} explore strategies for learning spatiotemporal features. For fully generated videos, DeMamba~\cite{demamba} models spatiotemporal inconsistencies, while ReStraV~\cite{restrav} exploits frame-feature trajectory geometry. Image-detector transfer offers another route to video detection~\cite{fakei2v}. Pretrained representations support frozen-feature classification~\cite{univfd} and adaptation through adapters, prompts, or subspace decomposition~\cite{foradapter,sepl,effort}. Unlike these approaches, \method focuses on reliable fusion of heterogeneous visual experts. It calibrates their scores against authentic videos and sets a threshold for the fused statistic, allowing reference updates without retraining encoders or collecting target deepfakes.

\vspace{-0.2cm}
\paragraph{MLLM-Based Detectors.}
MLLM-based detectors leverage multimodal reasoning to make authenticity judgments and provide explanations. Skyra~\cite{skyra} grounds predictions in annotations of visible artifacts, while VideoVeritas~\cite{videoveritas} combines preference alignment with reinforcement learning on grounding and counting tasks. VidGuard-R1~\cite{vidguard} uses rewards for temporal consistency and generation complexity, and BusterX++~\cite{busterx} jointly trains image and video detection through reinforcement learning. These methods rely on MLLMs to make detection decisions. In contrast, \method assigns scoring to visual experts and selectively investigates videos near the fusion threshold, avoiding MLLM-based investigation on every sample.

\vspace{-0.2cm}
\paragraph{Agent-Based Detection.}
Detection agents extend MLLM reasoning through specialist collaboration and tool use. ARGUS~\cite{argus} reconciles reports from multiple forensic experts, while Hermes~\cite{hermes} organizes evidence in a graph for multi-agent deliberation. TUE-Detector~\cite{tue} learns tool-use strategies from teacher-generated trajectories and reinforcement learning. Image-based agents further explore tool use~\cite{fact}, code-based analysis~\cite{codeloop}, and expert reliability for evidence arbitration~\cite{agentfox}. Selective investigation reduces inference costs by routing uncertain predictions to MLLMs and fusing their outputs with visual detector scores~\cite{cocovideo}, or using authentic-calibrated gates for sequential decisions and escalation~\cite{deer}. Agent evaluations also examine evidence arbitration and the reliability of visual explanations~\cite{dissecting}. Unlike these approaches, \method keeps its toolbox fixed and trains the investigator on generator-held-out predictions and counterfactual expert outputs. This exposes the investigator to misleading expert scores alongside unchanged video evidence, enabling an additional authentic-calibrated score to augment expert fusion.
\section{Preliminary}
\label{sec:analysis}

\paragraph{Notation.}
Let $v$ denote a video and $y\in\{0,1\}$ its label, with $0$ for authentic videos and $1$ for deepfakes, including generated and manipulated videos. The training set is $\mathcal{D}_{\mathrm{train}}=\{(v_i,y_i,g_i)\}_{i=1}^{n}$, where $g_i$ identifies the generator or manipulation method for a deepfake, or the capture domain for an authentic video. We denote the authentic calibration set by $\mathcal{C}$ and a test batch by $\mathcal{V}$. The calibration set is excluded from parameter fitting. A binary decision rule $\delta(v)$ flags a deepfake when it returns $1$, and $\alpha\in(0,1)$ is the prescribed false-alarm level. For a finite collection $\mathcal{A}$, $|\mathcal{A}|$ denotes its size, and $\mathbf{1}[\cdot]$ is the indicator function.

\vspace{-0.3cm}
\paragraph{Problem Formulation.}
We study detection under shifts in deepfake generators and authentic capture domains. Let $P_{\mathrm{eval}}$ denote the joint distribution of videos and labels in an evaluation domain. For the initial expert-fusion rule $\delta$, our population design objective is
\begin{equation}
\begin{aligned}
\max_{\delta}\ &\Pr_{P_{\mathrm{eval}}}\big[\delta(v)=1 \mid y=1\big]\\
\text{s.t.}\ &\mathrm{FAR}(\delta)=\Pr_{P_{\mathrm{eval}}}\big[\delta(v)=1\mid y=0\big]\le\alpha,
\end{aligned}
\label{eq:objective}
\end{equation}
where $\alpha$ is fixed before evaluation. This population criterion guides our design, since the distribution of future deepfakes is unknown. For the initial expert-fusion rule, Proposition~\ref{prop:main} establishes marginal false-alarm control under fixed-score and exchangeability assumptions. The final selective cascade is evaluated empirically, as detailed in Section~\ref{sec:system2}. We evaluate unseen-generator and in-distribution splits using BAcc and benchmark-specific metrics. Evaluation settings and metric definitions are detailed in Section~\ref{sec:setup} and Appendix~\ref{sec:supp_protocols}.

\vspace{-0.3cm}
\paragraph{Metadata Shortcuts.}
Container metadata can reveal dataset provenance without visual evidence of synthesis or manipulation. We use index-based frame sampling and exclude container metadata from model inputs. Baselines retain their released pipelines under consistent benchmark splits and evaluation metrics. Audit details, preprocessing and coverage are listed in Appendices~\ref{sec:supp_meta} and~\ref{sec:supp_baselines}.

\vspace{-0.3cm}
\paragraph{Authentic Scores as a Calibration Reference.}
On FaceVid, thresholds calibrated on held-out authentic videos yield FARs of 3.9--5.0\% in-domain and 6.5--10.2\% on the out-of-domain (OOD) split across the four experts of \method. Recall drops from 97.0--99.7\% on seen generators to 34.2--47.7\% on unseen ones. The fused statistic exhibits a similar recall drop, as shown in Figure~\ref{fig:motivation}. These results highlight distinct effects of authentic-domain and generator shifts: authentic-domain changes affect false alarms, while unseen generators substantially degrade recall. Authentic target videos can update the score reference and threshold when the capture domain changes, without requiring target deepfakes. For the comprehensive diagnostic settings, please refer to Appendix~\ref{sec:supp_calib}.
\section{Proposed Method: \method}
\label{sec:method}

\begin{figure*}[t]
  \centering
  \includegraphics[width=\textwidth]{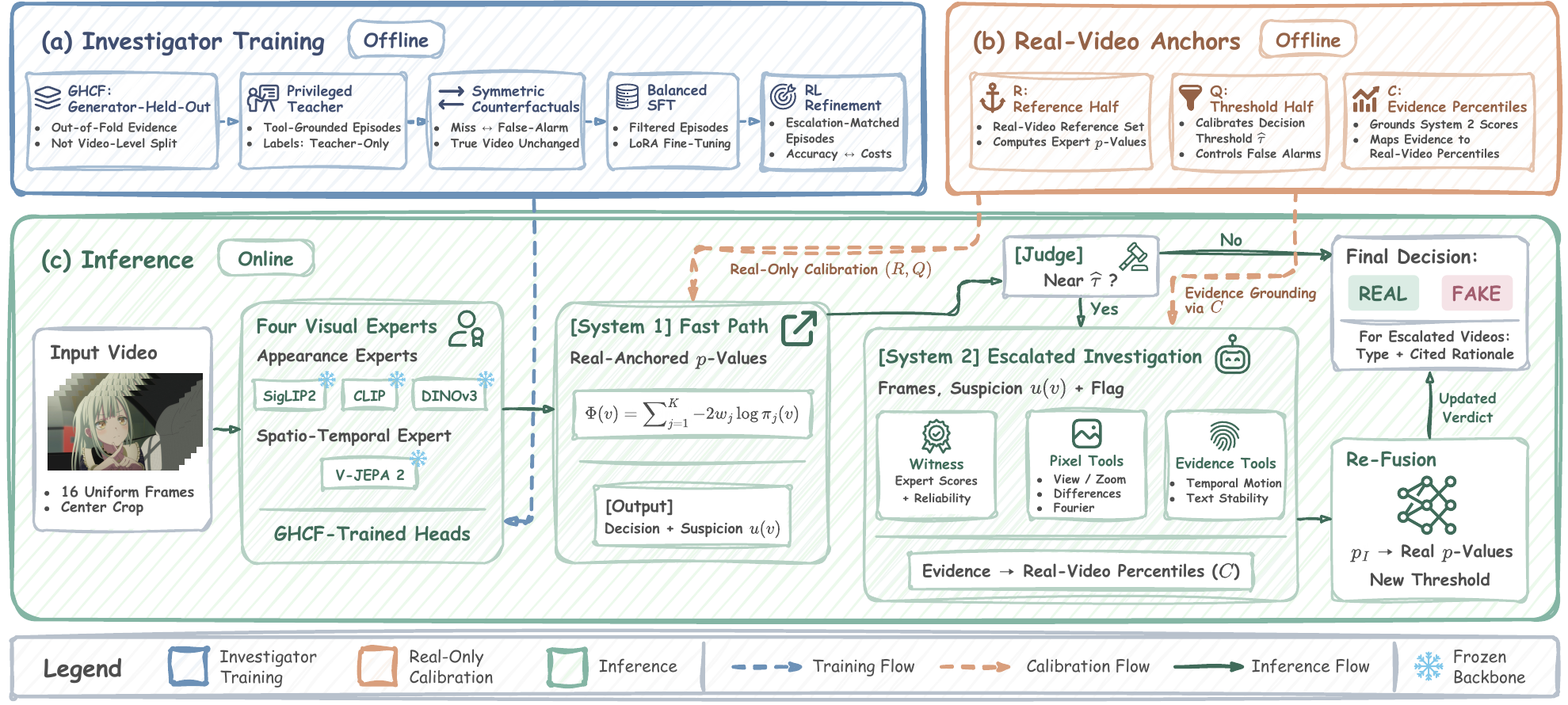}
  \vspace{-0.5cm}
  \caption{\textbf{Overview of our proposed framework:~\method.} (a) The investigator learns from generator-held-out predictions and counterfactual expert evidence. (b) Authentic-video references provide score-specific $p$-values through $\mathcal{R}$, decision thresholds through $\mathcal{Q}$, and tool-evidence percentiles through $C$. (c) Four visual experts support fast decisions through calibrated Fisher fusion. A batch gate selects videos near the threshold for tool-based investigation. The investigator's calibrated score is then fused with the expert scores, and a separately calibrated threshold determines the final decision.}
  \label{fig:framework}
  \vspace{-0.3cm}
\end{figure*}

\method combines calibrated expert fusion with selective investigation. Experts score every video, while a tool-using investigator examines selected videos near the threshold. Both sources of evidence enter a common decision rule through calibration on authentic videos. We train the investigator on generator-held-out predictions and counterfactual expert outputs to expose it to unreliable evidence. Figure~\ref{fig:framework} summarizes the framework.

\subsection{Frozen Experts and Generator-Held-Out Cross-Fitting}
\label{sec:experts}

\paragraph{Frozen Experts.}
We use $K=4$ frozen encoders, SigLIP2~\cite{siglip2}, CLIP~\cite{clip}, DINOv3~\cite{dinov3} and V-JEPA~2~\cite{vjepa2}, with lightweight heads that map their features to video logits. The first three encode individual frames, whose features are aggregated by temporal heads. V-JEPA~2 supplies video features for a temporal scorer. Only the heads are fitted for detection, preserving the pretrained representations. Larger scores indicate stronger evidence of a deepfake. Backbone configurations, feature extraction and input views are detailed in Appendices~\ref{sec:supp_agent} and~\ref{sec:supp_design}.

\vspace{-0.2cm}
\paragraph{Generator-Held-Out Cross-Fitting (GHCF).}
A random split of clips can leave the same generator families in both fitting and held-out data. To expose the investigator to errors on unfamiliar generators, we partition training videos into $F$ disjoint folds, keeping each deepfake generator family in one fold. Authentic videos follow a fixed domain- or path-based assignment. Let $f(i)$ identify the fold assigned to video $v_i$, and let $s_j^{(-f)}$ denote expert $j$ with its head fitted without fold $f$. Within a fixed input-view protocol, training videos receive out-of-fold (OOF) scores, while calibration and test videos receive the fold average,
\begin{equation}
\begin{aligned}
\tilde s_j(v_i)&=s_j^{(-f(i))}(v_i),\\
s_j(v)&=\frac{1}{F}\sum_{f=1}^{F}s_j^{(-f)}(v).
\end{aligned}
\label{eq:crossfit}
\end{equation}
The investigator thus receives predictions from heads fitted without the corresponding deepfake generator family, alongside the original video evidence. Further information on fold assignments appears in Appendix~\ref{sec:supp_protocols}.

\vspace{-0.2cm}
\paragraph{Calibration Set.}
We reserve authentic videos $\mathcal{C}$ and exclude them, together with linked videos, from parameter fitting. Calibration and test clips receive the same fold-averaged scores. Appendices~\ref{sec:supp_protocols} and~\ref{sec:supp_dev} provide further details on data exclusions and selection procedures.

\subsection{Real-Anchored Conformal Fusion}
\label{sec:conformal}

We use established conformal outlier testing and score combination~\cite{bates2023outliers,magesh2023ood,szabo2026fusion} to interpret heterogeneous expert outputs against a common authentic reference. A random index split partitions $\mathcal{C}$ into disjoint reference and threshold subsets $\mathcal{R}$ and $\mathcal{Q}$. The first defines the score transformation; the second sets the operating threshold for the resulting statistic. For expert $j\in\{1,\ldots,K\}$, the reference p-value and weighted Fisher statistic~\cite{fisher1928statistical} are
\begin{align}
\pi_j(v)&=\frac{1+|\{r\in\mathcal{R}:s_j(r)\ge s_j(v)\}|}{|\mathcal{R}|+1},
\label{eq:pvalue}\\
\Phi(v)&=-2\sum_{j=1}^{K}w_j\log\pi_j(v).
\label{eq:fusion}
\end{align}
A small $\pi_j(v)$ indicates that the expert's score exceeds most authentic reference scores. These upper-tail ranks give scores with different numerical scales a common interpretation, without estimating a posterior deepfake probability. The nonnegative weights $w_j$ specify each expert's contribution; we use $w_j=1$. Small p-values contribute more to $\Phi$, allowing an unusual score from one expert to contribute even when other experts give weaker evidence.

The experts can be dependent, and their p-values share $\mathcal{R}$. We calibrate $\Phi$ empirically on $\mathcal{Q}$ rather than use a theoretical chi-squared cutoff. The standard split-conformal quantile rule at level $\alpha$ gives a threshold $\hat\tau$, and a video is flagged when $\Phi(v)>\hat\tau$. Updating $\mathcal{R}$ and $\mathcal{Q}$ with authentic target videos adapts the score reference and threshold while keeping the experts fixed.

Standard split-conformal validity gives $\Pr[\Phi(v)>\hat\tau]\le\alpha$ for an authentic test video $v$ when the expert score functions and weights are fixed independently of calibration and test videos, and authentic calibration and test samples are exchangeable~\cite{vovk2005,bates2023outliers}. This bound averages over calibration and test draws and permits dependence among experts. Main-table configurations reuse calibration during development, so their FAR is evaluated empirically. Appendix~\ref{sec:supp_theory} presents the quantile rule, formal assumptions, and proof.

\subsection{Selective Investigation}
\label{sec:system2}

\paragraph{Escalation and Re-Fusion.}
Within a completed test batch $\mathcal{V}$, the budget $\rho\in[0,1]$ determines the fraction of videos investigated. We rank videos by their distance to the fusion threshold,
\begin{equation}
d(v)=\left|\log\bigl(\Phi(v)+\varepsilon\bigr)-\log\bigl(\hat\tau+\varepsilon\bigr)\right|,
\label{eq:gate_distance}
\end{equation}
where $\varepsilon>0$ ensures defined logarithms, and select the $\operatorname{round}(\rho|\mathcal{V}|)$ smallest distances to form $\mathcal{E}(\mathcal{V})$. Videos on either side of the threshold are eligible, so investigation can address potential misses and false alarms.

The investigator returns a score $p_I(v)\in[0,1]$ estimating the deepfake probability. We compute $\pi_I(v)$ using Eq.~\eqref{eq:pvalue} with $p_I$ in place of $s_j$ and fuse it with the expert statistic,
\begin{equation}
\Phi_2(v)=\Phi(v)-2w_I\log\pi_I(v),\qquad w_I\ge0.
\label{eq:refusion}
\end{equation}
Before test evaluation, the investigator scores all calibration videos. Its outputs on $\mathcal{R}$ supply reference scores, and the augmented statistic on $\mathcal{Q}$ calibrates a separate threshold $\hat\tau_2$ using the same quantile rule. Recalibrating the threshold accounts for the extra score contribution. The batch-dependent decision is
\begin{equation}
\delta_{\mathcal{V}}(v)=
\begin{cases}
\mathbf{1}[\Phi(v)>\hat\tau], & v\notin\mathcal{E}(\mathcal{V}),\\
\mathbf{1}[\Phi_2(v)>\hat\tau_2], & v\in\mathcal{E}(\mathcal{V}).
\end{cases}
\label{eq:cascade_decision}
\end{equation}
Since the augmented score is compared with its own threshold, investigation can confirm or reverse the initial decision, including removing a false alarm. The original expert scores remain part of this decision. The inference algorithm and calibration protocol are listed in Appendix~\ref{sec:supp_agent}.

\vspace{-0.2cm}
\paragraph{Investigator and Tools.}
The investigator receives sampled frames, the expert-fusion decision and a suspicion score $u(v)$ giving a normalized rank of $\Phi(v)$ against authentic calibration scores. A fixed toolbox supplies expert scores and reliability summaries, pixel-level views, and measurements of temporal, motion and text inconsistencies. Expert tools report model judgments, while pixel and evidence tools let the investigator inspect the underlying observations. It alternates tool calls with reasoning before returning $p_I(v)$ and a rationale referring to its observations. Because its prompts and tool statistics depend on calibration, we assess the cascade FAR empirically. We provide tool definitions and further clarify the scope of the guarantee in Appendices~\ref{sec:supp_agent} and~\ref{sec:supp_theory}.

\subsection{Training the Investigator}
\label{sec:training}

\paragraph{Privileged Distillation.}
We initialize the investigator by supervised fine-tuning (SFT) on teacher-generated tool-use demonstrations. The teacher has access to training labels and available artifact annotations or reference explanations. We filter demonstrations for valid tool use and accurate verdicts, and exclude privileged information from the student's inputs. The retained transcripts provide examples of requesting evidence, interpreting tool responses and submitting a verdict. Our selected recipe uses unmodified expert evidence during distillation and introduces counterfactual evidence during reinforcement learning (RL). Distillation details are given in Appendix~\ref{sec:supp_agent}.

\vspace{-0.2cm}
\paragraph{Counterfactual Expert Evidence.}
GHCF supplies predictions on held-out generator families, but experts remain correct on many training videos. We construct misleading expert evidence by replacing selected witness outputs with out-of-fold evidence from a donor video, while preserving the video and its label. An unflagged authentic donor simulates a missed detection for a deepfake, while a flagged deepfake donor simulates a false alarm for an authentic video. We replace the appearance witness, the temporal witness or both, and adjust the displayed suspicion using the donor's suspicion. Pixel and evidence tools continue to operate on the original video. These episodes create conflicts between expert testimony and visual observations while retaining the original label as the training target.

We train on a mixture of original and counterfactual episodes using a policy-gradient objective inspired by group relative policy optimization (GRPO)~\cite{grpo}. Multiple episodes are sampled for each video, and rewards are normalized within that group to form the update advantages. The reward combines the Brier accuracy term $1-(p_I-y)^2$ with penalties for tool calls, execution errors and ending without a pixel or evidence tool call. This objective penalizes confident mistakes and encourages pixel or evidence tool use while accounting for tool cost. Further information on reward definitions, the implemented loss, and model selection is provided in Appendices~\ref{sec:supp_agent} and~\ref{sec:supp_dev}.

\begin{table*}[t]
\centering
\caption{\textbf{Results across six benchmarks.} Train identifies training data (FV: FaceVid, GB: GenBuster, GB+: GB and Additional Sources, GenV: GenVideo, Img: Image Data, Each: Benchmark-Specific Training). Shading marks access to the benchmark's training split. Bold and underline mark the best and second-best completed results with at least 90\% verdict coverage on the relevant splits. Starred results exclude pipeline failures; Table~\ref{tab:supp_strict} counts them as errors. ViF counts failures within matched pairs. \textcolor{methodcolor}{\textsc{\textbf{Antiphony-F}}} uses expert fusion alone. Our results report ten-seed mean $\pm$ std; DVF uses 300 held-out authentic clips for calibration.}
\vspace{-0.2cm}
\label{tab:main}
\footnotesize
\setlength{\tabcolsep}{3.0pt}
\renewcommand{\arraystretch}{1.10}
\setlength{\lightrulewidth}{0.3pt}
\setlength{\arrayrulewidth}{0.3pt}
\resizebox{\textwidth}{!}{%
\begin{tabular}{cllcccccccc|cc}
\toprule
& & & \multicolumn{2}{c}{\textbf{FaceVid OOD}}
& \textbf{ViF}
& \multicolumn{2}{c}{\textbf{GenBuster}}
& \textbf{CoCo}
& \textbf{GenVideo}
& \textbf{DVF}
& \multicolumn{2}{|c}{\textbf{Overall}} \\
\cmidrule(lr){4-5}\cmidrule(lr){6-6}\cmidrule(lr){7-8}\cmidrule(lr){9-9}\cmidrule(lr){10-10}\cmidrule(lr){11-11}\cmidrule(lr){12-13}
\multicolumn{1}{c}{\textbf{Type}}
& \multicolumn{1}{l}{\textbf{Method}}
& \multicolumn{1}{l}{\textbf{Train}}
& \textbf{BAcc} & \textbf{F1} & \textbf{Acc}
& \textbf{Test} & \textbf{Closed} & \textbf{Acc}
& \textbf{BAcc} & \textbf{BAcc}
& \textbf{Avg} & \textbf{Min} \\
\midrule
\multirow{5}{*}{\textcolor{tableapricot!80!black}{\shortstack[c]{\textbf{Image}\\\textbf{Detector}}}} & AIDE~\cite{aide} & Img & 52.71 & 19.19 & 51.11 & 63.55 & 61.30 & 53.54 & 66.43$^*$ & 60.07 & 57.7 & 51.1 \\
 & DRCT~\cite{drct} & Img & 50.81 & 37.16 & 61.38 & 68.60 & 66.55 & 60.67 & 87.71 & \textbf{82.47} & 68.4 & 50.8 \\
 & NPR~\cite{npr} & Img & 49.97 & 0.00 & 50.44 & 73.65 & 70.00 & 50.85 & 66.16 & 54.28 & 57.3 & 50.0 \\
 & SAFE~\cite{safe} & Img & 49.57 & 0.40 & 50.03 & 56.90 & 50.95 & 49.95 & 50.20 & 50.18 & 50.6 & 49.6 \\
 & UnivFD~\cite{univfd} & Img & 44.03 & 2.90 & 50.16 & 61.35 & 49.55 & 50.13 & 55.72 & 55.43 & 51.8 & 44.0 \\
\midrule
\multirow{3}{*}{\textcolor{tablerose!80!black}{\shortstack[c]{\textbf{Face}\\\textbf{Specialist}}}} & DFD-FCG~\cite{dfdfcg} & FV & \cellcolor{tableslate!20}54.46$^*$ & \cellcolor{tableslate!20}17.90$^*$ & 52.15 & 65.05 & 65.15 & 49.87 & 53.69$^*$ & 52.17$^*$ & 54.6 & 49.9 \\
 & Effort~\cite{effort} & FV & \cellcolor{tableslate!20}62.51$^*$ & \cellcolor{tableslate!20}43.17$^*$ & 23.92 & 87.05$^*$ & 86.07$^*$ & 52.51$^*$ & 84.99$^*$ & 88.55$^*$ & 66.5 & 23.9 \\
 & TALL~\cite{tall} & FV & \cellcolor{tableslate!20}57.79$^*$ & \cellcolor{tableslate!20}36.36$^*$ & 25.66 & 83.74$^*$ & 84.09$^*$ & 52.53$^*$ & 68.41$^*$ & 79.33$^*$ & 61.3 & 25.7 \\
\midrule
\multirow{8}{*}{\textcolor{tablepurple!80!black}{\shortstack[c]{\textbf{Video}\\\textbf{Detector}}}} & CoCoDetect~\cite{cocovideo} & CoCo & 52.41 & 39.87 & 50.58 & 45.20 & 53.35 & \cellcolor{tableslate!20}87.21 & 35.36 & 69.94 & 57.5 & 35.4 \\
 & DeMamba~\cite{demamba} & Each & \cellcolor{tableslate!20}60.52 & \cellcolor{tableslate!20}37.49 & \cellcolor{tableslate!20}73.63 & \cellcolor{tableslate!20}\underline{95.55} & \cellcolor{tableslate!20}88.40 & \cellcolor{tableslate!20}94.33 & 72.38 & 68.00 & 76.8 & 60.5 \\
 & MM-Det~\cite{mmdet} & DVF & 50.00$^*$ & 44.60$^*$ & 48.85 & 51.85 & 50.00 & 49.10 & 49.39$^*$ & \cellcolor{tableslate!20}68.64$^*$ & 52.8 & 48.8 \\
 & NPR~\cite{npr} & Each & \cellcolor{tableslate!20}64.08 & \cellcolor{tableslate!20}46.99 & \cellcolor{tableslate!20}58.86 & \cellcolor{tableslate!20}57.55 & \cellcolor{tableslate!20}61.90 & \cellcolor{tableslate!20}56.36 & 62.85 & 42.55 & 57.4 & 42.5 \\
 & NSG-VD~\cite{nsgvd} & GenV & 50.38$^*$ & 5.86$^*$ & 52.56 & 58.70 & 52.95 & 49.22$^*$ & \cellcolor{tableslate!20}\textbf{95.23}$^*$ & 68.40 & 61.9 & 49.2 \\
 & ReStraV~\cite{restrav} & Each & -- & -- & \cellcolor{tableslate!20}52.53 & \cellcolor{tableslate!20}81.90 & \cellcolor{tableslate!20}57.25 & \cellcolor{tableslate!20}65.97 & 62.51 & 37.04 & -- & -- \\
 & STIL~\cite{stil} & Each & \cellcolor{tableslate!20}\underline{72.01} & \cellcolor{tableslate!20}56.95 & \cellcolor{tableslate!20}69.94 & \cellcolor{tableslate!20}82.00 & \cellcolor{tableslate!20}86.90 & \cellcolor{tableslate!20}67.95 & 74.20 & 64.68 & 72.2 & 64.7 \\
 & X-CLIP~\cite{xclip} & Each & \cellcolor{tableslate!20}62.25 & \cellcolor{tableslate!20}41.39 & \cellcolor{tableslate!20}78.73 & \cellcolor{tableslate!20}\textbf{96.95} & \cellcolor{tableslate!20}89.60 & \cellcolor{tableslate!20}94.33 & 67.83 & 64.45 & 76.8 & 62.2 \\
\midrule
\multirow{11}{*}{\textcolor{tableperiwinkleblue!80!black}{\shortstack[c]{\textbf{MLLM}}}}
 & DeepSeek-V4.1-Flash~\cite{xu2026deepseek} & -- & 53.15 & 37.90 & 62.68 & 77.50 & 81.75 & 54.72 & 76.41 & 69.52 & 66.0 & 53.2 \\
 & GPT-6.1 Sol~\cite{gpt61sol} & -- & 54.35 & 34.08 & 56.76 & 86.20 & 91.25 & 53.72 & 84.71 & 70.06 & 68.1 & 53.7 \\
 & InternVL3.5-8B~\cite{internvl35} & -- & 43.67 & 22.57 & 52.21 & 60.25 & 58.20 & 50.21 & 48.11 & 58.08 & 51.9 & 43.7 \\
 & Kimi K3.1 Pro~\cite{kimik31pro} & -- & 54.02 & 22.98 & 55.01 & 76.35 & 78.00 & 51.23 & 79.93 & 60.27 & 62.9 & 51.2 \\
 & Qwen2.5-VL-7B~\cite{qwen25vl} & -- & 40.31 & 9.75 & 49.86 & 51.40 & 51.10 & 50.36 & 44.28 & 48.42 & 47.4 & 40.3 \\
 & Qwen3-VL-8B~\cite{qwen3vl} & -- & 48.86 & 3.71 & 49.99 & 51.25 & 56.00 & 50.44 & 54.97 & 50.61 & 51.4 & 48.9 \\
 & Qwen3-VL-Plus~\cite{qwen3vl} & -- & 45.79 & 24.77 & 55.69 & 82.35 & 82.75 & 49.72 & 78.20 & 68.25 & 63.4 & 45.8 \\
 & Qwen3.8-27B~\cite{qwen3827b} & -- & 59.78 & 38.94 & 61.21 & 79.75 & 81.35 & 55.64 & 82.57 & 71.05 & 68.5 & 55.6 \\
 & Qwen3.8-Flash~\cite{qwen38flash} & -- & 58.46 & 37.83 & 63.30 & 79.35 & 80.15 & 55.85 & 86.38 & 73.00 & 69.5 & 55.8 \\
 & Qwen3.8-Max~\cite{qwen38max} & -- & 59.43 & 37.76 & 62.17 & 93.20 & 93.05 & 56.59 & 92.98 & 77.24 & 73.6 & 56.6 \\
 & Qwen3.8-Omni-Flash~\cite{qwen38omniflash} & -- & 61.06 & 42.63 & 69.07 & 73.45 & 74.80 & 58.46 & 88.03 & 75.28 & 71.0 & 58.5 \\
\midrule
\multirow{4}{*}{\textcolor{tableblue!80!black}{\shortstack[c]{\textbf{Specialized}\\\textbf{MLLM}}}} & BusterX++~\cite{busterx} & GB & 47.06$^*$ & 25.86$^*$ & 53.99 & \cellcolor{tableslate!20}90.05 & \cellcolor{tableslate!20}93.30 & 46.92 & 81.08$^*$ & 62.15$^*$ & 63.8 & 46.9 \\
 & Skyra~\cite{skyra} & ViF & 61.58 & 40.12 & \cellcolor{tableslate!20}\textbf{91.43} & 50.40 & 50.45 & 50.69 & 65.78 & 55.33 & 62.5 & 50.4 \\
 & TUE-Detector~\cite{tue} & ViF & 51.85 & 27.70 & \cellcolor{tableslate!20}56.89 & 34.45 & 34.45 & 36.79 & 69.26 & 59.93 & 51.5 & 34.5 \\
 & VideoVeritas~\cite{videoveritas} & GB+ & 57.09$^*$ & 44.58$^*$ & 69.88 & \cellcolor{tableslate!20}92.75 & \cellcolor{tableslate!20}92.60 & 57.38$^*$ & 81.79$^*$ & 80.60$^*$ & 73.2 & 57.1 \\
\midrule
\multirow{1}{*}{\textcolor{tablecyanblue!80!black}{\shortstack[c]{\textbf{Agent}}}} & ARGUS~\cite{argus} & FV & \cellcolor{tableslate!20}68.55 & \cellcolor{tableslate!20}52.31 & 53.90 & 84.60 & 89.95 & 48.69 & 82.14 & 67.74 & 68.1 & 48.7 \\
\midrule
\multirow{2}{*}{\textcolor{tablegreen!80!black}{\shortstack[c]{\textbf{Ours}}}} & \method & Each & \cellcolor{tableslate!20}\textbf{72.16}{\fontsize{6}{6.5}\selectfont$\pm$0.64} & \cellcolor{tableslate!20}\textbf{59.47}{\fontsize{6}{6.5}\selectfont$\pm$1.03} & \cellcolor{tableslate!20}\underline{90.91}{\fontsize{6}{6.5}\selectfont$\pm$0.41} & \cellcolor{tableslate!20}95.28{\fontsize{6}{6.5}\selectfont$\pm$0.29} & \cellcolor{tableslate!20}\textbf{95.42}{\fontsize{6}{6.5}\selectfont$\pm$0.31} & \cellcolor{tableslate!20}\underline{97.39}{\fontsize{6}{6.5}\selectfont$\pm$0.12} & 92.33{\fontsize{6}{6.5}\selectfont$\pm$0.68} & 81.64{\fontsize{6}{6.5}\selectfont$\pm$1.45} & \underline{88.3} & \textbf{72.2} \\
 & \textcolor{methodcolor}{\textsc{\textbf{Antiphony-F}}} & Each & \cellcolor{tableslate!20}71.77{\fontsize{6}{6.5}\selectfont$\pm$0.22} & \cellcolor{tableslate!20}\underline{58.85}{\fontsize{6}{6.5}\selectfont$\pm$0.35} & \cellcolor{tableslate!20}90.62{\fontsize{6}{6.5}\selectfont$\pm$0.13} & \cellcolor{tableslate!20}95.19{\fontsize{6}{6.5}\selectfont$\pm$0.30} & \cellcolor{tableslate!20}\underline{95.37}{\fontsize{6}{6.5}\selectfont$\pm$0.28} & \cellcolor{tableslate!20}\textbf{97.48}{\fontsize{6}{6.5}\selectfont$\pm$0.06} & \underline{93.09}{\fontsize{6}{6.5}\selectfont$\pm$0.25} & \underline{82.31}{\fontsize{6}{6.5}\selectfont$\pm$1.28} & \textbf{88.4} & \underline{71.8} \\
\bottomrule
\end{tabular}}
\vspace{-0.3cm}
\end{table*}

\section{Experiments}
\label{sec:experiments}

\subsection{Experimental Setup}
\label{sec:setup}

\paragraph{Benchmarks.}
Expert heads are trained separately on FaceVid~\cite{argus}, ViF-CoT-4K~\cite{skyra}, GenBuster~\cite{busterxv1} and CoCoVideo~\cite{cocovideo}. GenVideo~\cite{demamba} and DVF~\cite{mmdet} reuse the ViF-trained heads and investigator.
Source calibration uses held-out authentic training videos; main GenVideo results retain ViF calibration, while DVF uses 300 authentic target videos excluded from testing.
Evaluation covers FaceVid OOD with 20 unseen generators, ViF-Bench, GenBuster Test/Closed, CoCoVideo's test split, GenVideo and DVF; DF26~\cite{df26} serves as a supplementary stress test. Appendix~\ref{sec:supp_protocols} specifies data splits and calibration exclusions.

\vspace{-0.3cm}
\paragraph{Metrics.}
Detection performance is measured using BAcc on FaceVid OOD, GenVideo and DVF, paired accuracy averaged over generators on ViF, accuracy on GenBuster, and mean per-generator accuracy on CoCoVideo. We also report F1 on FaceVid to summarize precision and recall, FAR for false alarms, and amortized inference time per video for efficiency. Overall Avg and Min denote the unweighted mean and minimum of six benchmark scores after averaging GenBuster's Test and Closed scores. Appendices~\ref{sec:supp_protocols} and~\ref{sec:supp_cost} describe metric computation and timing protocols.

\vspace{-0.3cm}
\paragraph{Baselines.}
Our evaluation covers 32 baselines: image detectors applied per frame, face specialists, video detectors, general and specialized MLLMs, and detection agents. Representative methods include DRCT~\cite{drct}, Effort~\cite{effort}, STIL~\cite{stil}, Skyra~\cite{skyra}, and ARGUS~\cite{argus}. We use common evaluation splits and metrics, with each baseline's released preprocessing pipeline. See Appendix~\ref{sec:supp_baselines} for reproduction settings, verdict coverage and failure handling.

\begin{figure*}[t]
  \centering
  \includegraphics[width=\textwidth]{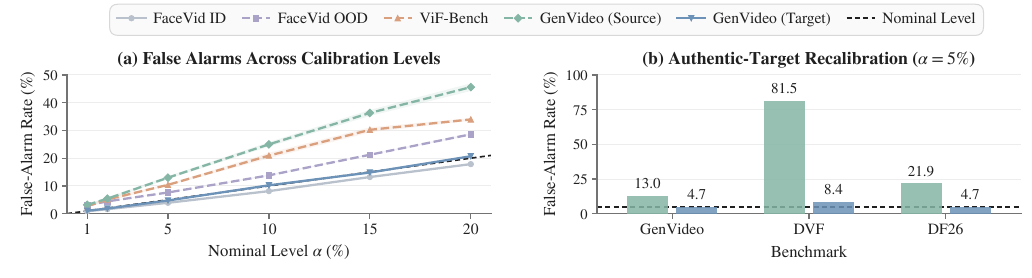}
  \vspace{-0.8cm}
  \caption{\textbf{Authentic target calibration reduces false alarms under domain shift.} Fusion-only FAR is shown (a)~across nominal levels $\alpha$ and (b)~with source (green) versus target (blue) calibration at $\alpha=5\%$. Bands show standard deviation across ten training seeds; dashed black lines mark the nominal FAR. DF26 uses two-fold splits of authentic videos for calibration and evaluation.}
  \label{fig:far}
  \vspace{-0.3cm}
\end{figure*}

\vspace{-0.3cm}
\paragraph{Parameter Settings.}
Expert heads are trained with four GHCF folds while all visual encoders remain frozen. The investigator is Qwen2.5-VL-7B~\cite{qwen25vl}. Expert fusion uses two 16-frame views; investigation uses one uniform 16-frame view. The main settings are $\alpha=5\%$, $\rho=0.3$ and $w_I=1$. Main-table results for \method report the mean and sample standard deviation across ten training seeds, after averaging five calibration splits within each seed. Appendices~\ref{sec:supp_agent} and~\ref{sec:supp_statistics} describe implementation and statistical procedures, including confidence interval (CI) estimation.

\vspace{-0.1cm}
\subsection{Main Results}
\label{sec:results}

\vspace{-0.1cm}
Across six benchmarks in Table~\ref{tab:main}, \method achieves a mean score of 88.3\%, exceeding the strongest evaluated MLLM and conventional detector baselines by 14.7 and 11.5 points, respectively, and ARGUS by 20.2 points. Its minimum benchmark score reaches 72.2, compared with 64.7 for STIL, the baseline with the highest minimum score.

\vspace{-0.3cm}
\paragraph{Unseen Generators and False Alarms.}
On FaceVid's 20 unseen generators, \method reaches 72.16\% BAcc, exceeding ARGUS by 3.61 points (95\% CI $[2.39,4.86]$). It achieves fewer false alarms than STIL at comparable BAcc, with FARs of 7.6\% and 22.7\%, respectively. When STIL is diagnostically thresholded using OOD test reals to match the 7.6\% FAR, its recall is 34.4\%, versus 52.0\% for \method. Matching FAR reveals a recall advantage that is not apparent from BAcc alone, as detailed in Appendix~\ref{sec:supp_baselines}.

\vspace{-0.2cm}
\paragraph{From Open to Commercial Generators.}
With heads trained on four open generators, \method reaches 95.28\% accuracy on GenBuster Test and 95.42\% on Closed, which contains eight commercial generators absent from expert-head training. X-CLIP, which leads Test at 96.95\%, falls to 89.60\% on Closed. The fusion-only variant also maintains similar accuracy across the two splits, supporting transfer beyond the experts' training generator families under GenBuster's shared data protocol. Appendix~\ref{sec:supp_protocols} describes the training data and the two evaluation splits.

\vspace{-0.3cm}
\paragraph{Adaptation with Reusable Components.}
With benchmark-specific heads on the same frozen encoders, \method reaches 97.39\% mean accuracy on CoCoVideo, exceeding the completed baselines in Table~\ref{tab:main}. Both the GenBuster and CoCoVideo configurations reuse the ViF-trained investigator, extending the framework through new expert heads without additional investigator training. With the ViF heads also fixed, the configuration reaches 92.33\% BAcc on GenVideo without target calibration and 81.64\% on DVF with 300 held-out authentic target videos. The next section examines the effect of changing the authentic reference for these fixed experts. Component reuse and additional transfer evaluations are detailed in Appendices~\ref{sec:supp_protocols} and~\ref{sec:supp_zeroshot}.

\subsection{Calibration Analysis}
\label{sec:far}

\paragraph{Adaptation with Authentic Target Videos.}
Updating the authentic reference improves the operating point while keeping the visual experts fixed. On GenVideo, recalibration with 2{,}000 held-out authentic target videos raises the framework's BAcc by 4.59 points to 96.92\%. At a nominal 5\%, target calibration reduces fusion-only FAR from 13.0\% to 4.7\% on GenVideo and from 81.5\% to 8.4\% on DVF (Figure~\ref{fig:far}). These results demonstrate a practical use of authentic target footage for adapting detection without collecting new deepfakes or retraining the experts.

A separate check fixes the scoring procedure before calibration and evaluation on previously unused authentic videos. The observed FAR is 5.57\% at a nominal 5\%, with a file-resampling 95\% interval of $[4.75,6.38]\%$. This evaluates the calibration procedure on fresh files independently of development reuse. We describe the sampling procedures and report variation across calibration draws in Appendix~\ref{sec:supp_theory}.

\vspace{-0.3cm}
\paragraph{Matched Calibration Access.}
We compare Fisher fusion with mean probability and source-trained logistic stacking using the same authentic target videos. In a separate fusion-only study with a total budget of 300 DVF clips for each rule, Fisher improves BAcc over these alternatives by 5.79 and 2.77 points, respectively. GenVideo shows the same ordering. The advantage persists when calibration access is matched, supporting the choice of score combination. Complete results, including standalone detectors and matched threshold subsets, are reported in Appendix~\ref{sec:supp_equal_calibration}.

\vspace{-0.3cm}
\paragraph{Adaptation to Video Processing.}
In the ViF post-processing study, recalibrating the fusion-only variant on authentic videos processed in the same way as the test clips raises paired accuracy under H.264 CRF~35 compression from 64.2\% to 79.4\%. Recalibration also recovers accuracy under blur and downscaling. This supports using calibration videos that reflect deployment conditions. We report the processing study, including FaceVid, in Appendix~\ref{sec:supp_robust}.

\subsection{Ablation Studies}
\label{sec:ablations}

\begin{table}[t]
\centering
\caption{\textbf{Key ablations (\%).}
(a)~Ten-seed fusion-only means; parentheses give FAR and Avg/Min summarize six benchmarks.
(b)~Single-seed investigator development results using uniform-view expert scores.
Validation uses family-balanced BAcc; testing uses FaceVid OOD BAcc and ViF paired accuracy.
CF denotes counterfactual evidence; -- denotes no RL.
Shading marks the configurations used in \method.}
\vspace{-0.2cm}
\label{tab:ablation_main}
\footnotesize
\renewcommand{\arraystretch}{1.18}
\setlength{\tabcolsep}{2.5pt}
\begin{tabular}{>{\raggedright\arraybackslash}p{\dimexpr0.27\linewidth-2\tabcolsep\relax}*{4}{>{\centering\arraybackslash}p{\dimexpr0.1825\linewidth-2\tabcolsep\relax}}}
\toprule
\multicolumn{5}{l}{\textbf{(a) Fusion and Experts}} \\
\midrule
Variant & \shortstack{FaceVid\\BAcc (FAR)} & \shortstack{ViF\\Acc (FAR)} & Avg & Min \\
\midrule
\rowcolor{methodcolor!15}
Fisher (Ours) & 71.77{\scriptsize\,(7.6)} & \textbf{90.62}{\scriptsize\,(10.4)} & \textbf{88.4} & 71.8 \\
Stouffer & 71.29{\scriptsize\,(7.1)} & 90.30{\scriptsize\,(10.7)} & 87.8 & 71.3 \\
Mean $p$-value & 69.33{\scriptsize\,(6.4)} & 88.68{\scriptsize\,(12.3)} & 85.8 & 69.3 \\
Minimum $p$-value & \textbf{71.89}{\scriptsize\,(9.4)} & 88.82{\scriptsize\,(10.1)} & 87.8 & \textbf{71.9} \\
DINOv3 Only & 67.88{\scriptsize\,(6.9)} & 85.83{\scriptsize\,(9.2)} & \textbf{88.4} & 67.9 \\
\midrule
\multicolumn{5}{l}{\textbf{(b) Investigator Training}} \\
\midrule
SFT / RL & \shortstack{FaceVid\\Val.} & \shortstack{ViF\\Val.} & \shortstack{FaceVid\\Test} & \shortstack{ViF\\Test} \\
\midrule
Clean / -- & 87.96 & 94.92 & 71.68 & 90.36 \\
CF / -- & 88.34 & 94.52 & 71.78 & \textbf{90.81} \\
Clean / Clean & 88.57 & 94.79 & 71.90 & 90.36 \\
CF / CF & 88.23 & 95.06 & 71.51 & 90.12 \\
\rowcolor{methodcolor!15}
Clean / CF & \textbf{89.88} & \textbf{95.26} & \textbf{72.22} & 90.73 \\
\bottomrule
\end{tabular}
\vspace{-0.4cm}
\end{table}

\paragraph{Fusion Rules and Expert Complementarity.}
Table~\ref{tab:ablation_main}(a) separates the effects of score combination and expert choice. With the same four experts and calibration data, Fisher has the highest six-benchmark mean among the tested p-value combination rules. It matches DINOv3's mean of 88.4\% while improving FaceVid OOD BAcc by 3.89 points and ViF paired accuracy by 4.79. The per-benchmark results show how complementary experts strengthen performance where a single representation is less effective. Appendix~\ref{sec:supp_design} presents the full expert-removal and encoder-adaptation comparisons.

\vspace{-0.2cm}
\paragraph{Investigator Training.}
The best validation results in Table~\ref{tab:ablation_main}(b) come from clean distillation followed by counterfactual RL. Compared with RL without counterfactuals, this recipe raises family-balanced validation BAcc by 0.47 points on ViF and 1.31 on FaceVid, with higher test scores on both benchmarks in the same development configuration. Using counterfactuals in both stages lowers validation scores. This development comparison supports the selected sequence of clean distillation and counterfactual RL. Complete recipe comparisons are provided in Appendix~\ref{sec:supp_dev}.

\vspace{-0.2cm}
\paragraph{Contribution of Selective Investigation.}
In the main ten-seed evaluation, adding investigation at $\rho=0.3$ improves FaceVid OOD BAcc by $0.39\pm0.55$ points and ViF paired accuracy by $0.29\pm0.39$. This paired comparison holds the expert heads and calibration pools fixed, measuring the contribution of the additional investigator score. Appendix~\ref{sec:supp_inv} reports tool-group ablations and transfer analyses. We present recorded investigation examples and explanation assessments in Appendices~\ref{sec:supp_cost} and~\ref{sec:supp_judge}.

\subsection{Sensitivity and Efficiency Analysis}
\label{sec:sensitivity}

\begin{figure}[t]
  \centering
  \includegraphics[width=\linewidth]{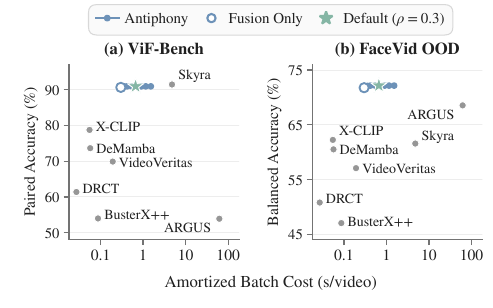}
  \vspace{-0.9cm}
  \caption{\textbf{Detection performance versus amortized batch cost.} Both panels use ViF timing estimates on one A100 with method-specific serving configurations; the cost axis is logarithmic. The \method curves show ten-seed mean scores as $\rho$ varies from 0 to 1 at $w_I=1$. We describe the timing protocol in Appendix~\ref{sec:supp_cost}.}
  \label{fig:cost}
  \vspace{-0.3cm}
\end{figure}

\paragraph{False-Alarm Level and Calibration Size.}
The nominal level $\alpha$ controls the tradeoff between detection and false alarms by adjusting the decision threshold. Calibration-size ablations find that 200 authentic clips keep benchmark scores within half a point of the full-pool results on FaceVid, ViF and GenVideo, although FAR estimates remain sensitive to calibration sampling. Parameter sweeps and the corresponding FARs are provided in Appendix~\ref{sec:supp_ablation}.

\vspace{-0.2cm}
\paragraph{Investigation Budget and Fusion Weight.}
A small budget retains nearly all the investigation gains observed on FaceVid and ViF. With $w_I=1$, reducing $\rho$ from 0.3 to 0.1 changes mean performance by less than 0.02 points on these benchmarks while lowering estimated amortized cost from 0.66 to 0.42\,s per video. The stratum analysis helps explain this pattern: investigation has its largest effect on videos nearest the threshold. The budget $\rho$ sets the fraction of videos investigated, while $w_I$ weights the investigator's contribution to fusion. Appendix~\ref{sec:supp_inv} presents the joint parameter sweep and stratum analysis.

\vspace{-0.3cm}
\paragraph{Amortized Inference Cost.}
At $\rho=0.3$, the estimated amortized cost is 0.66\,s per video for \method and 61.9\,s for released ARGUS. Under the measured one-A100 serving configurations, this corresponds to roughly $93\times$ higher batch throughput. Figure~\ref{fig:cost} shows this performance--cost tradeoff. The measurements include concurrent investigation and describe amortized batch cost rather than individual response latency. Appendix~\ref{sec:supp_cost} describes the timing protocol; Appendix~\ref{sec:supp_inv} compares alternative routing rules.
\section{Conclusion}
\label{sec:conclusion}

We presented \method, a deepfake video detection framework that combines complementary experts and reserves inspection for selected videos.
Calibrating expert and investigator scores against authentic videos lets both sources of evidence contribute to the final decision.
Across six benchmarks, the framework achieves a mean score of 88.3\% and a minimum score of 72.2\%.
Expert comparisons show that fusion can strengthen performance where a single representation is less effective.
Our calibration studies demonstrate how authentic target footage can reduce false alarms with the visual experts fixed.
On FaceVid and ViF, most investigation gains are retained with a small budget, suggesting that choosing which videos to inspect is important for efficient detection.
Future work could select investigations by their expected benefit relative to cost.
\section*{LLM Usage Statement}
Claude and OpenAI Codex helped draft and edit text, refine figures and the presentation of research claims, and generate and debug code.
ChatGPT Deep Research supported literature search and related-work synthesis.
Codex also supported experimental planning and analysis, executed and monitored experiments, and provided reviewer-style feedback and consistency checks.
As part of the research method, Qwen3.8-Max generated investigator training trajectories and revised rationales, while Qwen3.8-Max and Gemini-3.8-Flash assessed explanation grounding.
The research procedures and evaluated MLLMs are described in Sections~\ref{sec:method} and~\ref{sec:experiments} and Appendices~\ref{sec:supp_agent} and~\ref{sec:supp_judge}.
The authors directed the work, reviewed and edited tool outputs, and take full responsibility for all submitted content.
{
    \small
    \bibliographystyle{ieeenat_fullname}
    \bibliography{main}
}
\clearpage
\appendix
\renewcommand{\thepage}{S\arabic{page}}
\setcounter{page}{1}
\maketitlesupplementary
\raggedbottom
\renewcommand{\topfraction}{0.9}
\renewcommand{\dbltopfraction}{0.9}
\renewcommand{\textfraction}{0.05}

\section*{Contents}
\begin{itemize}[itemsep=3pt, parsep=1.5pt]
\item \textbf{Appendix~\ref{sec:supp_protocols}: Datasets and Evaluation Protocols.}
  \begin{itemize}
  \item \ref{sec:supp_datasets}: Datasets.
  \item \ref{sec:supp_training_sources}: Training and Calibration Sources.
  \item \ref{sec:supp_evaluation_protocol}: Evaluation Splits and Metrics.
  \item \ref{sec:supp_statistics}: Result Aggregation and Uncertainty.
  \item \ref{sec:supp_meta}: Metadata Shortcuts and Input Preprocessing.
  \end{itemize}
\item \textbf{Appendix~\ref{sec:supp_agent}: Model Configurations and Training Details.}
  \begin{itemize}
  \item \ref{sec:supp_heads}: Expert Heads.
  \item \ref{sec:supp_tools}: Inference and Tool Use.
  \item \ref{sec:supp_training}: Investigator Training.
  \end{itemize}
\item \textbf{Appendix~\ref{sec:supp_theory}: False-Alarm Analysis.}
  \begin{itemize}
  \item \ref{sec:supp_theory_setting}: Calibration Setting and Assumptions.
  \item \ref{sec:supp_marginal_control}: Marginal False-Alarm Control.
  \item \ref{sec:supp_calibration_uncertainty}: Uncertainty for a Single Calibration Set.
  \item \ref{sec:supp_calibration_dependencies}: Calibration Dependencies in the Implemented System.
  \item \ref{sec:supp_fresh_calibration}: Calibration with Previously Unused Videos.
  \end{itemize}
\item \textbf{Appendix~\ref{sec:supp_baselines}: Baseline Implementation and Comparison Protocols.}
  \begin{itemize}
  \item \ref{sec:supp_baseline_protocol}: Evaluation Settings and Failure Handling.
  \item \ref{sec:supp_baseline_implementations}: Model Implementations and API Settings.
  \item \ref{sec:supp_detection_behavior}: Recall, False Alarms, and Coverage.
  \end{itemize}
\item \textbf{Appendix~\ref{sec:supp_design}: Expert Design and Calibration Analysis.}
  \begin{itemize}
  \item \ref{sec:supp_representations}: Representation Adaptation and Input Views.
  \item \ref{sec:supp_fusion_ablation}: Score Fusion and Expert Complementarity.
  \item \ref{sec:supp_calib}: Calibration Design and Sensitivity.
  \item \ref{sec:supp_equal_calibration}: Matched Calibration Access.
  \end{itemize}
\item \textbf{Appendix~\ref{sec:supp_investigation_analysis}: Selective Investigation Analysis.}
  \begin{itemize}
  \item \ref{sec:supp_dev}: Investigator Training and Selection.
  \item \ref{sec:supp_inv}: Routing, Tools, and Investigation Budget.
  \item \ref{sec:supp_cost}: Inference Cost and Investigation Examples.
  \item \ref{sec:supp_judge}: Explanation Grounding.
  \end{itemize}
\item \textbf{Appendix~\ref{sec:supp_additional}: Generalization across Generators and Video Conditions.}
  \begin{itemize}
  \item \ref{sec:supp_pergen}: Variation across Generators.
  \item \ref{sec:supp_df26}: Recent Talking-Head Generators.
  \item \ref{sec:supp_zeroshot}: Cross-Benchmark Transfer with Fixed Experts.
  \item \ref{sec:supp_robust}: Robustness to Video Processing.
  \end{itemize}
\item \textbf{Appendix~\ref{sec:supp_llm_usage}: Ethics and Research Transparency.}
  \begin{itemize}
  \item \ref{sec:supp_ethics_considerations}: Ethical Considerations.
  \item \ref{sec:supp_llm_disclosure}: LLM Usage Statement.
  \end{itemize}
\end{itemize}

\newpage
\section{Datasets and Evaluation Protocols}
\label{sec:supp_protocols}

Our main evaluation covers six benchmarks for facial manipulation and fully generated videos, with DF26 providing an additional benchmark for public-speaking videos.

\subsection{Datasets}
\label{sec:supp_datasets}
\paragraph{FaceVid-Forensics-100K.}
FaceVid~\cite{argus} contains face-swapped, reenacted, and fully generated videos, allowing facial-manipulation and generated-video detectors to be compared under a common protocol. Its official training, in-domain test, and out-of-domain (OOD) splits contain 68{,}906, 23{,}458, and 7{,}636 clips, respectively. The OOD split includes 20 generators absent from training. Authentic training videos come from AIGVDBench's OpenVid-HD collection, FaceForensics++, and GenVidBench. Authentic OOD videos come from GenVidBench and ViF-Bench, so these benchmarks share some source collections.

\begin{table*}[t]
\centering
\caption{\textbf{Training and calibration sources.} Training counts are shown before removing calibration videos and their linked clips. These excluded videos are not used to fit any expert head. Calibration counts include both the reference and threshold subsets. The last column gives the dataset used to train the investigator.}
\label{tab:protocol_map}
\resizebox{\textwidth}{!}{%
\begin{tabular}{llcccc}
\toprule
\textbf{Benchmark}
& \textbf{Expert-Head Training Pool}
& \textbf{Clips (Real / Fake)}
& \textbf{Excluded Entries}
& \textbf{Authentic Calibration Pool}
& \textbf{Investigator} \\
\midrule
FaceVid OOD & FaceVid Train & 11{,}626 / 57{,}280
& 2{,}711 & 2{,}001 FaceVid Training Clips & FaceVid \\
ViF & ViF-CoT-Derived & 3{,}502 / 3{,}502
& 1{,}204 & 602 ViF Training Clips & ViF \\
GenBuster & Balanced GenBuster Subset & 10{,}000 / 10{,}000
& 600 & 600 GenBuster Training Clips & ViF \\
CoCoVideo & Official Train & 9{,}100 / 9{,}100
& 1{,}200 & 600 CoCoVideo Training Clips & ViF \\
\midrule
GenVideo & Reuse ViF Heads & --
& -- & The Same 602 ViF Clips & ViF \\
DVF & Reuse ViF Heads & --
& -- & 300 Held-Out DVF Clips & ViF \\
\bottomrule
\end{tabular}}
\end{table*}

\paragraph{ViF-CoT-4K.}
ViF-CoT-4K~\cite{skyra} provides annotated generated videos paired with authentic clips from Kinetics-400 and Panda-70M. It covers seven generator configurations and includes explanations of visible artifacts together with their temporal intervals and spatial regions. These annotations support the study of both authenticity judgments and the visual evidence used to explain them.

\paragraph{ViF-Bench.}
ViF-Bench~\cite{skyra} covers 19 generator configurations, including open and commercial text-to-video and image-to-video systems. The benchmark links generated videos to authentic references and evaluates detection across these generation settings. Its paired evaluation protocol balances authentic and generated examples within each generator before averaging across generators.

\paragraph{GenBuster.}
GenBuster~\cite{busterxv1} contains a training set of 200{,}000 videos, equally divided between authentic videos and videos generated by four open models: CogVideoX, EasyAnimate, HunyuanVideo and LTX-Video. The Test split uses the same four generators, while Closed covers eight commercial generators absent from the training set, including Sora, Gen3 and Kling. Each evaluation split contains 1{,}000 authentic and 1{,}000 generated videos, supporting comparisons within and across generation systems.

\paragraph{CoCoVideo-26K.}
CoCoVideo~\cite{cocovideo} contains 13{,}000 authentic--generated video pairs across 13 commercial generators. Authentic videos come from OpenVid-1M, and their first frames and text descriptions condition the paired generations. Its official training, validation, and test splits contain 18{,}200, 3{,}900, and 3{,}900 clips, respectively, with balanced authentic and generated examples within each generator group.

\paragraph{GenVideo.}
GenVideo~\cite{demamba} is a million-scale benchmark for AI-generated video detection. Its authentic training videos come from Kinetics-400 and Youku-mPLUG. The original paper reports an evaluation set containing 10{,}000 authentic videos from MSR-VTT and 8{,}588 generated videos across ten source groups, covering diverse content and generation methods.

\paragraph{DVF.}
Diffusion Video Forensics (DVF)~\cite{mmdet} contains 2{,}750 authentic videos and 3{,}938 generated videos from eight generation sources. Authentic videos come from YouTube-8M and InternVid-10M, while the generated subset includes Stable Video Diffusion outputs. The collection supports evaluating detectors across diffusion-based video generation methods.

\paragraph{DF26.}
DF26~\cite{df26} contains 271 authentic and 2{,}420 generated public-speaking videos across three scenarios: direct-to-camera address, studio interviews, and official statements. Its seven models form ten generator configurations: four commercial generators and three open models, with the latter evaluated in both image-to-video and text-to-video modes. The open image-to-video configurations use the first frame of an authentic clip, allowing detection to be studied when generation is conditioned on real visual content.

\begin{table*}[t]
\centering
\caption{\textbf{Evaluation settings for the six main benchmarks.} Counts describe the evaluation sets before ViF pair matching. Baseline pipelines may fail on some videos; Appendix~\ref{sec:supp_baselines} describes their coverage. ``Benchmark training fakes'' refers to deepfakes from the training split, not test videos or necessarily the same generators.}
\label{tab:protocol_evaluation}
\resizebox{\textwidth}{!}{%
\begin{tabular}{lcll}
\toprule
\textbf{Benchmark}
& \textbf{Clips (Real / Fake)}
& \textbf{Training and Test Setting}
& \textbf{Main Metric} \\
\midrule
FaceVid OOD
& 5{,}716 / 1{,}920
& Benchmark Training Fakes; 20 Unseen Test Generators
& BAcc; Fake-Class F1 \\
ViF
& 165 / 2{,}995
& ViF-CoT Training Fakes; ViF-Bench Test
& Paired Accuracy \\
GenBuster
& 1{,}000 / 1{,}000 in Each Split
& Open-Generator Training Fakes; Test and Closed Evaluated Separately
& Accuracy in Each Split \\
CoCoVideo
& 1{,}950 / 1{,}950
& Benchmark Training Fakes; Official Test Pairs
& Mean Accuracy over 13 Generators \\
\midrule
GenVideo
& 8{,}000 / 8{,}302
& No Target Fakes; Source-Calibrated Transfer
& BAcc \\
DVF
& 2{,}450 / 3{,}938
& No Target Fakes; Target-Authentic Calibration
& BAcc \\
\bottomrule
\end{tabular}}
\end{table*}

\subsection{Training and Calibration Sources}
\label{sec:supp_training_sources}

Tables~\ref{tab:protocol_map} and~\ref{tab:protocol_evaluation} summarize our training, calibration, and evaluation sets. The full method and its fusion-only variant use the same expert heads and calibration videos.

\paragraph{Expert-Head Training.}
The four visual encoders remain frozen throughout training. We fit separate sets of expert heads on FaceVid, ViF-CoT-4K, GenBuster, and CoCoVideo. FaceVid and CoCoVideo use their official training splits. For ViF, we use 3{,}502 annotated deepfakes from ViF-CoT-4K and their 3{,}502 authentic counterparts. Our 20{,}000-video GenBuster subset contains 10{,}000 authentic videos and 2{,}500 deepfakes from each of CogVideoX, EasyAnimate, HunyuanVideo, and LTX-Video. Authentic videos reserved for calibration are removed before fitting the heads, as described below.

\paragraph{Investigator Training and Reuse.}
For each seed, we train one investigator on FaceVid and one on ViF. Each learns tool use from clean demonstrations, then undergoes counterfactual RL on videos near the fusion threshold. For ViF, artifact annotations guide teacher demonstrations but are omitted from the student inputs. The FaceVid investigator is used for FaceVid evaluation. The ViF investigator is reused on ViF, GenBuster, CoCoVideo, GenVideo, and DVF without further training. For GenBuster and CoCoVideo, the investigator's CLIP and V-JEPA~2 ViT-L witness heads are fitted on the corresponding dataset, alongside the visual expert heads. Appendices~\ref{sec:supp_agent} and~\ref{sec:supp_dev} describe investigator training.

\paragraph{Generator-Held-Out Folds.}
To obtain expert predictions for investigator training, we divide the training videos into four folds. Each expert head is fitted on three folds and predicts the remaining fold, giving out-of-fold (OOF) scores for videos it has not seen during training. Deepfakes are assigned to the generator groups specified below. Calibration and test videos are scored by averaging the logits from the four heads, as described in Section~\ref{sec:experts}.

For ViF, we organize the seven training configurations into four groups: HunyuanVideo, CogVideoX, Wan2.1, and Wan2.2, assigned to folds 0--3, respectively. Image-to-video and text-to-video variants within each group stay together. GenBuster assigns CogVideoX, EasyAnimate, HunyuanVideo, and LTX-Video to these four folds. CoCoVideo uses the groups \{Jimeng, Runway, Pika, Sora\}, \{Hailuo, Seedance, Vidu\}, \{Luma, Veo, Vivago\} and \{PixVerse, Kling, Midjourney\}. The general splitting procedure places larger generator groups first, assigning each to the fold with the fewest clips. Ties follow the dataset's file-list order and then the fold index. Authentic videos are assigned by a deterministic hash of the video path, except for the FaceVid sources described next.

FaceVid uses predefined folds. Fold 0 contains VidProM's \texttt{vc2\_videos}; fold 1 contains \texttt{pika\_videos} and \texttt{t2vz\_videos}; fold 2 contains \texttt{ms\_videos}, GenVidBench's CogVideo, and AIGVDBench's CausVid, Gen2 and Gen3; fold 3 contains the five FaceForensics++ manipulation methods. Authentic FaceForensics++ and GenVidBench clips are placed in folds 3 and 2, respectively; other authentic clips use the path hash. We assign folds before removing calibration videos and keep the assignments fixed across training seeds. All exclusions are applied before fitting any head.

\paragraph{Holding Out Calibration Videos.}
We select authentic calibration videos once with seed 0 and keep this selection fixed across training seeds. For FaceVid and ViF, we sample whole source groups within each authentic domain, approximately in proportion to the domain's size. We also remove training videos linked to the selected source identities. This excludes 2{,}711 FaceVid training clips, including the 2{,}001 calibration clips, and 1{,}204 ViF clips, including the 602 calibration clips. For GenBuster and CoCoVideo, we sample authentic training videos uniformly without replacement. GenBuster excludes the 600 selected clips; CoCoVideo excludes 600 authentic clips and their 600 generated counterparts.

After these exclusions, FaceVid has 9{,}625 authentic and 56{,}570 deepfake training clips, and ViF has 2{,}900 of each. GenBuster retains 9{,}400 authentic and 10{,}000 deepfake clips; CoCoVideo retains 8{,}500 of each. Each head then trains on three of the four folds within these remaining videos.

We identify linked FaceForensics++ videos using both source and donor identity identifiers in each manipulated clip's filename. For ViF, the first 11 filename characters identify the source, and all associated authentic and generated clips are excluded together. CoCoVideo pairs clips by filename within each generator group. These rules keep calibration videos and their linked clips out of head training; they do not make every GHCF fold disjoint by authentic source identity.

\paragraph{Use of Calibration during Development.}
Calibration statistics are reused when selecting models and investigator training recipes. The investigator also receives suspicion scores and evidence-tool percentiles computed from calibration data (See Section~\ref{sec:system2}). ViF's 602 calibration clips come from 402 source identities, and we split these clips, rather than identities, into reference and threshold subsets. Remark~\ref{rem:departures} explains how calibration reuse and shared identities affect the assumptions of the false-alarm bound.

\subsection{Evaluation Splits and Metrics}
\label{sec:supp_evaluation_protocol}

\paragraph{Detection and False-Alarm Metrics.}
We report BAcc on FaceVid OOD, GenVideo and DVF, together with deepfake-class F1 on FaceVid. BAcc is $(\mathrm{Recall}+\mathrm{Specificity})/2$, so the two classes contribute equally even when their counts differ. FAR is $1-\mathrm{Specificity}$, and deepfake-class F1 is $2\mathrm{TP}/(2\mathrm{TP}+\mathrm{FP}+\mathrm{FN})$. We compute AUROC for detectors that provide continuous scores.

\paragraph{Paired Accuracy on ViF.}
The released matching rule provides authentic references for 2{,}991 of the 2{,}995 ViF-Bench deepfakes; all methods omit the same four unmatched clips. We evaluate each matched pair, compute accuracy within each generator, and then average over the 19 generators. Some of the 165 authentic videos serve as references for multiple deepfakes and are counted once in each pair. For a generator $g$ with $N_g$ matched pairs, paired accuracy is $(\mathrm{TP}_g+\mathrm{TN}_g)/(2N_g)$. This differs from ordinary accuracy on the full set of 3{,}160 videos, which counts each video once.

\paragraph{Accuracy on GenBuster and CoCoVideo.}
GenBuster Test contains 300 CogVideoX, 400 EasyAnimate, 100 HunyuanVideo, and 200 LTX-Video deepfakes. Closed contains 100 deepfakes each from Gen3, Jimeng, Kling, Luma, and Pika, 200 from Sora, and 150 each from Vidu and Wanx. With 1{,}000 authentic videos in each split, we report accuracy over all 2{,}000 clips, giving each clip equal weight. CoCoVideo has 150 authentic/generated test pairs for each of its 13 generators. We compute accuracy for each generator and then average these scores.

\paragraph{Six-Benchmark Summary.}
We average GenBuster's Test and Closed scores to obtain one score for that benchmark. Avg is then the unweighted mean of the six benchmark scores, and Min is their minimum. These summaries combine each benchmark's own metric and training setup; they are not accuracy pooled over all test videos. Appendix~\ref{sec:supp_statistics} describes aggregation across training seeds and calibration splits.

\paragraph{Transfer to GenVideo and DVF.}
Both benchmarks use the ViF-trained heads and investigator. Our local GenVideo-Val collection contains 10{,}000 authentic and 8{,}302 generated videos across ten source groups. We reserve 2{,}000 authentic videos for a separate recalibration experiment, leaving 8{,}000 authentic and 8{,}302 generated videos for testing, or 16{,}302 videos in total. The main result in Table~\ref{tab:main} uses ViF calibration; the recalibration experiment uses the reserved GenVideo clips and the same test set. On DVF, the main result uses 300 authentic target videos for calibration and evaluates the remaining 6{,}388 videos. We sample both target calibration sets uniformly without replacement using seed 0. None of these reserved clips are included in testing, and neither protocol uses target deepfakes for training or calibration.

The additional commercial-generator transfer study in Appendix~\ref{sec:supp_zeroshot} also keeps the ViF heads fixed and uses 300 authentic target videos for calibration. This differs from the main GenBuster and CoCoVideo experiments, where the heads are trained on each benchmark's own training data.

\paragraph{Supplementary DF26 Evaluation.}
We evaluate \method with FaceVid or ViF source calibration and, separately, recalibration on authentic DF26 clips using two folds. Appendix~\ref{sec:supp_df26} describes these experiments.

\paragraph{Labels and Source Groups.}
Before computing metrics, we use $y=0$ for authentic clips and $y=1$ for deepfakes. CoCoVideo's released indices instead assign label 1 to original videos, so we check the original/generated path and generation-method field when converting its labels. Video paths are used to match predictions to labels, source groups, and exclusion lists. Group counts follow the dataset's configuration or collection names, which do not always correspond to distinct model families. For example, GenVideo includes the mixed \texttt{WildScrape} collection, while DVF lists \texttt{StableVideo} and \texttt{StableVideoDiffusion} separately. Re-encoded and post-processed copies are evaluated within their respective processing studies, not as additional benchmarks.

\paragraph{Shared Settings.}
Unless stated otherwise, the main results use two expert views, $\alpha=5\%$, $w_I=1$, and an investigation budget of $\rho=0.3$. We report results across ten training seeds, with five calibration splits per seed. Appendix~\ref{sec:supp_agent} describes the input views and training settings, and Appendix~\ref{sec:supp_statistics} explains averaging and confidence intervals. Individual experiments specify any changes to the seeds or input views.
\subsection{Result Aggregation and Uncertainty}
\label{sec:supp_statistics}

We distinguish variation across training seeds from uncertainty due to the evaluation sample. The reported standard deviations summarize the former, while paired confidence intervals assess the latter with fitted runs held fixed.

\paragraph{Result Aggregation.}
We align each method's per-video predictions with the evaluation manifests using file paths and labels, then compute metrics under the coverage and failure rules in Appendix~\ref{sec:supp_baselines}. Each run contributes its own metric; predictions are neither pooled nor voted across seeds. The main results use the six-benchmark summary defined in Appendix~\ref{sec:supp_evaluation_protocol}. Development and matched-calibration comparisons follow their respective protocols, and all entries and gains are rounded only after aggregation.

\paragraph{Training Seeds and Calibration Splits.}
Let $M_{s,h}$ denote a benchmark metric for training seed $s$ and reference/threshold split $h$. We first average the five calibration splits within each seed, then compute the mean and sample standard deviation over ten seeds:
\begin{equation}
\begin{aligned}
\bar M_s&=\frac{1}{5}\sum_{h=0}^{4}M_{s,h},\qquad
\bar M=\frac{1}{10}\sum_{s=0}^{9}\bar M_s,\\
\mathrm{SD}&=\sqrt{\frac{1}{9}\sum_{s=0}^{9}(\bar M_s-\bar M)^2}.
\end{aligned}
\end{equation}
The investigator episodes stored for each seed remain fixed across its five calibration splits. The standard deviation therefore describes variation between seed-level averages, rather than treating all splits as independent training runs. Captions identify single-seed development comparisons and alternative input views.

\paragraph{Paired Confidence Intervals.}
Each bootstrap replicate uses the same resampled evaluation items for both methods and all training seeds, then averages the metric differences. FaceVid, GenBuster, GenVideo, and DVF comparisons resample videos within each class. CoCoVideo resamples within generator and class before averaging over generators. Fitted models, investigator episodes, and calibration splits remain fixed, so these intervals measure evaluation-sample variation and do not include training or model-selection uncertainty. Comparisons with incomplete baseline coverage use the shared covered subset.

\paragraph{Shared References on ViF.}
ViF's 2{,}991 matched deepfake videos share 165 authentic references across 19 generators. To preserve this dependence, we resample authentic-source groups, retaining every associated deepfake/reference pair across generators, and recompute the mean paired accuracy. With 4{,}000 replicates and resampling seed 0, the 95\% interval for \method minus Skyra is $[-3.04,2.04]$ points.

\subsection{Metadata Shortcuts and Input Preprocessing}
\label{sec:supp_meta}

\begin{table}[t]
\centering
\caption{\textbf{Metadata-only diagnostics without pixel inputs.} The FaceVid classifier is a class-balanced depth-3 tree fitted on the training split. Authentic ViF training clips are resampled to match their paired deepfake's frame rate; all authentic ViF test clips run at 16\,fps.}
\label{tab:supp_meta}
\resizebox{\columnwidth}{!}{%
\begin{tabular}{llr}
\toprule
\textbf{Benchmark}
& \textbf{Metadata-Only Predictor}
& \textbf{Score} \\
\midrule
FaceVid-Forensics-100K OOD
& Depth-3 Tree on Frame Rate
& 98.89 BAcc \\
GenVideo (Val)
& Frame Height Alone
& 0.9999 AUROC \\
GenVideo (Val)
& Depth-4 Tree on Container Metadata
& 99.6 BAcc \\
ViF-Bench Test
& Frame Rate + Duration (Timestamps)
& 69.5 Paired Acc \\
ViF-Bench Test
& + Frame Size (Gradient Boosting)
& 79.4 Paired Acc \\
\bottomrule
\end{tabular}
}
\end{table}

\paragraph{Differences between Video Sources.}
Authentic and generated videos may differ in frame rate or resolution because they originate from different data sources. Such differences can predict the label without providing visual evidence of synthesis or manipulation. We examine the container metadata before decoding and preprocessing, linking each record to its label and source group through the video path.

FaceVid clips are $256\times256$, but the OOD split has median frame rates of 29.97\,fps for authentic videos and 16\,fps for deepfakes. Authentic ViF training clips match their generated counterparts' frame rates, whereas all authentic ViF-Bench clips use 16\,fps. Authentic GenVideo clips use $298\times224$ frames at 3\,fps; its generated clips have varied dimensions and frame rates.

\paragraph{Prediction from Metadata Alone.}
Table~\ref{tab:supp_meta} measures how well these properties distinguish authentic videos from deepfakes without pixel inputs. Frame rate alone is highly predictive on FaceVid OOD, and frame height almost perfectly separates the classes on GenVideo validation. The ViF results show that timing and frame-size information can support detection. These diagnostics motivate controlling access to metadata when evaluating visual evidence.

\paragraph{Input Preprocessing.}
\method samples frames by index and uses square crops to avoid directly supplying these cues. For a video with $N$ decoded frames, the uniform view uses 16 frames at indices $\mathrm{round}(\mathrm{linspace}(0, N-1, 16))$. Each frame is center-cropped to a square and resized to $256\times256$, using area interpolation for downscaling and bicubic interpolation for upscaling. Short clips may repeat indices, so the 16 inputs need not be distinct frames. Expert fusion also uses the offset view described in Appendix~\ref{sec:supp_agent}, followed by any encoder-specific resizing. The investigator starts with the uniform view and can request additional pixel-tool outputs. Frame rate, duration, timestamps, native resolution, and native aspect ratio are not provided. Baselines retain their released preprocessing, as specified in Appendix~\ref{sec:supp_baselines}.

\paragraph{Residual Association with Metadata.}
We also check whether the fused score remains associated with metadata on FaceVid OOD. Among authentic videos, its correlations with frame rate and frame count are near zero. BAcc on frame-rate-matched subsets falls both above and below the overall score, showing no consistent direction of change. Authentic videos with low frame rates are nevertheless flagged somewhat more often. These checks suggest limited association with the measured metadata, while leaving open the possibility of reliance on correlated visual cues.

\section{Model Configurations and Training Details}
\label{sec:supp_agent}

\begin{table*}[t]
\centering
\caption{\textbf{Visual encoders and fitted heads.} All inputs contain 16 frames; resolutions are square. Feature tensors are shown after concatenating the selected depths and token summaries. Frozen counts include the active visual encoder, excluding text encoders and unused V-JEPA predictors. Head counts are per fold; four fitted heads are retained for each expert.}
\label{tab:supp_encoders}
\resizebox{\textwidth}{!}{%
\begin{tabular}{lccccll}
\toprule
\textbf{Encoder} & \textbf{Resolution} & \textbf{Feature Tensor} & \textbf{Frozen (M)} & \textbf{Head Parameters} & \textbf{Head} & \textbf{Role} \\
\midrule
SigLIP2-L/16 & 256 & $16\times6{,}144$ & 315.956 & 9{,}498{,}626 & Temporal Transformer & Fusion \\
CLIP-L/14 & 224 & $16\times6{,}144$ & 303.180 & 9{,}498{,}626 & Temporal Transformer & Fusion and Witness \\
DINOv3 ViT-H+/16 & 256 & $16\times7{,}680$ & 840.593 & 10{,}288{,}130 & Temporal Transformer & Fusion \\
V-JEPA~2 ViT-g & 256 & $8\times4{,}224$ & 1{,}012.174 & 12{,}673 & Linear Scorer & Fusion \\
V-JEPA~2 ViT-L & 256 & $8\times3{,}072$ & 303.885 & 9{,}217 & Linear Scorer & Witness Only \\
\bottomrule
\end{tabular}}
\end{table*}

\subsection{Expert Heads}
\label{sec:supp_heads}

We use three appearance experts and one temporal expert for fusion, keeping their pretrained visual encoders frozen and training only the detection heads. The investigator accesses CLIP and a smaller V-JEPA~2 encoder through two witness tools. Below, we specify the pretrained checkpoints and how we extract features for detection; Table~\ref{tab:supp_encoders} summarizes their dimensions and parameter counts.

\paragraph{SigLIP2-L/16.}
SigLIP2~\cite{siglip2} combines image--text learning with captioning and self-supervised objectives. We use its visual encoder, \nolinkurl{google/siglip2-large-patch16-256}, as an appearance expert, with $256\times256$ frames and features from depths 8, 16, and 24. Since this encoder has no class token, the global feature is the patch-token mean at the intermediate depths and the attention-pooled output at the final depth. Each input channel is normalized with mean and standard deviation 0.5.

\paragraph{CLIP-L/14.}
CLIP~\cite{clip} learns visual representations through image--text contrastive training. Its visual encoder contributes to expert fusion and serves as the investigator's appearance witness. We use \nolinkurl{openai/clip-vit-large-patch14} with $224\times224$ frames and CLIP normalization. At depths 8, 16, and 24, we extract the class token and mean patch feature before the image--text projection. CLIP is selected as the appearance witness based on single-expert validation performance, with witness scores computed from the uniform input view.

\paragraph{DINOv3 ViT-H+/16.}
DINOv3~\cite{dinov3} provides global and dense image representations learned through self-supervision. We use the distilled ViT-H+/16 checkpoint, \nolinkurl{facebook/dinov3-vith16plus-pretrain-lvd1689m}, as the third appearance encoder. Inputs are $256\times256$ frames with ImageNet normalization. At depths 10, 21, and 32, we apply the encoder's final LayerNorm and extract the class token and mean patch feature, excluding register tokens from the patch average.

\paragraph{V-JEPA~2 ViT-g.}
V-JEPA~2~\cite{vjepa2} learns video representations by predicting features of masked video regions from visible context. Its ViT-g encoder supplies the temporal score for fusion. We use \nolinkurl{facebook/vjepa2-vitg-fpc64-256} with $256\times256$ frames and features from depths 13, 26, and 40. Although the checkpoint is configured for 64-frame clips, our extraction protocol uses 16 frames. The tubelet size of two produces eight temporal units, each represented by concatenated spatial means from the three depths. Inputs use ImageNet normalization; the pretrained predictor is not used.

\paragraph{V-JEPA~2 ViT-L.}
The smaller ViT-L encoder serves as the investigator's temporal witness. Its score is available through the witness tool and is not part of the four-expert fusion. We use \nolinkurl{facebook/vjepa2-vitl-fpc64-256} with features from depths 8, 16, and 24. As with ViT-g, we use 16 frames at $256\times256$ resolution, ImageNet normalization, and spatial averaging within each tubelet, without the pretrained predictor. Witness scores use the uniform view.

\paragraph{Detection Heads.}
For each appearance expert, we concatenate the global and mean patch features from the three selected depths. The head projects these frame features to width 512 and adds learned positional embeddings. A two-layer temporal transformer, with eight attention heads, feed-forward width 2{,}048 and dropout 0.1, aggregates the sequence through a learned query token. LayerNorm and a linear layer map the query output to a video logit; a separate linear layer maps each projected frame feature to a unit logit. The V-JEPA heads apply LayerNorm and a linear scorer to each tubelet and average the logits. Both head types minimize video-level BCE plus $0.5$ times the mean unit-level BCE, with each unit inheriting the video label.

\begin{table*}[t]
\centering
\caption{\textbf{Training configuration.} Head epochs are shown as FaceVid / ViF, GenBuster, and CoCoVideo. Investigator training updates only the language model's low-rank adaptation parameters; all base-model weights remain frozen. The supervised fine-tuning batch uses 2 GPUs, one transcript per GPU and 8 gradient-accumulation steps. For reinforcement learning, batch size denotes video draws per iteration.}
\label{tab:supp_training_config}
\resizebox{\textwidth}{!}{%
\begin{tabular}{llcccll}
\toprule
\textbf{Stage}
& \textbf{Optimizer}
& \textbf{Learning Rate}
& \textbf{Weight Decay}
& \textbf{Batch Size}
& \textbf{Sampling Unit}
& \textbf{Schedule} \\
\midrule
Appearance Heads
& AdamW
& Peak $3\times10^{-4}$
& 0.05
& 256
& Clips
& 8 / 15 Epochs; One-Cycle \\

Temporal Heads
& AdamW
& Peak $10^{-3}$
& 0.05
& 256
& Clips
& 10 / 20 Epochs; One-Cycle \\

Supervised Fine-Tuning
& AdamW
& $10^{-4}$
& 0
& 16
& Transcripts
& 2 Epochs; Warmup + Cosine Decay \\

Reinforcement Learning
& AdamW
& $10^{-5}$
& 0
& 32
& Video Draws (6 Episodes Each)
& 30 Iterations; 4 Updates Each \\
\bottomrule
\end{tabular}}
\end{table*}

\paragraph{Input Views and Scoring.}
We center-crop decoded frames to a square and resize them to $256\times256$ before encoder-specific normalization and any further resizing. Encoder-side resizing uses antialiased bicubic interpolation. Expert fusion uses two 16-frame views: a uniform frame-index grid and the same grid shifted by half a sampling interval, clipped to the video boundaries. The investigator and its tools use the uniform view. Short clips may repeat sampled frames.

During GHCF training, each video receives predictions from the head fitted without its assigned fold. Calibration and test videos receive the mean logit across the four fitted heads. Two-view fusion then averages the uniform and offset video logits; witness scores retain the uniform view. Appendix~\ref{sec:supp_protocols} specifies the folds and calibration exclusions.

\paragraph{Witness-Head Reuse.}
GenBuster and CoCoVideo use witness heads fitted on their respective training splits while reusing the ViF-trained investigator. GenVideo and DVF reuse both the ViF heads and investigator. All detection heads receive only visual encoder features.

\paragraph{Head Optimization.}
Within each fitting fold, we weight positive-class BCE terms by the authentic-to-deepfake count ratio and add Gaussian feature noise with standard deviation 0.02. The one-cycle schedule increases the learning rate over the first 5\% of its steps. Table~\ref{tab:supp_training_config} summarizes the optimization settings for expert-head training and investigator fine-tuning.

\subsection{Inference and Tool Use}\label{sec:supp_tools}

\paragraph{Qwen2.5-VL-7B-Instruct.}
We use \nolinkurl{Qwen/Qwen2.5-VL-7B-Instruct}~\cite{qwen25vl}, an instruction-tuned vision-language model, as the investigator. The model supports visual text understanding, object localization, and tool use. In our framework, it examines sampled frames and tool responses through a multimodal dialogue to gather evidence for its verdict. Training updates only LoRA adapters in the language model; the base model, including its visual tower, remains frozen. The selected adapter contains 161{,}480{,}704 trainable parameters, or 1.91\% of the 8{,}453{,}647{,}360 parameters including adapters.

\paragraph{Inference and Calibration Settings.}
The main configuration uses $\alpha=5\%$, $w_I=1$ and $\rho=0.3$. Each investigation allows at most eight tool calls, excluding verdict submission, and ten dialogue rounds. Responses are sampled at temperature 0.2 with a limit of 2{,}048 tokens. A random permutation assigns the first $\lfloor|\mathcal{C}|/2\rfloor$ calibration clips to $\mathcal{R}$ and the remainder to $\mathcal{Q}$. The gate ranks videos in a batch by
\[
d(v)=\left|\log(\Phi(v)+\varepsilon)-\log(\hat\tau+\varepsilon)\right|,
\qquad \varepsilon=10^{-12},
\]
and selects $\operatorname{round}(\rho|\mathcal{V}|)$ videos, with nearest-integer rounding and ties at half-integers rounded to even. Equal distances retain input order. Before test evaluation, the investigator scores every video in $\mathcal{C}$, including clips the test-batch gate would not select. These outputs define its reference p-values and the re-fusion threshold.

We apply the order-statistic rule in Eq.~\eqref{eq:fusion_threshold} to $\Phi$ and $\Phi_2$ on the same threshold subset $\mathcal{Q}$. Algorithm~\ref{alg:inference} summarizes batch inference.

If the investigator submits no verdict within ten rounds, its score defaults to the prompt suspicion $u(v)$. We calibrate and fuse this fallback score in the same way as a submitted score; the fallback does not restore the fusion-only decision. This occurs in 142 of 149{,}148 stored test episodes and 42 of 41{,}030 calibration episodes under the main protocol, approximately 0.10\% in each case.

\begin{algorithm}[t]
\caption{Batch inference of \method after calibration}
\label{alg:inference}
\small
\begin{algorithmic}[1]
\Require calibrated expert statistic $\Phi$ and threshold $\hat\tau$; configured investigator $I$, reference outputs $\{p_I(r):r\in\mathcal{R}\}$, threshold $\hat\tau_2$ and weight $w_I$; batch $\mathcal{V}$ and rate $\rho$
\State Compute $\Phi(v)$ for every $v\in\mathcal{V}$
\State $\mathcal{E}(\mathcal{V})\gets$ the $\operatorname{round}(\rho|\mathcal{V}|)$ videos closest to $\hat\tau$ in log-score distance
\For{$v\in\mathcal{V}$}
  \If{$v\notin\mathcal{E}(\mathcal{V})$}
    \State $\delta_{\mathcal{V}}(v)\gets\mathbf{1}[\Phi(v)>\hat\tau]$
  \Else
    \State Obtain $p_I(v)$ and its reference p-value $\pi_I(v)$
    \State $\Phi_2(v)\gets\Phi(v)-2w_I\log\pi_I(v)$
    \State $\delta_{\mathcal{V}}(v)\gets\mathbf{1}[\Phi_2(v)>\hat\tau_2]$
  \EndIf
\EndFor
\State \Return $\{\delta_{\mathcal{V}}(v):v\in\mathcal{V}\}$
\end{algorithmic}
\end{algorithm}

\begin{table*}[t]
\centering
\caption{\textbf{Investigator toolbox.} Expert outputs provide learned scores; pixel and evidence tools inspect the unchanged video. All tools are fixed during investigator training.}
\label{tab:supp_toolbox}
\renewcommand{\arraystretch}{1.15}
\setlength{\tabcolsep}{6pt}
\setlength{\lightrulewidth}{0.3pt}
\resizebox{\textwidth}{!}{%
\begin{tabular}{lll}
\toprule
\textbf{Tool}
& \textbf{Arguments}
& \textbf{Returned Evidence} \\
\midrule
\rowcolor[HTML]{E1EBE5}
\multicolumn{3}{c}{\textbf{Witness}} \\
\midrule
\texttt{frame\_artifact\_expert}
& None
& CLIP video and per-frame probabilities, the three highest-scoring frames, and the reliability card. \\

\texttt{temporal\_physics\_expert}
& None
& V-JEPA~2 ViT-L video probability and reliability card. \\
\midrule
\rowcolor[HTML]{E1EBE5}
\multicolumn{3}{c}{\textbf{Pixel}} \\
\midrule
\texttt{view\_frames}
& \texttt{indices}; optional \texttt{zoom}
& Up to four frames, optionally cropped to a normalized zoom box. \\

\texttt{frame\_differences}
& Optional \texttt{start}, \texttt{end}
& Per-step change magnitudes and an amplified image of the largest change. \\

\texttt{frequency\_spectrum}
& Optional \texttt{index}
& Frame log spectrum and high-to-low-frequency ratio. \\
\midrule
\rowcolor[HTML]{E1EBE5}
\multicolumn{3}{c}{\textbf{Evidence}} \\
\midrule
\texttt{temporal\_profile}
& None
& Luminance, sharpness, saturation, and color variation, with authentic-reference percentiles. \\

\texttt{motion\_consistency}
& None
& Camera-motion estimates, residual tracks, and motion irregularities, with reference percentiles. \\

\texttt{text\_stability}
& None
& \begin{tabular}[c]{@{}l@{}}
Optical character recognition (OCR) region counts and text variation \\
across sampled frames, with reference percentiles.
\end{tabular} \\
\midrule
\rowcolor[HTML]{E1EBE5}
\multicolumn{3}{c}{\textbf{Decision}} \\
\midrule
\texttt{submit\_verdict}
& \begin{tabular}[c]{@{}l@{}}
\texttt{p\_fake}, \texttt{forgery\_type}, \\
\texttt{rationale}
\end{tabular}
& Deepfake probability $p_I$, forgery category and rationale. \\
\bottomrule
\end{tabular}}
\end{table*}

\paragraph{Prompt Calibration.}
With $m=|\mathcal{Q}|$, the displayed suspicion is
\[
u(v)=\frac{|\{q\in\mathcal{Q}:\Phi(q)<\Phi(v)\}|}{m+1},
\]
which expresses the video's rank relative to authentic threshold samples, not a posterior probability. Prompts use a fixed reference/threshold split with seed 0. Each calibration clip is removed from its assigned half before computing its suspicion; test clips use the full reference and threshold subsets. Evaluation over five calibration splits reuses these stored episodes and recomputes fusion thresholds and escalation sets. Thus, the split comparison holds the investigations fixed. Reliability cards summarize out-of-fold recall on held-out generator families and FAR on $\mathcal{C}$; evidence-tool percentiles also use $\mathcal{C}$. Appendix~\ref{sec:supp_theory} explains how these dependencies affect the scope of the guarantee.

\paragraph{Investigator Prompts.}
Figures.~\ref{fig:supp_system_prompt} and~\ref{fig:supp_user_template} reproduce the system prompt and first user-message template used in the selected configuration, respectively. Braced fields are filled from the case's calibrated suspicion, flag level, and tool budget; the first user message is followed by the 16 overview images. The templates' phrase ``uniformly in time'' refers to the implemented frame-index sampling. Their nominal false-alarm wording is a prompt instruction; realized FAR is measured on the evaluation videos.

\paragraph{Tools and Observations.}
The investigator combines expert reports with inspection of frames and measurements of temporal consistency. Table~\ref{tab:supp_toolbox} lists the tools, which return structured text and, where applicable, images. We cache witness logits, frame statistics, point tracks, and OCR outputs; pixel tools operate on the same decoded frames. Evidence tools express each measurement as a percentile of authentic calibration values, highlighting unusual observations for inspection. These descriptive percentiles are distinct from the p-values used in fusion. Motion analysis uses offline CoTracker3~\cite{cotracker3} with a $20\times20$ query grid on the first frame. RapidOCR probes frames 0, 8, and 15, then reads the remaining frames if any probe detects text.

\paragraph{Tool Interface and Serving.}
Frame indices refer to the 16 overview frames, numbered 0--15. \texttt{view\_frames} returns at most four images; its optional \texttt{zoom} is $[x_0,y_0,x_1,y_1]$ in normalized crop coordinates. Difference intervals require an end index larger than the start. The terminal \texttt{submit\_verdict} call requires \texttt{p\_fake} in $[0,1]$, a \texttt{forgery\_type} from \texttt{real}, \texttt{face\_swap}, \texttt{face\_reenactment}, or \texttt{fully\_synthetic}, and a \texttt{rationale} of 2--4 sentences citing the decisive evidence. Measured evidence returns scalar values with \texttt{pct\_of\_authentic\_below}; these descriptive percentiles are not calibrated deepfake probabilities.

We serve the investigator with vLLM 0.30.0 in bfloat16 without quantization, using one replica per GPU and tensor parallel size one. The maximum context is 32{,}768 tokens, with at most 64 images per request. Training and serving use the same Qwen2.5 tool-call chat format. Appendix~\ref{sec:supp_cost} describes hardware and amortized inference measurements.

\subsection{Investigator Training}\label{sec:supp_training}

\paragraph{Qwen3.8-Max.}
Qwen3.8-Max~\cite{qwen38max} is designed for coding, research, and long-horizon tasks that require multi-step problem solving. We use it as a privileged teacher to generate complete tool-use demonstrations for the smaller investigator. Each demonstration records the teacher's tool requests, returned observations, and final verdict, showing the student how an investigation proceeds from initial expert reports to visual inspection and a decision.

The teacher receives the video, its label, and any available artifact annotations. For ViF deepfakes, we map up to four annotated artifacts to the sampled-frame indices and visible crop coordinates. For FaceVid, we provide a reference explanation truncated to 900 characters. Authentic examples receive an authentic-label hint. These annotations appear only in the teacher's system prompt. Student transcripts retain the common system prompt and the observed interaction; episodes whose responses mention privileged information are excluded.

Teacher episodes are generated from eligible training videos, filtered for valid tool use and accurate verdicts, and balanced between classes. We duplicate episodes near the escalation threshold to give them greater training weight. Table~\ref{tab:supp_clean_demonstrations} distinguishes the number of source videos from the resulting transcript counts.

The alternative counterfactual supervised fine-tuning (SFT) sets contain 5{,}973 ViF and 4{,}374 FaceVid transcripts. For each training seed, we fit investigator adapters using the same demonstration sets but different adapter initialization, minibatch order, and reinforcement learning (RL) sampling. Rationale grounding is assessed separately in Appendix~\ref{sec:supp_judge}.

\begin{table}[t]
\centering
\caption{\textbf{Selected clean demonstrations after filtering and balancing.} Counts include escalation-band duplication.}
\label{tab:supp_clean_demonstrations}
\resizebox{\linewidth}{!}{%
\begin{tabular}{lcccc}
\toprule
\textbf{Source}
& \textbf{Videos}
& \textbf{Transcripts}
& \textbf{Authentic}
& \textbf{Deepfake} \\
\midrule
ViF & 2{,}023 & 3{,}638 & 1{,}819 & 1{,}819 \\
FaceVid & 2{,}517 & 2{,}910 & 1{,}455 & 1{,}455 \\
\bottomrule
\end{tabular}}
\end{table}

\paragraph{Counterfactual Evidence.}
Counterfactual evidence creates conflicts between expert reports and the target video's visual observations. For a deepfake, the donor is an authentic training video not flagged by expert fusion; for an authentic video, it is a flagged training deepfake. We uniformly choose to replace the appearance witness, the temporal witness, or both, leaving the target video, label, and pixel/evidence-tool outputs unchanged. Replacing both witnesses also replaces the displayed suspicion with the donor's value; replacing one uses the mean of the original and donor suspicions. The selected recipe introduces these counterfactual outputs only during reinforcement learning.

\paragraph{Filtering and Supervised Fine-Tuning.}
For each video and evidence type, we retain up to two episodes with the lowest Brier error, breaking ties in favor of fewer tool calls. Each episode must submit a verdict satisfying $|p_I-y|\le0.4$, contain no malformed or failed tool calls, corrective prompts, or truncated responses, and avoid mentioning privileged information. Calibration exclusions and held-out validation deepfakes are removed from the training pool. Episodes whose displayed suspicion lies in $[0.82,0.99]$ are duplicated. The clean set retains only unmodified expert evidence and subsamples the larger class. We fine-tune the student on assistant turns for two epochs using low-rank adaptation (LoRA) with rank 64.

SFT uses approximately 5\% warmup with cosine decay; Table~\ref{tab:supp_training_config} specifies the optimizer and effective batch size. LoRA uses alpha 128 and dropout 0.05 on the language model's attention projections (query, key, value, and output) and MLP projections (gate, up, and down); biases remain frozen. The transcript limit is 12{,}000 tokens, and the longest selected clean transcript contains 7{,}567 tokens. The base model runs in bfloat16, with gradient norms clipped at 1.0 for the LoRA updates. The two-epoch runs perform 456 ViF and 364 FaceVid optimizer updates.

\paragraph{Reward.}
For an episode that submits a verdict, the reward is
\begin{equation}
\begin{aligned}
r={}&1-(p_I-y)^2-\lambda_c n_{\mathrm{calls}}\\
&-\lambda_e n_{\mathrm{err}}-\lambda_v\mathbf{1}[n_{\mathrm{vis}}=0],
\end{aligned}
\label{eq:reward}
\end{equation}
where $n_{\mathrm{calls}}$ counts tool calls, $n_{\mathrm{err}}$ counts tool errors, corrective prompts, and truncated outputs, and $n_{\mathrm{vis}}$ counts calls to pixel or evidence tools. The Brier term penalizes confident errors in either class. We use $\lambda_c=0.005$ per tool call and $\lambda_e=0.1$ per error event. The penalty $\lambda_v=0.1$ encourages additional inspection by penalizing episodes with no pixel or evidence calls, even though the overview frames are already visible. This term records tool invocation, not visual attention or call success. Episodes without a submitted verdict receive reward $-0.5$.

\paragraph{Escalation-Band Reinforcement Learning.}
We focus RL on videos near the decision threshold by drawing equal numbers of authentic and deepfake training videos from the out-of-fold suspicion band $[0.82,0.99]$. Each rollout replaces witness evidence with probability 0.3 using an eligible opposite-class donor from this band. We sample six episodes per video at temperature 1 and normalize their advantages within the group. Groups without reward variation are dropped. Training uses 30 iterations of 32 video draws, with four LoRA optimizer steps per iteration at learning rate $10^{-5}$ and zero weight decay. We retain the final adapter. One A100 generates rollouts and a second performs updates. The schedule requests 960 video draws and 5{,}760 episodes per run, with repeated draws permitted.

\paragraph{Policy-Gradient Objective.}
The update uses group-relative reward normalization inspired by group relative policy optimization (GRPO)~\cite{grpo}. For an episode $e$ of training video $v$, we form the detached advantage
\[
A_e=\frac{r_e-\bar r_v}{\sigma_v+\varepsilon_A},\qquad
\varepsilon_A=10^{-6},
\]
where $\bar r_v$ and $\sigma_v$ are the mean and population standard deviation of rewards across the returned episodes for $v$. Groups with $\sigma_v<10^{-6}$ are excluded. Within an update minibatch $\mathcal{B}$, let $\mathcal{B}_{\mathrm{valid}}$ contain episodes whose transcripts can be re-rendered within the token limit. If $\mathcal{T}_e$ indexes the assistant tokens of episode $e$, the minimized loss is
\begin{equation}
\begin{aligned}
\mathcal{L}_{\mathrm{RL}}(\theta)
={}& -\frac{1}{|\mathcal{B}|}
\sum_{e\in\mathcal{B}_{\mathrm{valid}}}
\frac{A_e}{|\mathcal{T}_e|}\\
&\qquad\cdot
\sum_{t\in\mathcal{T}_e}
\log P_\theta(a_{e,t}\mid h_{e,t}).
\end{aligned}
\label{eq:supp_rl_loss}
\end{equation}
Here $a_{e,t}$ is an assistant token, $h_{e,t}$ is its preceding multimodal conversation, and $P_\theta$ is the student policy. We optimize this group-normalized REINFORCE loss without probability-ratio clipping or a reference-policy KL penalty. Gradient norms are clipped at 1.0. The four minibatches share a rollout batch, with no importance-ratio correction after the first optimizer step. Episodes skipped during transcript rendering remain in the group reward statistics. RL in the development tables denotes this update.

\clearpage
\begin{figure*}[t]
\centering
\begin{prompt}{Investigator System Prompt}
You are a video forensics investigator. Decide whether a short video is authentic camera footage or fake (face swap, face reenactment, or fully AI-generated).

You see 16 frames sampled uniformly in time (index 0-15). You are NOT given frame rate, duration, file size, or any container metadata; do not speculate about them.

Investigate with the tools, then call submit_verdict exactly once.

How to weigh evidence:

- Expert tools return probabilities together with a reliability note. Treat them as noisy witnesses, not ground truth.
- The appearance expert is precise but misses many unseen or image-to-video generators: a LOW appearance score is only weak evidence of authenticity, while a HIGH score is strong evidence of forgery.
- Your own visual impression is prone to false alarms on real footage (compression blur, motion blur, low light, unusual scenes). Only count concrete, localized anomalies: warping or melting geometry, identity or texture drift between frames, inconsistent hands/teeth/eyes/text, physically implausible motion, lighting that contradicts the scene, boundary shimmer around a face.
- When the experts disagree, look for decisive visual or temporal evidence before deciding, and express remaining uncertainty through p_fake rather than forcing 0 or 1.
- The evidence tools (temporal_profile, motion_consistency, text_stability) report each metric with pct_of_authentic_below, the share of authentic reference videos with a smaller value. Values above about 0.95 are unusual for real footage; treat them as leads to verify by looking at the indicated frames and regions, not as proof on their own.

Budget: at most {max_tool_calls} tool calls. Prefer cheap, informative steps; stop as soon as the evidence is decisive.
\end{prompt}
\caption{\textbf{Investigator system prompt.} The same system prompt is stored in student demonstrations.}
\label{fig:supp_system_prompt}
\end{figure*}

\begin{figure*}[t]
\centering
\begin{prompt}{First User-Message Template}
New case.
Fast detector (fusion of the appearance and temporal experts, calibrated on authentic videos): suspicion = {suspicion:.3f}, i.e. this video looks more suspicious than {pct:.0f}% of authentic reference videos. The fast detector flags videos with suspicion >= {flag_level:.2f} (about {alpha_pct:.0f}% false alarms on authentic footage) and has {flag_state} this one. It can miss fakes from unseen generators and occasionally flags authentic footage, so verify the evidence yourself.
Below are all 16 frames (indices 0-15, in temporal order).
\end{prompt}
\caption{\textbf{First user-message template.} Teacher-only annotations are appended to the teacher's request and omitted from the stored student input.}
\label{fig:supp_user_template}
\end{figure*}

\clearpage

\section{False-Alarm Analysis}
\label{sec:supp_theory}

This appendix studies false alarms for the expert-fusion rule in
Section~\ref{sec:conformal} using standard split-conformal
results~\cite{angelopoulos2021,bates2023outliers,vovk2012,vovk2005}.
We distinguish marginal false-alarm control from uncertainty for a single
calibration set, then discuss the selective cascade and evaluate calibration
on previously unused authentic videos.

\subsection{Calibration Setting and Assumptions}
\label{sec:supp_theory_setting}

Let $P_0$ denote the authentic-video distribution represented by the
calibration set $\mathcal{C}$, and let $v$ be an authentic test video.
For a prespecified level $0<\alpha<1$, we split $\mathcal{C}$ into a reference
subset $\mathcal{R}$ and a threshold subset $\mathcal{Q}$ of fixed sizes, with
$m=|\mathcal{Q}|\ge1$. We use the following assumptions.
\begin{enumerate}[label=\textup{(A\arabic*)},ref=\textup{(A\arabic*)},leftmargin=*,itemsep=2pt,topsep=3pt]
  \item \label{asm:fixed}
  \emph{Fixed Scoring.} The measurable expert score functions $s_1,\dots,s_K$
  and weights $w_1,\dots,w_K\ge0$ are fixed independently of
  $\mathcal{C}\cup\{v\}$. Calibration and test videos use the same score
  functions and weights. All probabilities below condition on these functions
  and weights.
  \item \label{asm:split}
  \emph{Independent Split.} The index partition assigning calibration videos
  to $\mathcal{R}$ and $\mathcal{Q}$ is drawn independently of the videos in
  $\mathcal{C}\cup\{v\}$.
  \item \label{asm:exch}
  \emph{Exchangeability.} Conditional on the score functions and weights in
  \ref{asm:fixed}, the videos in $\mathcal{C}\cup\{v\}$ are exchangeable.
  Independent draws from $P_0$ satisfy this assumption.
\end{enumerate}
If scoring is randomized, we treat each observation as $(v,\xi)$, where the
auxiliary draws $\xi$ are i.i.d.\ and independent of the videos and the split.
Probabilities then include these draws, and conditioning on $\mathcal{R}$
fixes the reference observations and their auxiliary draws.
Remark~\ref{rem:departures} relates these assumptions to the implementation.

Given $\mathcal{R}$, the reference p-values in Eq.~\eqref{eq:pvalue} and
$\Phi(v)=\sum_{j=1}^{K}w_j\bigl(-2\log\pi_j(v)\bigr)$ are fixed functions of
the observation, augmented by $\xi$ when needed.
Let $k=\lceil(1-\alpha)(m+1)\rceil$, and let
$\Phi_{(k)}(\mathcal{Q})$ denote the $k$-th smallest threshold-set score.
The finite-sample threshold is
\begin{equation}
  \hat{\tau}=
  \begin{cases}
    \Phi_{(k)}(\mathcal{Q}), & k\le m,\\[2pt]
    +\infty, & k>m,
  \end{cases}
  \label{eq:fusion_threshold}
\end{equation}
and the expert-fusion decision is
$\delta_{\mathrm{F}}(v)=\mathbf{1}[\Phi(v)>\hat{\tau}]$.
The strict inequality assigns videos with $\Phi(v)=\hat{\tau}$ to the
authentic class.

\subsection{Marginal False-Alarm Control}
\label{sec:supp_marginal_control}

\begin{proposition}[Split-Conformal False-Alarm Control]
\label{prop:main}
\label{prop:marginal}
Under assumptions~\ref{asm:fixed}--\ref{asm:exch},
\begin{equation}
  \Pr\bigl[\Phi(v)>\hat{\tau}\bigr]
  \le\frac{m+1-k}{m+1}\le\alpha,
  \label{eq:far_bound}
\end{equation}
regardless of dependence among experts. If, conditional on $\mathcal{R}$,
ties among the scores on $\mathcal{Q}\cup\{v\}$ have probability zero, the
first inequality is an equality and
$\Pr[\Phi(v)>\hat{\tau}]\ge\alpha-1/(m+1)$.
\end{proposition}

\begin{proof}
Condition on the index split and $\mathcal{R}$. Let
$Z_1,\dots,Z_m$ be the scores on $\mathcal{Q}$ and let $Z_{m+1}=\Phi(v)$.
Assumptions~\ref{asm:split} and~\ref{asm:exch} preserve exchangeability of the
remaining observations after conditioning on $\mathcal{R}$. Applying the same
fixed function to each observation therefore makes
$Z_1,\dots,Z_{m+1}$ conditionally exchangeable.

Since $0<\alpha<1$, we have $1\le k\le m+1$. If $k=m+1$, the threshold is
$+\infty$ and the false-alarm probability is zero, equal to
$(m+1-k)/(m+1)$. Now suppose $k\le m$. Write $Z_{(k)}$ for the $k$-th smallest
of $Z_1,\dots,Z_m$ and $Z^*_{(k)}$ for the $k$-th smallest of all $m+1$
scores. Adding a score cannot increase the $k$-th order statistic, so
$Z^*_{(k)}\le Z_{(k)}$ and
\[
  \{Z_{m+1}>Z_{(k)}\}\subseteq\{Z_{m+1}>Z^*_{(k)}\}.
\]
At most $m+1-k$ scores strictly exceed $Z^*_{(k)}$. Thus, for
$N=\sum_{i=1}^{m+1}\mathbf{1}[Z_i>Z^*_{(k)}]$, exchangeability gives
\begin{align*}
  \Pr[Z_{m+1}>\hat{\tau}\mid\mathcal{R}]
  &\le\Pr[Z_{m+1}>Z^*_{(k)}\mid\mathcal{R}]\\
  &=\frac{\mathbb{E}[N\mid\mathcal{R}]}{m+1}
  \le\frac{m+1-k}{m+1}.
\end{align*}
Averaging over the reference set and index split gives the first inequality
in Eq.~\eqref{eq:far_bound}.

If ties have probability zero, the test score has a uniform rank among the
$m+1$ scores, and it exceeds $Z_{(k)}$ exactly when its rank exceeds $k$.
Its conditional false-alarm probability is then $(m+1-k)/(m+1)$.
Finally, the definition of $k$ gives
\[
  \alpha-\frac{1}{m+1}
  <\frac{m+1-k}{m+1}\le\alpha,
\]
which proves the remaining claims.
\end{proof}

Each reference p-value takes finitely many values, so the fused statistic can
have ties. Under assumptions~\ref{asm:fixed}--\ref{asm:exch}, the strict
decision rule preserves the upper bound, although ties may make it
conservative. In a seed-0 diagnostic, we pool threshold-set and held-out
authentic scores and round them to nine decimal places before counting ties.
Repeated scores account for at most 0.13\% of this pool on GenVideo and
FaceVid with 1{,}000 reference videos, and 1.6\% on DVF with 150 reference
videos.

The bound concerns authentic videos only. It therefore remains valid under
generator shift if the authentic observations and scoring procedure continue
to satisfy the assumptions; threshold calibration requires no deepfake
examples. Dependence among experts does not invalidate this argument.
In contrast, the classical $\chi^2_{2K}$ reference
of Fisher's method~\cite{fisher1928statistical} is exact for the unweighted
statistic with $K$ independent, continuous, uniform p-values.

\begin{remark}[Dependence among Experts]
\label{rem:dependence}
\normalfont
The experts have mean pairwise Spearman correlations of 0.60 on authentic
FaceVid videos and 0.42 on authentic GenVideo videos, and all reference
p-values share $\mathcal{R}$. In a seed-0 diagnostic averaged over five
reference/threshold splits, the classical $\chi^2_8$ threshold of 15.51 at
$\alpha=5\%$ yields FARs of 10.8\% on FaceVid and 10.9\% on GenVideo.
The calibrated threshold in Eq.~\eqref{eq:fusion_threshold} gives 3.9\% and
4.8\%, respectively. These measurements use held-out authentic videos from
the calibration domain.
\end{remark}

\subsection{Uncertainty for a Single Calibration Set}
\label{sec:supp_calibration_uncertainty}


Proposition~\ref{prop:marginal} averages over calibration sets. For a fixed
split into $\mathcal{R}$ and $\mathcal{Q}$, the population FAR is
\[
  \mathrm{FAR}(\mathcal{R},\mathcal{Q})
  =\Pr[\Phi(v)>\hat{\tau}\mid\mathcal{R},\mathcal{Q}],
\]
where $v$ is an independent authentic draw from $P_0$.
The next result gives a high-probability upper bound for this FAR under
independent sampling. It considers $k\le m$, since $k>m$ gives an infinite
threshold and zero FAR.

\begin{proposition}[Calibration-Conditional False-Alarm Rate]
\label{prop:conditional}
Assume \ref{asm:fixed} and~\ref{asm:split}, let
$\mathcal{C}\cup\{v\}$ consist of independent draws from $P_0$, and let
$k\le m$. Conditional on $\mathcal{R}$,
$\mathrm{FAR}(\mathcal{R},\mathcal{Q})$ is stochastically dominated by
$\mathrm{Beta}(m+1-k,k)$. Equality in distribution holds if the conditional
distribution of $\Phi(v)$ has no atoms. Consequently, for $0<\eta<1$, with
probability at least $1-\eta$ over the calibration set,
$\mathrm{FAR}(\mathcal{R},\mathcal{Q})$ is at most the $(1-\eta)$-quantile
$b_{m,1-\eta}$ of this Beta distribution.
\end{proposition}

\begin{proof}
Condition on $\mathcal{R}$. The scores $Z_1,\dots,Z_m$ on $\mathcal{Q}$ and
$Z_{m+1}=\Phi(v)$ are then i.i.d.\ with distribution function
$H(z)=\Pr[Z_{m+1}\le z\mid\mathcal{R}]$. The strict decision rule gives
\[
  \mathrm{FAR}(\mathcal{R},\mathcal{Q})=1-H(Z_{(k)}).
\]
For $0<u<1$, define $H^{-1}(u)=\inf\{z:H(z)\ge u\}$.
Let $U_1,\dots,U_m$ be i.i.d.\ uniform variables on $(0,1)$ and set
$Y_i=H^{-1}(U_i)$. The variables $Y_i$ have distribution function $H$,
so $1-H(Z_{(k)})$ has the same conditional distribution as
$W=1-H(Y_{(k)})$. Monotonicity of $H^{-1}$ gives
$Y_{(k)}=H^{-1}(U_{(k)})$, while right-continuity of $H$ implies
$H(H^{-1}(u))\ge u$. Hence, under this coupling,
\[
  W=1-H\bigl(H^{-1}(U_{(k)})\bigr)\le1-U_{(k)}.
\]
Since $U_{(k)}\sim\mathrm{Beta}(k,m+1-k)$,
$1-U_{(k)}\sim\mathrm{Beta}(m+1-k,k)$, proving stochastic dominance.
If $H$ is continuous, $H(H^{-1}(u))=u$ and equality in distribution follows.

By the definition of $b_{m,1-\eta}$,
\[
  \Pr\bigl[\mathrm{FAR}(\mathcal{R},\mathcal{Q})
  \le b_{m,1-\eta}\mid\mathcal{R}\bigr]\ge1-\eta.
\]
Averaging over $\mathcal{R}$ proves the final claim.
\end{proof}

This is a special case of training-conditional validity for inductive
conformal predictors~\cite{vovk2012}. At $\alpha=5\%$, the 95\% upper
quantile $b_{m,0.95}$ is approximately 7.7\% for $m=150$ threshold videos and
6.2\% for $m=1{,}000$. The corresponding probabilities that population FAR
exceeds 5\% are at most 0.373 and 0.480, respectively. These bounds describe
variation across independent calibration sets; they do not require every
realized set to attain FAR at most $\alpha$.

\begin{figure*}[t]
\centering
\includegraphics[width=\textwidth]{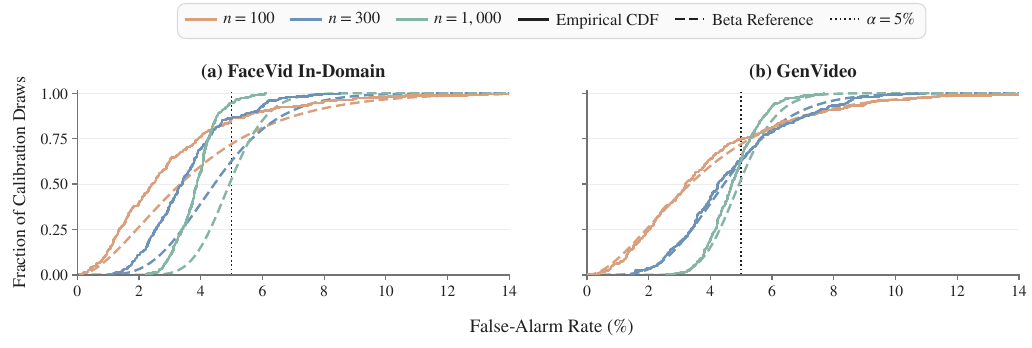}
\caption{\textbf{Variation across calibration draws.}
Empirical cumulative distribution functions (CDFs) of FAR over 300 random
calibration subsets (solid) and the Beta reference from
Proposition~\ref{prop:conditional} (dashed), shown for
(a) FaceVid in-domain and (b) GenVideo.
Each subset contains $n$ authentic videos, split equally into reference
and threshold sets. Dotted lines mark $\alpha=5\%$.
Both panels use seed-0, two-view expert scores.}
\label{fig:theory_beta}
\end{figure*}

Figure.~\ref{fig:theory_beta} compares the Beta reference with empirical
FARs over 300 calibration subsets, each containing $n=2m$ authentic videos
and one reference/threshold split. GenVideo calibration and test videos are disjoint subsets of one pool,
and scoring was selected without its calibration videos. With $n=300$, the
measured FAR has mean 4.67\% and 95th percentile 8.43\%, compared with
4.64\% and 7.74\% for the Beta reference. The means are close, but the
measured upper tail is higher. These empirical FARs use subsets of a finite
pool and a fixed test set, whereas Proposition~\ref{prop:conditional}
concerns independent draws from $P_0$ and population FAR.

FaceVid additionally reuses calibration data during model selection and
investigator-recipe selection (Remark~\ref{rem:departures}), and sampling
without replacement from the fixed FaceVid pool changes the spread of
the measured FARs relative to independent draws.

On DVF, we draw 300 calibration clips from the pool of 2{,}750 authentic
videos and evaluate on the remainder using seed-0 experts at $\alpha=5\%$.
Each draw averages five reference/threshold splits. Across 300 draws, FAR is
$4.65\pm1.32\%$ (mean and standard deviation), and 1.3\% of draws reach or
exceed 8.7\%, the FAR for the main experiment's calibration clips with seed-0
experts. This places that selection near the upper end of the observed
variation and quantifies sensitivity to calibration selection within the
fixed DVF pool.

If scoring is fixed before collecting independent target calibration and
test videos from one authentic distribution, both propositions apply with
$P_0$ equal to that target distribution. Section~\ref{sec:far} evaluates
authentic-target recalibration empirically.

\subsection{Calibration Dependencies in the Implemented System}
\label{sec:supp_calibration_dependencies}

\begin{remark}[Scope for the Implemented System]
\label{rem:departures}
\normalfont
Expert-head fitting excludes $\mathcal{C}$ and linked videos, and calibration
and test clips receive the same fold-averaged scores. Once expert fusion is
fixed, independently collected target calibration and test data can satisfy
\ref{asm:fixed}--\ref{asm:exch}.

During development, source calibration statistics are reused for model and
investigator-recipe selection. The selected configuration therefore need not
be independent of its source calibration set. Investigator training also uses
a calibration-based suspicion band. At inference, prompt suspicion depends
on the threshold subset $\mathcal{Q}$, and evidence-tool percentiles use the
full calibration set $\mathcal{C}$. Calibration clips receive leave-one-out
suspicion scores, whereas test clips use the complete reference and threshold
subsets. Target recalibration retains these dependencies.

The proofs already condition on the reference-dependent score transformation.
The investigator's additional dependence on $\mathcal{Q}$ and the asymmetric
treatment of calibration and test clips do not establish the conditional
exchangeability required by the argument. We therefore assess investigator
and cascade FAR empirically.

FaceVid and ViF calibration clips are held out by source identity and then
split into $\mathcal{R}$ and $\mathcal{Q}$ by video. The 602 ViF calibration
clips come from 402 source identities. These source exclusions separate head
training from calibration, but shared-source clips need not be independent
or satisfy file-level exchangeability. In particular, source holdout alone
does not establish the i.i.d.\ assumption of
Proposition~\ref{prop:conditional}.
\end{remark}

\paragraph{Cascade False Alarms.}
Let $\Phi_2(v)$ denote the re-fused statistic obtained if video $v$ is
investigated, including videos not selected by the gate. For any escalation
set $\mathcal{E}(\mathcal{V})$, Eq.~\eqref{eq:cascade_decision} gives
\[
  \{\delta_{\mathcal{V}}(v)=1\}\subseteq
  \{\Phi(v)>\hat{\tau}\}\cup\{\Phi_2(v)>\hat{\tau}_2\}.
\]
If both branches have marginal FAR at most $\alpha$ under the same authentic
distribution before selection, the union bound gives cascade FAR at most
$2\alpha$, without requiring independence between the branches or the gate.
This does not give an $\alpha$ bound for the cascade or a conditional bound
within the selected subset. The implemented investigator has the calibration
dependencies described in Remark~\ref{rem:departures}, so we evaluate the
cascade empirically using target calibration videos excluded from parameter
fitting and recipe selection.

With 2{,}000 authentic GenVideo calibration videos, cascade FAR is
$4.77\pm0.19\%$ and fusion-only FAR is $4.75\pm0.16\%$ across ten training
seeds; investigation adds 0.40 false-alarm points and removes 0.38. With 300
held-out authentic DVF videos, the corresponding FARs are $7.87\pm1.21\%$
and $8.36\pm0.96\%$; investigation adds 0.99 points and removes 1.48.
Both DVF rates exceed the nominal 5\%. The resampling analysis in
Appendix~\ref{sec:supp_calibration_uncertainty} measures the fusion-only
rule's sensitivity to calibration selection.

\subsection{Calibration with Previously Unused Videos}
\label{sec:supp_fresh_calibration}

We evaluate expert-fusion calibration on authentic files unused during
development. We first freeze the ten ViF-trained expert configurations,
including their heads, two-view scoring, and Fisher fusion. We then screen
previously unused files from the official GenBuster training pool, excluding
files already used in the project and duplicates identified from file or
decoded-frame hashes. After screening and before feature extraction, the
4{,}000 retained files are uniformly assigned to 2{,}000 calibration and
2{,}000 evaluation files. Screening and assignment do not use detector
scores. All 4{,}000 files have complete expert scores.

The primary estimate averages five prespecified reference/threshold splits
within each training seed, then reports the mean and sample standard
deviation across ten seeds. We also evaluate 100 prespecified splits of the
same calibration pool. Both analyses retain the same calibration/evaluation
partition. The 95\% confidence intervals use 2{,}000 bootstrap replicates
that resample evaluation files with fitted models and calibration splits
fixed. Source identities are unavailable, so files are the resampling unit
(See Table~\ref{tab:supp_fresh_calibration}).

\begin{table}[t]
\centering
\caption{\textbf{FAR on previously unused authentic videos (\%).} Primary results average five calibration splits per training seed; the final column averages 100 prespecified splits of the same pool. Means and sample standard deviations are computed across ten seeds. Confidence intervals resample evaluation files with fitted models and calibration splits fixed.}
\label{tab:supp_fresh_calibration}
\resizebox{\linewidth}{!}{%
\begin{tabular}{crrr}
\toprule
\textbf{Nominal $\alpha$} & \textbf{Primary FAR} & \textbf{95\% CI} & \textbf{100-Split FAR} \\
\midrule
1 & $1.24\pm0.29$ & $[0.93,1.56]$ & $1.40\pm0.22$ \\
5 & $5.57\pm0.54$ & $[4.75,6.38]$ & $6.07\pm0.52$ \\
10 & $10.72\pm0.64$ & $[9.58,11.81]$ & $11.38\pm0.55$ \\
\bottomrule
\end{tabular}}
\end{table}

With scoring fixed, uniform assignment makes the threshold videos and a
randomly selected evaluation file exchangeable conditional on the screened
pool and reference subset. The rank argument in
Proposition~\ref{prop:marginal} therefore controls FAR averaged over random
assignments within this pool. The FAR of a realized assignment may exceed
$\alpha$; the population Beta model in Proposition~\ref{prop:conditional}
requires the stronger i.i.d.\ assumptions.

\section{Baseline Implementation and Comparison Protocols}
\label{sec:supp_baselines}

\subsection{Evaluation Settings and Failure Handling}
\label{sec:supp_baseline_protocol}

\paragraph{Training Access and Inputs.}
Table~\ref{tab:supp_protocols} specifies the checkpoint or training source, video input, and decision rule for each of the 32 baseline configurations in Table~\ref{tab:main}. We use released implementations in pinned software environments, with released checkpoints or benchmark-specific retraining as indicated. Evaluation manifests, labels, and metrics are shared, while frame sampling, preprocessing, and computational budgets follow the implementations listed below. In particular, DeMamba, STIL, X-CLIP, and video-based NPR use the DeMamba codebase. Overall aggregates require all seven main-table splits. Gemini-3.8-Flash appears in supplementary comparisons; D3 is evaluated separately using its released average precision (AP) metric.

\begin{table*}[!t]
\centering
\caption{\textbf{Baseline implementations and evaluation settings.} Each row corresponds to one main-table baseline configuration. Image detectors average frame probabilities; video detectors retain their implementation-specific aggregation and thresholds. Per-benchmark training and transfer are described below. Table~\ref{tab:supp_api_settings} specifies hosted-model access and request settings.}
\label{tab:supp_protocols}
\footnotesize
\renewcommand{\arraystretch}{1.15}
\setlength{\tabcolsep}{5pt}
\setlength{\lightrulewidth}{0.3pt}
\resizebox{\textwidth}{!}{%
\begin{tabular}{llll}
\toprule
\textbf{Method} & \textbf{Checkpoint and Training} & \textbf{Evaluation Input} & \textbf{Decision Rule} \\
\midrule
\rowcolor[HTML]{EDF3EF}
\multicolumn{4}{c}{\textbf{Image Detectors}} \\
\midrule
AIDE~\cite{aide} & Released GenImage SD1.4 Checkpoint & 8 Uniform Frames; Released Transforms & Mean Frame Probability $\ge0.5$ \\
DRCT~\cite{drct} & Released DRCT-2M ConvNeXt-B & 8 Uniform Frames; $224^2$ Crops & Mean Frame Probability $\ge0.5$ \\
NPR (Image)~\cite{npr} & Released ProGAN-4class Checkpoint & 8 Uniform Frames; Resize to 256 & Mean Frame Probability $\ge0.5$ \\
SAFE~\cite{safe} & Released Image Checkpoint & 8 Uniform Frames; $256^2$ Crops & Mean Frame Probability $\ge0.5$ \\
UnivFD~\cite{univfd} & Released CLIP ViT-L/14 Classifier & 8 Uniform Frames; $224^2$ Center Crops & Mean Frame Probability $\ge0.5$ \\
\midrule
\rowcolor[HTML]{EDF3EF}
\multicolumn{4}{c}{\textbf{Face Specialists}} \\
\midrule
DFD-FCG~\cite{dfdfcg} & Re-trained on FaceVid & Tracked Faces; 10 Frames per 3-s Clip & Mean Clip Probability $>0.5$ \\
Effort~\cite{effort} & Re-trained on FaceVid & First 8 Aligned Face Crops; $224^2$ & Mean Frame Probability $\ge0.5$ \\
TALL~\cite{tall} & Re-trained on FaceVid & 8 Four-Frame Clips; $2\times2$ Layout & Argmax of Mean Clip Logits \\
\midrule
\rowcolor[HTML]{EDF3EF}
\multicolumn{4}{c}{\textbf{Video Detectors}} \\
\midrule
CoCoDetect~\cite{cocovideo} & Released CoCoVideo R3D-18 & 32 Frames from Index 29; $224^2$ & Video Probability $\ge0.5$ \\
DeMamba~\cite{demamba} & Re-trained per Benchmark & 8-Frame DeMamba Pipeline; $224^2$ & Video Probability $>0.5$ \\
MM-Det~\cite{mmdet} & Released DVF Checkpoint & Reconstructions; 10-Frame Windows & Mean Window Probability $\ge0.5$ \\
NPR (Video)~\cite{npr} & Re-trained per Benchmark & 8-Frame DeMamba Pipeline; $224^2$ & Video Probability $>0.5$ \\
NSG-VD~\cite{nsgvd} & Released GenVideo SEINE Checkpoint & 8 Frames; 100 Authentic Calibration Clips & MMD Statistic $>1$ \\
ReStraV~\cite{restrav} & Re-trained on ViF, GenBuster and CoCoVideo & 24 Frames from Central 2\,s; DINOv2 & Real Score Below Fitted Threshold \\
STIL~\cite{stil} & Re-trained per Benchmark & 8-Frame DeMamba Pipeline; $224^2$ & Video Probability $>0.5$ \\
X-CLIP~\cite{xclip} & Re-trained per Benchmark & 8-Frame DeMamba Pipeline; $224^2$ & Video Probability $>0.5$ \\
\midrule
\rowcolor[HTML]{EDF3EF}
\multicolumn{4}{c}{\textbf{General MLLMs}} \\
\midrule
DeepSeek-V4.1-Flash & API; Zero-Shot & Up to 16 Frames; ARGUS Single-Agent Prompt & ARGUS Answer Parser \\
GPT-6.1 Sol & API; Zero-Shot & Up to 16 Frames; ARGUS Single-Agent Prompt & ARGUS Answer Parser \\
InternVL3.5-8B~\cite{internvl35} & Released Checkpoint; Zero-Shot & 16 Frames; ViF/Skyra Front End & Released Parser; Greedy \\
Kimi K3.1 Pro & API; Zero-Shot & Up to 16 Frames; ARGUS Single-Agent Prompt & ARGUS Answer Parser \\
Qwen2.5-VL-7B~\cite{qwen25vl} & Released Instruct Checkpoint; Zero-Shot & 16 Frames; ViF/Skyra Front End & Released Parser; Greedy \\
Qwen3-VL-8B~\cite{qwen3vl} & Released Instruct Checkpoint; Zero-Shot & 16 Frames; ViF/Skyra Front End & Released Parser; Greedy \\
Qwen3-VL-Plus & API; Zero-Shot & Up to 16 Frames; ARGUS Single-Agent Prompt & ARGUS Answer Parser \\
Qwen3.8-27B & API; Zero-Shot & Up to 16 Frames; ARGUS Single-Agent Prompt & ARGUS Answer Parser \\
Qwen3.8-Flash & API; Zero-Shot & Up to 16 Frames; ARGUS Single-Agent Prompt & ARGUS Answer Parser \\
Qwen3.8-Max & API; Zero-Shot & Up to 16 Frames; ARGUS Single-Agent Prompt & ARGUS Answer Parser \\
Qwen3.8-Omni-Flash & API; Zero-Shot & Up to 16 Frames; ARGUS Single-Agent Prompt & ARGUS Answer Parser \\
\midrule
\rowcolor[HTML]{EDF3EF}
\multicolumn{4}{c}{\textbf{Specialized MLLMs}} \\
\midrule
BusterX++~\cite{busterx} & Released GenBuster Checkpoint & 2\,fps; 64 Image Tokens per Frame & Released Judge; Temperature 0 \\
Skyra~\cite{skyra} & Released Skyra-RL; ViF-CoT-4K & 16 Frames; Short Side 256 & Released Real/Fake Parser; Greedy \\
TUE-Detector~\cite{tue} & Released Stage-1 Checkpoint & Skyra Front End; Released Tool Card & First-Round Verdict; Greedy \\
VideoVeritas~\cite{videoveritas} & Released Checkpoint & Video Input at 3\,fps & Released Parser; Temperature 0.7 \\
\midrule
\rowcolor[HTML]{EDF3EF}
\multicolumn{4}{c}{\textbf{Detection Agents}} \\
\midrule
ARGUS~\cite{argus} & Released Observer/Judge LoRAs & Up to 16 Frames; Full Multi-Agent Pipeline & Released Scorer; Greedy \\
\bottomrule
\end{tabular}}
\end{table*}

\paragraph{Benchmark-Specific Retraining.}
Effort, TALL and DFD-FCG are fitted on the FaceVid training split. The DeMamba-codebase models are fitted separately on FaceVid, ViF, GenBuster and CoCoVideo; ReStraV uses the latter three datasets. Their ViF-trained checkpoints are reused for GenVideo and DVF. The DeMamba-codebase runs use 30 epochs, AdamW with learning rate $10^{-6}$, an effective batch size of 128, and learning-rate milestones at epochs 20 and 25. Checkpoint selection uses a label-stratified 10\% holdout of the training split with seed 0, or CoCoVideo's official validation split. Released checkpoints for the remaining methods are evaluated without additional detector training.

\paragraph{Input Construction and Score Aggregation.}
The image baselines receive eight uniformly sampled frames and apply their released image transforms before averaging fake probabilities. The DeMamba pipeline instead extracts frames at its released rate and selects eight consecutive frames, using the release's center crop and resize. CoCoDetect uses unnormalized inputs; it starts at frame 0 when fewer than 30 frames are available and repeats the last frame if needed. NSG-VD retains its own maximum mean discrepancy (MMD) test and 100-video authentic calibration procedure. Its 100 reference videos are sampled from the held-out authentic calibration data using seed 0.

\paragraph{Verdict Coverage.}
Coverage is the fraction of evaluation videos for which a pipeline returns a usable verdict. Missing outputs can arise from failed face extraction, short clips, unsupported image sizes, or unavailable reconstructions. We distinguish missing outputs from implemented fallback predictions: DFD-FCG assigns probability 0.5 to some clips without a tracked face, which its strict $>0.5$ rule classifies as authentic. The main table follows each pipeline's scoring convention and marks scores that exclude videos without a verdict with a star. Table~\ref{tab:supp_strict} instead counts those missing verdicts as errors.

Unparsable API and ARGUS answers count as errors, whereas VideoVeritas excludes them under its released convention. ViF counts failures within the matched pairs defined by its released pairing rule. MM-Det's released DVF scores cover 3{,}695 videos after its training and validation exclusions; Table~\ref{tab:supp_dvfclean} provides comparisons on this shared subset. Raw model outputs and missing-output records are retained for checking coverage and failure handling.

Main-table boldface and underlining require at least 90\% verdict coverage on the relevant split, and on every main-table split for Overall Avg and Min. This requirement determines which results are highlighted; starred scores are still computed on the videos retained by each pipeline.

\begin{table*}[!t]
\centering
\caption{\textbf{Results across six benchmarks with pipeline failures counted as errors.} Evaluation splits, metrics, and training labels follow Table~\ref{tab:main}. Shading marks access to the benchmark's training split. Bold and underline indicate the best and second-best results with at least 90\% verdict coverage. Our results report the mean $\pm$ standard deviation across ten training seeds.}
\label{tab:supp_strict}
\footnotesize
\setlength{\tabcolsep}{3.0pt}
\renewcommand{\arraystretch}{1.10}
\setlength{\lightrulewidth}{0.3pt}
\resizebox{\textwidth}{!}{%
\begin{tabular}{cllcccccccc}
\toprule
& & & \multicolumn{2}{c}{\textbf{FaceVid OOD}} & \textbf{ViF} & \multicolumn{2}{c}{\textbf{GenBuster}} & \textbf{CoCo} & \textbf{GenVideo} & \textbf{DVF} \\
\cmidrule(lr){4-5}\cmidrule(lr){6-6}\cmidrule(lr){7-8}\cmidrule(lr){9-9}\cmidrule(lr){10-10}\cmidrule(lr){11-11}
\textbf{Type} & \textbf{Method} & \textbf{Train} & \textbf{BAcc} & \textbf{F1} & \textbf{Acc} & \textbf{Test} & \textbf{Closed} & \textbf{Acc} & \textbf{BAcc} & \textbf{BAcc} \\
\midrule
\multirow{5}{*}{\textcolor{tableapricot!80!black}{\shortstack[c]{\textbf{Image}\\\textbf{Detector}}}} & AIDE~\cite{aide} & Img & 52.71 & 19.19 & 51.11 & 63.55 & 61.30 & 53.54 & 66.43 & 60.07 \\
 & DRCT~\cite{drct} & Img & 50.81 & 37.16 & 61.38 & 68.60 & 66.55 & 60.67 & 87.71 & \textbf{82.47} \\
 & NPR~\cite{npr} & Img & 49.97 & 0.00 & 50.44 & 73.65 & 70.00 & 50.85 & 66.16 & 54.28 \\
 & SAFE~\cite{safe} & Img & 49.57 & 0.40 & 50.03 & 56.90 & 50.95 & 49.95 & 50.20 & 50.18 \\
 & UnivFD~\cite{univfd} & Img & 44.03 & 2.90 & 50.16 & 61.35 & 49.55 & 50.13 & 55.72 & 55.43 \\
\midrule
\multirow{3}{*}{\textcolor{tablerose!80!black}{\shortstack[c]{\textbf{Face}\\\textbf{Specialist}}}} & DFD-FCG~\cite{dfdfcg} & FV & \cellcolor{tableslate!20}44.82 & \cellcolor{tableslate!20}9.44 & 52.15 & 65.05 & 65.15 & 49.87 & 52.77 & 50.73 \\
 & Effort~\cite{effort} & FV & \cellcolor{tableslate!20}52.14 & \cellcolor{tableslate!20}29.69 & 23.92 & 49.40 & 45.10 & 33.92 & 29.65 & 24.70 \\
 & TALL~\cite{tall} & FV & \cellcolor{tableslate!20}45.31 & \cellcolor{tableslate!20}23.20 & 25.66 & 49.45 & 47.05 & 34.74 & 20.53 & 26.74 \\
\midrule
\multirow{8}{*}{\textcolor{tablepurple!80!black}{\shortstack[c]{\textbf{Video}\\\textbf{Detector}}}} & CoCoDetect~\cite{cocovideo} & CoCo & 52.41 & 39.87 & 50.58 & 45.20 & 53.35 & \cellcolor{tableslate!20}87.21 & 35.36 & 69.94 \\
 & DeMamba~\cite{demamba} & Each & \cellcolor{tableslate!20}60.52 & \cellcolor{tableslate!20}37.49 & \cellcolor{tableslate!20}73.63 & \cellcolor{tableslate!20}\underline{95.55} & \cellcolor{tableslate!20}88.40 & \cellcolor{tableslate!20}94.33 & 72.38 & 68.00 \\
 & MM-Det~\cite{mmdet} & DVF & 45.68 & 37.36 & 48.85 & 51.85 & 50.00 & 49.10 & 45.10 & \cellcolor{tableslate!20}39.00 \\
 & NPR~\cite{npr} & Each & \cellcolor{tableslate!20}64.08 & \cellcolor{tableslate!20}46.99 & \cellcolor{tableslate!20}58.86 & \cellcolor{tableslate!20}57.55 & \cellcolor{tableslate!20}61.90 & \cellcolor{tableslate!20}56.36 & 62.85 & 42.55 \\
 & NSG-VD~\cite{nsgvd} & GenV & 39.71 & 3.43 & 52.56 & 58.70 & 52.95 & 49.15 & \cellcolor{tableslate!20}\textbf{95.19} & 68.40 \\
 & ReStraV~\cite{restrav} & Each & -- & -- & \cellcolor{tableslate!20}52.53 & \cellcolor{tableslate!20}81.90 & \cellcolor{tableslate!20}57.25 & \cellcolor{tableslate!20}65.97 & 62.51 & 37.04 \\
 & STIL~\cite{stil} & Each & \cellcolor{tableslate!20}\underline{72.01} & \cellcolor{tableslate!20}56.95 & \cellcolor{tableslate!20}69.94 & \cellcolor{tableslate!20}82.00 & \cellcolor{tableslate!20}86.90 & \cellcolor{tableslate!20}67.95 & 74.20 & 64.68 \\
 & X-CLIP~\cite{xclip} & Each & \cellcolor{tableslate!20}62.25 & \cellcolor{tableslate!20}41.39 & \cellcolor{tableslate!20}78.73 & \cellcolor{tableslate!20}\textbf{96.95} & \cellcolor{tableslate!20}89.60 & \cellcolor{tableslate!20}94.33 & 67.83 & 64.45 \\
\midrule
\multirow{11}{*}{\textcolor{tableperiwinkleblue!80!black}{\shortstack[c]{\textbf{MLLM}}}} & DeepSeek-V4.1-Flash~\cite{xu2026deepseek} & -- & 53.15 & 37.90 & 62.68 & 77.50 & 81.75 & 54.72 & 76.41 & 69.52 \\
 & GPT-6.1 Sol~\cite{gpt61sol} & -- & 54.35 & 34.08 & 56.76 & 86.20 & 91.25 & 53.72 & 84.71 & 70.06 \\
 & InternVL3.5-8B~\cite{internvl35} & -- & 43.67 & 22.57 & 52.21 & 60.25 & 58.20 & 50.21 & 48.11 & 58.08 \\
 & Kimi K3.1 Pro~\cite{kimik31pro} & -- & 54.02 & 22.98 & 55.01 & 76.35 & 78.00 & 51.23 & 79.93 & 60.27 \\
 & Qwen2.5-VL-7B~\cite{qwen25vl} & -- & 40.31 & 9.75 & 49.86 & 51.40 & 51.10 & 50.36 & 44.28 & 48.42 \\
 & Qwen3-VL-8B~\cite{qwen3vl} & -- & 48.86 & 3.71 & 49.99 & 51.25 & 56.00 & 50.44 & 54.97 & 50.61 \\
 & Qwen3-VL-Plus~\cite{qwen3vl} & -- & 45.79 & 24.77 & 55.69 & 82.35 & 82.75 & 49.72 & 78.20 & 68.25 \\
 & Qwen3.8-27B~\cite{qwen3827b} & -- & 59.78 & 38.94 & 61.21 & 79.75 & 81.35 & 55.64 & 82.57 & 71.05 \\
 & Qwen3.8-Flash~\cite{qwen38flash} & -- & 58.46 & 37.83 & 63.30 & 79.35 & 80.15 & 55.85 & 86.38 & 73.00 \\
 & Qwen3.8-Max~\cite{qwen38max} & -- & 59.43 & 37.76 & 62.17 & 93.20 & 93.05 & 56.59 & 92.98 & 77.24 \\
 & Qwen3.8-Omni-Flash~\cite{qwen38omniflash} & -- & 61.06 & 42.63 & 69.07 & 73.45 & 74.80 & 58.46 & 88.03 & 75.28 \\
\midrule
\multirow{4}{*}{\textcolor{tableblue!80!black}{\shortstack[c]{\textbf{Specialized}\\\textbf{MLLM}}}} & BusterX++~\cite{busterx} & GB & 46.79 & 25.60 & 53.99 & \cellcolor{tableslate!20}90.05 & \cellcolor{tableslate!20}93.30 & 46.92 & 81.03 & 62.15 \\
 & Skyra~\cite{skyra} & ViF & 61.58 & 40.12 & \cellcolor{tableslate!20}\textbf{91.43} & 50.40 & 50.45 & 50.69 & 65.78 & 55.33 \\
 & TUE-Detector~\cite{tue} & ViF & 51.85 & 27.70 & \cellcolor{tableslate!20}56.89 & 34.45 & 34.45 & 36.79 & 69.26 & 59.93 \\
 & VideoVeritas~\cite{videoveritas} & GB+ & 55.37 & 42.26 & 69.88 & \cellcolor{tableslate!20}92.75 & \cellcolor{tableslate!20}92.60 & 57.36 & 81.74 & 80.59 \\
\midrule
\multirow{1}{*}{\textcolor{tablecyanblue!80!black}{\shortstack[c]{\textbf{Agent}}}} & ARGUS~\cite{argus} & FV & \cellcolor{tableslate!20}68.55 & \cellcolor{tableslate!20}52.31 & 53.90 & 84.60 & 89.95 & 48.69 & 82.14 & 67.74 \\
\midrule
\multirow{2}{*}{\textcolor{tablegreen!80!black}{\shortstack[c]{\textbf{Ours}}}} & \method & Each & \cellcolor{tableslate!20}\textbf{72.16}{\fontsize{6}{6.5}\selectfont$\pm$0.64} & \cellcolor{tableslate!20}\textbf{59.47}{\fontsize{6}{6.5}\selectfont$\pm$1.03} & \cellcolor{tableslate!20}\underline{90.91}{\fontsize{6}{6.5}\selectfont$\pm$0.41} & \cellcolor{tableslate!20}95.28{\fontsize{6}{6.5}\selectfont$\pm$0.29} & \cellcolor{tableslate!20}\textbf{95.42}{\fontsize{6}{6.5}\selectfont$\pm$0.31} & \cellcolor{tableslate!20}\underline{97.39}{\fontsize{6}{6.5}\selectfont$\pm$0.12} & 92.33{\fontsize{6}{6.5}\selectfont$\pm$0.68} & 81.64{\fontsize{6}{6.5}\selectfont$\pm$1.45} \\
 & \textcolor{methodcolor}{\textsc{\textbf{Antiphony-F}}} & Each & \cellcolor{tableslate!20}71.77{\fontsize{6}{6.5}\selectfont$\pm$0.22} & \cellcolor{tableslate!20}\underline{58.85}{\fontsize{6}{6.5}\selectfont$\pm$0.35} & \cellcolor{tableslate!20}90.62{\fontsize{6}{6.5}\selectfont$\pm$0.13} & \cellcolor{tableslate!20}95.19{\fontsize{6}{6.5}\selectfont$\pm$0.30} & \cellcolor{tableslate!20}\underline{95.37}{\fontsize{6}{6.5}\selectfont$\pm$0.28} & \cellcolor{tableslate!20}\textbf{97.48}{\fontsize{6}{6.5}\selectfont$\pm$0.06} & \underline{93.09}{\fontsize{6}{6.5}\selectfont$\pm$0.25} & \underline{82.31}{\fontsize{6}{6.5}\selectfont$\pm$1.28} \\
\bottomrule
\end{tabular}}
\end{table*}

\subsection{Model Implementations and API Settings}
\label{sec:supp_baseline_implementations}

\paragraph{Reproduction Checks.}
We check the reproduced pipelines against published results and released outputs. Skyra reaches 91.43 paired accuracy on ViF-Bench, compared with the reported 91.02. NSG-VD matches the released per-video accuracy of 94.00 and obtains 98.41 AUROC, compared with 98.38. MM-Det reproduces the released reconstructions pixel for pixel through its zero-shot pipeline. ARGUS agrees with its released OOD predictions on 90.0\% of 300 checked videos.

\paragraph{Compatibility Adjustments.}
\label{sec:supp_tue}
The ViF-Bench frame release renames 34 files that its official loader would skip, and VideoVeritas omits a dataset factory imported by its inference script. We repair these file-loading issues. CoCoDetect's released checkpoint uses 32 frames without input normalization, which differs from the training-script defaults. Matching the checkpoint configuration gives 87.64 validation accuracy, close to the stored 87.26.

For TUE-Detector, the public release includes training loops for text-only trajectories, a 41-entry tool interface, twelve initial analyzers implemented as stubs that return raw measurements, and the merged post-SFT checkpoint (``Stage 1'', reported at 93.12 paired accuracy). The analyzer feature extractors, output text, evolved tools, and verdict-update logic are unavailable, preventing reproduction of the complete tool-using detector. We selected Skyra's prompt format on 80 videos from the checkpoint's training set, comprising 40 authentic and 40 deepfake clips, without using test videos. Evaluation uses Skyra's official front end with 16 frames and a short side of 256, TUE's released tool card, and its three-round protocol with greedy decoding. Table~\ref{tab:main} reports the first-round verdict before any tool output, which reaches 56.89 paired accuracy on ViF-Bench, compared with 50.73 for the third-round verdict after stub outputs. Skyra-RL, the starting model identified by the checkpoint's provenance file, reaches 91.43 in the same front end.

\paragraph{Training and Metric Compatibility.}
F3Net is excluded from the DeMamba-based comparison because its released model requires $299\times299$ inputs while the pipeline supplies $224\times224$. At the required resolution, a four-video run uses about 21\,GB, preventing the released batch size of 128 within available memory. For DeMamba, two-step gradient accumulation ($2\times64$) preserves the official effective batch size of 128, although floating-point execution can differ from a single larger batch. D3's released evaluation provides only AP: 57.5 on ViF-Bench, 86.4 and 92.4 on GenBuster, 87.6 on CoCoVideo, 98.5 on GenVideo and 80.4 on DVF. We therefore report these values separately from the metrics in Table~\ref{tab:main}.

\begin{table*}[t]
\centering
\caption{\textbf{API access and inference settings.} Eight models appear in the main comparison; Gemini-3.8-Flash is supplementary. The API service identifies the access route used in our implementation. All token-limit fields are set to 8{,}192. A dash indicates a parameter not explicitly supplied, leaving its behavior to the provider; it does not imply that reasoning is disabled.}
\label{tab:supp_api_settings}
\renewcommand{\arraystretch}{1.15}
\setlength{\tabcolsep}{4pt}
\resizebox{\textwidth}{!}{%
\begin{tabular}{lllcll}
\toprule
\textbf{Model}
& \textbf{Request Model ID}
& \textbf{Token-Limit Field}
& \textbf{Temperature}
& \textbf{Additional Request Settings}
& \textbf{API Service} \\
\midrule
DeepSeek-V4.1-Flash
& \texttt{deepseek-v4.1-flash-1017}
& \texttt{max\_tokens}
& 0
& \begin{tabular}[c]{@{}l@{}}
  \texttt{enable\_thinking=true};\\
  \texttt{reasoning\_effort=low}
  \end{tabular}
& Alibaba Cloud MaaS \\

Gemini-3.8-Flash
& \texttt{gemini-3.8-flash}
& \texttt{max\_tokens}
& 0
& --
& Google Gemini API \\

GPT-6.1 Sol
& \texttt{gpt-6.1-sol}
& \texttt{max\_completion\_tokens}
& --
& --
& OpenAI API \\

Kimi K3.1 Pro
& \texttt{kimi-k3.1-pro}
& \texttt{max\_completion\_tokens}
& 1.0
& \begin{tabular}[c]{@{}l@{}}
  \texttt{top\_p=0.95};\\
  \texttt{presence\_penalty=0}
  \end{tabular}
& Alibaba Cloud MaaS \\

Qwen3-VL-Plus
& \texttt{qwen3-vl-plus}
& \texttt{max\_tokens}
& 0
& --
& Alibaba Cloud MaaS \\

Qwen3.8-27B
& \texttt{qwen3.8-27b}
& \texttt{max\_tokens}
& 0
& --
& Alibaba Cloud MaaS \\

Qwen3.8-Flash
& \texttt{qwen3.8-flash}
& \texttt{max\_tokens}
& 0
& --
& Alibaba Cloud MaaS \\

Qwen3.8-Max
& \texttt{qwen3.8-max}
& \texttt{max\_tokens}
& 0
& --
& Alibaba Cloud MaaS \\

Qwen3.8-Omni-Flash
& \texttt{qwen3.8-omni-flash}
& \texttt{max\_tokens}
& 0
& --
& Alibaba Cloud MaaS \\
\bottomrule
\end{tabular}}
\end{table*}

\paragraph{Locally Evaluated MLLMs.}
Qwen2.5-VL-7B, Qwen3-VL-8B and InternVL3.5-8B use the released ViF/Skyra evaluation front end and system prompt without task-specific fine-tuning. This front end supplies 16 sampled frames with timestamps. The two Qwen wrappers use greedy decoding with an 8{,}192-token limit. InternVL uses $448\times448$ dynamic image tiles, greedy decoding, and a 1{,}024-token limit. Skyra uses its released Skyra-RL checkpoint and maps responses other than ``Real'' to the fake class. TUE-Detector's use of the same front end is described under \emph{Compatibility Adjustments}.

\paragraph{API Access and Model Identifiers.}
Table~\ref{tab:supp_api_settings} lists the eight API baselines in the main table and Gemini-3.8-Flash from the supplementary comparisons. The identifiers are the model strings sent in requests, rather than immutable provider checkpoint hashes. The configured access routes are \nolinkurl{https://api.openai.com/v1} for OpenAI and \nolinkurl{https://generativelanguage.googleapis.com/v1beta/openai/} for Gemini's OpenAI-compatible interface. Qwen, Kimi and DeepSeek use Alibaba Cloud Model-as-a-Service (MaaS), with workspace-specific endpoints of the form \nolinkurl{https://llm-<workspace>.cn-beijing.maas.aliyuncs.com/compatible-mode/v1}; the workspace identifier is omitted for anonymity. Saved request settings and provider aliases document access, but do not establish that hosted weights remain unchanged.

\paragraph{Shared API Inputs.}
All nine hosted models use the released ARGUS single-agent system and user prompts. Each request contains up to 16 frames sampled uniformly by frame index; short clips use all available frames. Frames retain their aspect ratio and are downscaled only when the longer side exceeds 512 pixels, then encoded as RGB JPEG images at quality 90. We submit the images with frame-order labels through the \texttt{image\_url} interface and \texttt{detail=high}. Requests contain no audio or container metadata, and the single-agent baseline does not call tools. The prompt asks for an explanation and a real/fake verdict in \texttt{<explanation>} and \texttt{<answer>} tags. These inputs are shared; each provider's visual tokenization remains internal to its service.

\paragraph{API Inference Settings.}
Table~\ref{tab:supp_api_settings} gives the explicit request settings. All models request an 8{,}192-token output limit, using the field accepted by the corresponding service. This is a request budget, not a common count of generated answer tokens: reasoning tokens may be counted differently. GPT's temperature is omitted; Kimi uses temperature 1.0, top-$p$ 0.95, and presence penalty 0. DeepSeek enables reasoning at low effort. This setting was fixed after compatibility checks found truncated outputs at default effort, without selecting by prediction accuracy. Other models receive no explicit reasoning override. The client reads complete, non-streaming responses.

\paragraph{API Response Handling.}
The ARGUS parser first looks for an answer tag and otherwise uses the last real/fake token in the response; a nonempty explanation is also required. For Qwen3-VL-Plus, Qwen3.8-Omni-Flash, Kimi K3.1 Pro and DeepSeek-V4.1-Flash, the response must additionally contain a nonempty final answer and \texttt{finish\_reason=stop} before parsing. Their truncations, unparsable final answers, and content refusals count as failed verdicts and are scored as errors. Successful predictions are retained while failed records receive at most three passes; unresolved transport, quota, or authentication errors prevent a split from being marked complete. GPT, Gemini, and Qwen3.8-Max/Flash/27B retain the original ARGUS response-reading and parsing procedure, without these additional completion checks.

\subsection{Recall, False Alarms, and Coverage}
\label{sec:supp_detection_behavior}

\paragraph{Recall and False Alarms on Unseen Generators.}
Table~\ref{tab:supp_facevid} separates missed deepfakes from false alarms on FaceVid OOD, which contains 1{,}920 fakes from 20 unseen generators and 5{,}716 authentic videos from partly new capture domains. Face specialists re-trained on FaceVid (Effort~\cite{effort}, TALL~\cite{tall}, DFD-FCG~\cite{dfdfcg}) retain 86--99\% specificity but detect only 10--32\% of unseen fakes. The API baselines span a wider recall range of 16.1\% to 62.4\%, but their most sensitive model, DeepSeek, flags 56.1\% of authentic videos. VideoVeritas~\cite{videoveritas} detects 92.0\% of deepfakes while flagging 77.8\% of authentic videos. Higher recall can therefore come at the cost of substantially more false alarms.

ARGUS~\cite{argus} reaches 68.6\% BAcc at 69.0\% specificity, while STIL~\cite{stil} reaches 72.0\% at 77.3\% specificity. Reporting recall and FAR alongside BAcc makes these operating points explicit. This contrast between missed deepfakes and false alarms resembles the credulity--sensitivity dilemma reported for generated images~\cite{fraudbench}.

\begin{table*}[!t]
\centering
\caption{\textbf{Detailed results on FaceVid OOD.} Results follow each method's official scoring convention; starred scores exclude pipeline failures, as in Table~\ref{tab:main}. Face specialists, DeMamba, STIL, X-CLIP, and NPR (Video) are re-trained on FaceVid; general MLLMs are evaluated zero-shot. AUROC is computed from continuous detector scores. For the fusion-only variant, it uses the expert-fusion statistic and is averaged over ten seeds.}
\label{tab:supp_facevid}
\begin{tabular}{lccccc}
\toprule
\textbf{Method}
& \textbf{BAcc}
& \textbf{F1}
& \textbf{Recall}
& \textbf{Spec.}
& \textbf{AUROC} \\
\midrule
\rowcolor[HTML]{EDF3EF}
\multicolumn{6}{c}{\textbf{Image Detectors}} \\
\midrule
AIDE & 52.71 & 19.19 & 13.02 & 92.39 & 54.74 \\
DRCT & 50.81 & 37.16 & 67.76 & 33.85 & 49.29 \\
NPR (Image) & 49.97 & 0.00 & 0.00 & 99.95 & 27.62 \\
SAFE & 49.57 & 0.40 & 0.21 & 98.93 & 35.41 \\
UnivFD & 44.03 & 2.90 & 2.08 & 85.97 & 34.95 \\
\midrule
\rowcolor[HTML]{EDF3EF}
\multicolumn{6}{c}{\textbf{Face Specialists}} \\
\midrule
DFD-FCG & 54.46$^*$ & 17.90$^*$ & 10.27$^*$ & 98.65$^*$ & 42.74$^*$ \\
Effort & 62.51$^*$ & 43.17$^*$ & 32.32$^*$ & 92.70$^*$ & 73.28$^*$ \\
TALL & 57.79$^*$ & 36.36$^*$ & 29.15$^*$ & 86.43$^*$ & 63.07$^*$ \\
\midrule
\rowcolor[HTML]{EDF3EF}
\multicolumn{6}{c}{\textbf{Video Detectors}} \\
\midrule
CoCoDetect & 52.41 & 39.87 & 82.45 & 22.38 & 47.78 \\
DeMamba & 60.52 & 37.49 & 27.50 & 93.54 & 70.25 \\
MM-Det & 50.00$^*$ & 44.60$^*$ & 100.00$^*$ & 0.00$^*$ & 47.39$^*$ \\
NPR (Video) & 64.08 & 46.99 & 58.02 & 70.14 & 69.58 \\
NSG-VD & 50.38$^*$ & 5.86$^*$ & 3.21$^*$ & 97.54$^*$ & 50.82$^*$ \\
STIL & 72.01 & 56.95 & 66.67 & 77.34 & 77.71 \\
X-CLIP & 62.25 & 41.39 & 31.67 & 92.83 & 75.60 \\
\midrule
\rowcolor[HTML]{EDF3EF}
\multicolumn{6}{c}{\textbf{General MLLMs}} \\
\midrule
DeepSeek-V4.1-Flash & 53.15 & 37.90 & 62.45 & 43.86 & -- \\
Gemini-3.8-Flash & 55.82 & 33.07 & 31.56 & 80.07 & -- \\
GPT-6.1 Sol & 54.35 & 34.08 & 39.17 & 69.54 & -- \\
InternVL3.5-8B & 43.67 & 22.57 & 28.18 & 59.17 & -- \\
Kimi K3.1 Pro & 54.02 & 22.98 & 16.09 & 91.95 & -- \\
Qwen2.5-VL-7B & 40.31 & 9.75 & 9.53 & 71.10 & -- \\
Qwen3-VL-8B & 48.86 & 3.71 & 2.14 & 95.59 & -- \\
Qwen3-VL-Plus & 45.79 & 24.77 & 30.52 & 61.06 & -- \\
Qwen3.8-27B & 59.78 & 38.94 & 36.04 & 83.52 & -- \\
Qwen3.8-Flash & 58.46 & 37.83 & 37.86 & 79.06 & -- \\
Qwen3.8-Max & 59.43 & 37.76 & 33.23 & 85.64 & -- \\
Qwen3.8-Omni-Flash & 61.06 & 42.63 & 47.81 & 74.30 & -- \\
\midrule
\rowcolor[HTML]{EDF3EF}
\multicolumn{6}{c}{\textbf{Specialized MLLMs}} \\
\midrule
BusterX++ & 47.06$^*$ & 25.86$^*$ & 31.04$^*$ & 63.07$^*$ & -- \\
Skyra & 61.58 & 40.12 & 30.83 & 92.32 & -- \\
TUE-Detector & 51.85 & 27.70 & 27.45 & 76.24 & -- \\
VideoVeritas & 57.09$^*$ & 44.58$^*$ & 92.02$^*$ & 22.17$^*$ & -- \\
\midrule
\rowcolor[HTML]{EDF3EF}
\multicolumn{6}{c}{\textbf{Detection Agents}} \\
\midrule
ARGUS & 68.55 & 52.31 & 68.07 & 69.03 & -- \\
\midrule
\rowcolor[HTML]{EDF3EF}
\multicolumn{6}{c}{\textbf{Ours}} \\
\midrule
\method & 72.16$\pm$0.64 & 59.47$\pm$1.03 & 51.95$\pm$1.30 & 92.37$\pm$0.31 & -- \\
\textcolor{methodcolor}{\textsc{\textbf{Antiphony-F}}}
& 71.77$\pm$0.22 & 58.85$\pm$0.35 & 51.18$\pm$0.52 & 92.36$\pm$0.11 & 82.53$\pm$0.25 \\
\bottomrule
\end{tabular}
\end{table*}

\begin{figure}[t]
  \centering
  \includegraphics[width=\linewidth]{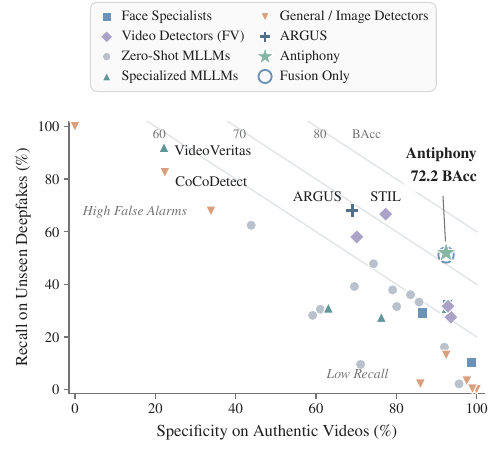}
  \caption{\textbf{The credulity--sensitivity dilemma on FaceVid OOD.} Recall on deepfakes from 20 unseen generators is plotted against specificity on authentic videos. Gray lines indicate constant BAcc; results for \method are averaged over ten seeds.}
  \label{fig:dilemma}
\end{figure}

\paragraph{Complementary Errors and Pairwise Fusion.}
To assess whether different errors can support better combined decisions, we compare Effort and ARGUS on the 6{,}359 FaceVid OOD videos for which both methods return predictions, including 1{,}878 deepfakes. A label-informed oracle that selects a correct verdict whenever either detector is correct reaches 82.61\% BAcc. Simple combinations recover much less of this potential gain. Conjunction, which flags a video only when both detectors predict a deepfake, reaches 61.36\% BAcc; disjunction, which flags it when either detector does, reaches 68.20\%, compared with 67.06\% for ARGUS alone on the same subset.

A gate based on Effort's confidence leaves most of this gap unresolved. It uses Effort when $|p_{\mathrm{fake}}-0.5|\ge t$ and ARGUS otherwise. Even when $t$ is retrospectively selected from nine values between 0.05 and 0.45 using evaluation BAcc, the best margin, $t=0.40$, yields only 67.83\%. Effort assigns scores below 0.1 to 28.1\% of deepfakes in this subset, illustrating why confident predictions can be misleading. Complementary errors thus motivate fusion, but the tested confidence gate does not reliably select between these detectors.

\begin{table}[t]
\centering
\caption{\textbf{DVF results on the clean subset and full evaluation set.} The clean subset excludes MM-Det's training data and contains 3{,}695 videos, including 2{,}138 deepfakes. Both columns report BAcc under each method's official scoring convention. Starred scores exclude videos without a verdict, as in Table~\ref{tab:main}; MM-Det therefore uses the same 3{,}695 videos in both columns.}
\label{tab:supp_dvfclean}
\begin{tabular}{lcc}
\toprule
\textbf{Method} & \textbf{Clean Subset} & \textbf{Full DVF} \\
\midrule
\rowcolor[HTML]{EDF3EF}
\multicolumn{3}{c}{\textbf{Image Detectors}} \\
\midrule
AIDE & 63.32 & 60.07 \\
DRCT & 74.72 & 82.47 \\
NPR (Image) & 58.45 & 54.28 \\
SAFE & 50.12 & 50.18 \\
UnivFD & 55.57 & 55.43 \\
\midrule
\rowcolor[HTML]{EDF3EF}
\multicolumn{3}{c}{\textbf{Video Detectors}} \\
\midrule
CoCoDetect & 59.18 & 69.94 \\
DeMamba & 78.91 & 68.00 \\
MM-Det & 68.64 & 68.64$^*$ \\
NPR (Video) & 42.62 & 42.55 \\
NSG-VD & 69.38 & 68.40 \\
ReStraV & 44.37 & 37.04 \\
STIL & 66.28 & 64.68 \\
X-CLIP & 81.13 & 64.45 \\
\midrule
\rowcolor[HTML]{EDF3EF}
\multicolumn{3}{c}{\textbf{General MLLMs}} \\
\midrule
InternVL3.5-8B & 59.61 & 58.08 \\
Qwen2.5-VL-7B & 47.09 & 48.42 \\
Qwen3-VL-8B & 50.92 & 50.61 \\
\midrule
\rowcolor[HTML]{EDF3EF}
\multicolumn{3}{c}{\textbf{Specialized MLLMs}} \\
\midrule
BusterX++ & 73.55$^*$ & 62.15$^*$ \\
Skyra & 59.75 & 55.33 \\
TUE-Detector & 68.04 & 59.93 \\
TUE-Detector (Turn 3) & 51.25 & 50.38 \\
VideoVeritas & 94.52$^*$ & 80.60$^*$ \\
\midrule
\rowcolor[HTML]{EDF3EF}
\multicolumn{3}{c}{\textbf{Detection Agents}} \\
\midrule
ARGUS & 76.19 & 67.74 \\
\midrule
\rowcolor[HTML]{EDF3EF}
\multicolumn{3}{c}{\textbf{Ours}} \\
\midrule
\method & 85.40{\scriptsize$\pm$1.10} & 81.64 \\
\textcolor{methodcolor}{\textsc{\textbf{Antiphony-F}}}
& 85.98{\scriptsize$\pm$0.95} & 82.31 \\
\bottomrule
\end{tabular}
\end{table}

\paragraph{Matched False-Alarm Rates on FaceVid OOD.}
We next compare recall at a common FAR to separate detection performance from the choice of operating point. STIL, re-trained on FaceVid, reaches 72.01 BAcc compared with $72.16\pm0.64$ for \method; the paired difference is $+0.16$ points (95\% CI $[-1.19, 1.52]$). Their FARs are 22.7\% and 7.6\%, respectively. Adjusting STIL's threshold using authentic videos from the OOD test split to match 7.6\% FAR gives 34.4\% recall on unseen fakes, compared with 52.0\% for \method. At 5\% and 10\% FAR, STIL's recall is 25.1\% and 41.4\%, respectively. Its AUROC is 77.7, compared with $82.5\pm0.3$ for the fusion-only variant. Matching FAR reveals a recall advantage that the similar BAcc values obscure. This threshold adjustment uses authentic test videos as an evaluation diagnostic and is separate from the deployment protocol.

\paragraph{DVF Without MM-Det's Training Data.}
To compare methods on the same videos, we restrict this analysis to the subset covered by MM-Det's release. The full DVF evaluation set contains 1{,}800 SVD fakes and 893 YouTube reals from MM-Det's training and validation data. Its released scores cover the remaining 1{,}557 real and 2{,}138 fake videos, or 57.8\% of the full evaluation set. Table~\ref{tab:supp_dvfclean} compares methods on these 3{,}695 videos. The fusion-only variant reaches $85.98\pm0.95$ BAcc, and \method reaches 85.40, compared with 68.64 for MM-Det. The advantage over MM-Det therefore persists when all three methods are evaluated on the subset covered by its release.
\section{Expert Design and Calibration Analysis}
\label{sec:supp_design}

This appendix examines how expert representations, score combination, and calibration affect detection and false alarms. We compare encoder adaptation and input views, study the contributions of individual experts, and evaluate calibration choices under both the main settings and matched access to authentic target videos.

\subsection{Representation Adaptation and Input Views}
\label{sec:supp_representations}

\paragraph{Frozen Versus Adapted Encoders.}
We compare frozen and Effort-adapted expert configurations~\cite{effort} in Table~\ref{tab:adapt}. The adapted-expert configuration reaches 84.85 mean validation BAcc, compared with 83.46 for the frozen configuration, and performs better on ViF but worse on FaceVid OOD and GenVideo. Higher validation performance therefore does not translate into consistent gains under transfer. We retain frozen encoders in \method; Appendix~\ref{sec:supp_agent} describes their backbones and fitted heads.

\begin{table}[t]
\centering
\caption{\textbf{Frozen and adapted expert configurations.} The rows use four frozen experts, those experts plus Effort-CLIP, and Effort-CLIP with Effort-DINOv3, respectively. Fusion-only results use one uniform 16-frame view and training seed 0. Val. is family-balanced validation BAcc averaged over FaceVid and ViF. Test metrics are ViF paired accuracy, FaceVid OOD BAcc, and GenVideo zero-shot BAcc using ViF-trained experts under the equalized protocol in Appendix~\ref{sec:supp_zeroshot}. All scores are percentages; bold marks the highest value in each column.}
\label{tab:adapt}
\resizebox{\linewidth}{!}{%
\begin{tabular}{lcccc}
\toprule
\textbf{Experts} & \textbf{Val.} & \textbf{ViF} & \textbf{FaceVid OOD} & \textbf{GenVideo} \\
\midrule
Frozen (\method) & 83.46 & 90.26 & \textbf{71.86} & \textbf{96.03} \\
Frozen + Effort-CLIP & 84.19 & 92.29 & 71.70 & 94.87 \\
Effort-CLIP + Effort-DINOv3 & \textbf{84.85} & \textbf{95.08} & 68.26 & 92.99 \\
\bottomrule
\end{tabular}}
\end{table}

\paragraph{Input Views.}
A second temporal view can expose the experts to frames missed by uniform sampling. We compare a single uniform 16-frame view with averaging video logits from uniform and half-interval-offset views. Across ten expert-head seeds, the second view changes validation BAcc by $+0.08\pm0.08$ points. Paired test differences are $+0.24\pm0.08$ on ViF, $-0.16\pm0.08$ on FaceVid OOD, and $-0.92\pm0.50$ on DVF. Additional temporal coverage thus has a modest, dataset-dependent effect. Table~\ref{tab:supp_pergen} breaks down the ViF results by generator. Input-view selection uses the same held-out development clips as investigator selection. The encoder-adaptation comparison instead uses GHCF out-of-fold training deepfakes for validation.

\begin{table*}[t]
\centering
\caption{\textbf{Effects of score fusion and expert selection.} Entries show fusion-only benchmark scores with FAR in parentheses, all in percentages. Results average five calibration splits within each of ten training seeds, then average across seeds. Fusion rules use the same four experts; expert-removal and single-expert variants use Fisher fusion with recalibrated thresholds. The final block uses fixed or source-fitted decision rules without authentic-only calibration.}
\label{tab:supp_ablation}
\resizebox{\textwidth}{!}{%
\begin{tabular}{lccccccc}
\toprule
\textbf{Variant} & \textbf{FaceVid OOD} & \textbf{ViF} & \textbf{GB Test} & \textbf{GB Closed} & \textbf{CoCo} & \textbf{GenVideo} & \textbf{DVF} \\
\midrule
\rowcolor[HTML]{EDF3EF}
\multicolumn{8}{c}{\textbf{Fusion Rules}} \\
\midrule
Fisher (\textcolor{methodcolor}{\textsc{\textbf{Antiphony-F}}}) & 71.77 (7.6) & 90.62 (10.4) & 95.19 (9.6) & 95.37 (8.6) & 97.48 (4.0) & 93.09 (13.0) & 82.31 (8.4) \\
Stouffer & 71.29 (7.1) & 90.30 (10.7) & 95.07 (9.9) & 95.14 (8.9) & 97.36 (4.0) & 92.86 (13.3) & 79.81 (8.2) \\
Mean $p$-Value & 69.33 (6.4) & 88.68 (12.3) & 94.48 (10.5) & 94.51 (9.5) & 94.76 (4.6) & 91.80 (14.3) & 75.59 (7.7) \\
Minimum $p$-Value & 71.89 (9.4) & 88.82 (10.1) & 96.16 (7.7) & 95.86 (7.2) & 97.34 (3.8) & 94.61 (9.8) & 77.98 (5.5) \\
\midrule
\rowcolor[HTML]{EDF3EF}
\multicolumn{8}{c}{\textbf{Expert Removal}} \\
\midrule
Without SigLIP2 & 70.85 (8.3) & 91.03 (8.3) & 95.20 (9.6) & 95.31 (8.4) & 97.23 (4.4) & 91.19 (16.6) & 84.00 (7.9) \\
Without CLIP & 71.67 (8.0) & 89.00 (11.8) & 95.24 (9.5) & 95.07 (8.7) & 96.87 (4.0) & 94.86 (9.5) & 87.49 (8.4) \\
Without DINOv3 & 71.65 (8.1) & 90.14 (10.8) & 95.25 (9.5) & 95.23 (8.6) & 97.08 (4.0) & 93.38 (11.7) & 71.48 (6.7) \\
Without V-JEPA~2 & 71.41 (6.8) & 90.67 (8.8) & 95.49 (9.0) & 95.65 (8.4) & 97.46 (4.3) & 94.17 (10.8) & 76.13 (7.9) \\
\midrule
\rowcolor[HTML]{EDF3EF}
\multicolumn{8}{c}{\textbf{Single Experts}} \\
\midrule
SigLIP2 Only & 70.21 (6.1) & 85.24 (12.4) & 96.62 (6.7) & 95.92 (7.2) & 96.05 (3.5) & 97.46 (1.8) & 58.49 (5.1) \\
CLIP Only & 69.06 (7.5) & 91.42 (4.6) & 96.21 (7.6) & 96.18 (6.7) & 97.60 (3.9) & 89.18 (18.4) & 52.84 (5.4) \\
DINOv3 Only & 67.88 (6.9) & 85.83 (9.2) & 96.67 (6.7) & 95.48 (6.3) & 96.53 (3.1) & 93.37 (11.8) & 90.65 (5.8) \\
V-JEPA~2 Only & 61.84 (10.1) & 79.11 (6.4) & 88.61 (10.4) & 87.37 (8.8) & 78.16 (3.6) & 90.21 (7.1) & 74.12 (5.0) \\
\midrule
\rowcolor[HTML]{EDF3EF}
\multicolumn{8}{c}{\textbf{Decision Rules Without Authentic-Only Calibration}} \\
\midrule
Mean Probability $>0.5$ & 62.88 (2.1) & 90.25 (6.7) & 99.81 (0.0) & 89.62 (0.1) & 98.08 (1.1) & 95.14 (8.3) & 60.52 (78.1) \\
Logistic Stacking & 64.88 (3.1) & 89.64 (14.8) & 99.90 (0.0) & 88.30 (0.1) & 98.07 (2.9) & 87.44 (24.4) & 56.54 (86.8) \\
OOF BAcc-Optimal Threshold & 63.81 (2.5) & 89.54 (15.2) & 99.77 (0.2) & 92.91 (0.3) & 97.52 (4.1) & 89.04 (21.4) & 55.58 (88.7) \\
\bottomrule
\end{tabular}}
\end{table*}

\subsection{Score Fusion and Expert Complementarity}
\label{sec:supp_fusion_ablation}

\paragraph{Compared Decision Rules.}
To isolate the effect of score combination, we compare four rules using the same expert scores and reference/threshold split: Fisher, Stouffer, negative mean $p$-value, and negative minimum $p$-value. Stouffer sums standard-normal inverse survival transforms after clipping $p$-values to $[10^{-12},1-10^{-12}]$. Each statistic receives its own empirical threshold on $\mathcal{Q}$ using Eq.~\eqref{eq:fusion_threshold}. The comparison therefore changes how evidence is combined while keeping threshold calibration; no rule uses an independence-based analytical cutoff.

We also compare three decision rules without authentic-only calibration: mean sigmoid probability thresholded at 0.5, class-balanced logistic stacking of source OOF logits thresholded at 0.5, and a mean-logit threshold selected to maximize source OOF BAcc over 2{,}001 empirical quantiles. Logistic stacking uses $C=1$, LBFGS, and at most 1{,}000 iterations; fitting excludes calibration clips and linked training entries. Finally, we remove one expert at a time or retain a single expert, recomputing reference $p$-values and thresholds for each retained set.

Table~\ref{tab:supp_ablation} reports fusion-only means across ten training seeds, with FAR on authentic test videos in parentheses. Full-pool results first average five calibration splits within each seed. Tables~\ref{tab:supp_alpha} and~\ref{tab:supp_csize} use the same convention, with additional subset sampling for reduced calibration pools.

\paragraph{Combining Expert Evidence.}
Fisher exceeds Stouffer and mean $p$-value fusion on all six benchmarks in Table~\ref{tab:supp_ablation}. Minimum $p$-value fusion performs better on FaceVid, GenBuster, and GenVideo, but worse on ViF and DVF. These differences show that the fusion rule affects transfer even when expert predictions and calibration data are unchanged.

The expert ablations reveal why no single representation is equally effective across domains. CLIP performs well on ViF and CoCoVideo, SigLIP2 on GenVideo, and DINOv3 on DVF. Removing CLIP improves GenVideo and DVF BAcc but lowers ViF accuracy. The experts offer complementary strengths, although including a particular expert does not improve every benchmark.

\paragraph{Decisions Without Authentic-Only Calibration.}
Fixed-threshold and source-fitted rules improve GenBuster Test and CoCoVideo accuracy but reduce FaceVid OOD and DVF BAcc; mean probability also improves GenVideo BAcc. Strong performance on source benchmarks therefore does not establish that the same decision rule transfers well. On DVF, this comparison changes both the rule and access to authentic target calibration videos. Appendix~\ref{sec:supp_equal_calibration} separates these factors by matching calibration access.

\begin{table}[t]
\centering
\caption{\textbf{Held-out versus out-of-fold calibration.} Results use development experts, one uniform view, training seed 0, and $\alpha=5\%$. BAcc and F1 are measured on FaceVid OOD; FAR is measured on authentic in-domain test videos. All scores are percentages. OOF authentic training scores come from individual fold heads; test scores use fold-averaged logits.}
\label{tab:ablation}
\begin{tabular}{lrrr}
\toprule
\textbf{Calibration Design} & \textbf{BAcc} & \textbf{F1} & \textbf{FAR} \\
\midrule
GHCF + Held-Out $\mathcal{C}$ (Ours) & 71.10 & 57.84 & 3.54 \\
In-Fold + Held-Out $\mathcal{C}$ & 70.88 & 57.59 & 3.39 \\
GHCF + OOF Scores & 73.78 & 61.03 & 6.92 \\
\bottomrule
\end{tabular}
\end{table}

\subsection{Calibration Design and Sensitivity}
\label{sec:supp_calib}
\label{sec:supp_ablation}

\begin{table*}[t]
\centering
\caption{\textbf{Matched authentic calibration at $\alpha=5\%$.} BAcc and FAR are percentages. Column headings give total pool sizes; in the second block, all methods set thresholds using Fisher's threshold half. Expert combinations report mean $\pm$ standard deviation across ten training seeds after averaging 20 calibration selections/splits. Standalone detectors use one checkpoint each. All methods have complete calibration and evaluation coverage. Investigation is disabled; these settings differ from the main evaluation.}
\label{tab:supp_equal_calibration}
\begin{tabular}{lcccccc}
\toprule
& \multicolumn{2}{c}{\textbf{GenVideo, 300 Clips}} & \multicolumn{2}{c}{\textbf{GenVideo, 2{,}000 Clips}} & \multicolumn{2}{c}{\textbf{DVF, 300 Clips}} \\
\cmidrule(lr){2-3}\cmidrule(lr){4-5}\cmidrule(lr){6-7}
\textbf{Method} & \textbf{BAcc} & \textbf{FAR} & \textbf{BAcc} & \textbf{FAR} & \textbf{BAcc} & \textbf{FAR} \\
\midrule
\rowcolor[HTML]{EDF3EF}
\multicolumn{7}{c}{\textbf{Same Total Calibration Budget}} \\
\midrule
Fisher (\textcolor{methodcolor}{\textsc{\textbf{Antiphony-F}}}) & 97.04{\scriptsize$\pm$0.05} & 4.17 & 97.05{\scriptsize$\pm$0.07} & 4.62 & 81.04{\scriptsize$\pm$1.36} & 7.03 \\
Mean Probability & 96.46{\scriptsize$\pm$0.10} & 4.69 & 96.57{\scriptsize$\pm$0.11} & 4.40 & 75.25{\scriptsize$\pm$0.25} & 4.61 \\
Source-Trained Logistic Stacking & 95.63{\scriptsize$\pm$0.16} & 5.27 & 95.68{\scriptsize$\pm$0.19} & 5.14 & 78.27{\scriptsize$\pm$0.96} & 4.81 \\
STIL & 70.23 & 5.86 & 69.86 & 5.28 & 60.44 & 6.08 \\
X-CLIP & 71.26 & 5.63 & 71.60 & 5.84 & 64.65 & 5.84 \\
DRCT & 90.23 & 4.64 & 90.30 & 4.74 & 84.77 & 5.18 \\
\midrule
\rowcolor[HTML]{EDF3EF}
\multicolumn{7}{c}{\textbf{Same Threshold Subset}} \\
\midrule
Fisher (\textcolor{methodcolor}{\textsc{\textbf{Antiphony-F}}}) & 97.04{\scriptsize$\pm$0.05} & 4.17 & 97.05{\scriptsize$\pm$0.07} & 4.62 & 81.04{\scriptsize$\pm$1.36} & 7.03 \\
Mean Probability & 96.60{\scriptsize$\pm$0.09} & 4.24 & 96.62{\scriptsize$\pm$0.10} & 4.25 & 75.08{\scriptsize$\pm$0.24} & 4.41 \\
Source-Trained Logistic Stacking & 95.70{\scriptsize$\pm$0.15} & 4.61 & 95.74{\scriptsize$\pm$0.17} & 4.86 & 78.16{\scriptsize$\pm$0.82} & 4.96 \\
STIL & 69.66 & 5.71 & 69.73 & 5.34 & 59.94 & 5.51 \\
X-CLIP & 70.69 & 5.31 & 71.51 & 5.70 & 64.71 & 5.54 \\
DRCT & 89.90 & 4.38 & 90.26 & 4.65 & 84.11 & 4.82 \\
\bottomrule
\end{tabular}
\end{table*}

\paragraph{Authentic-Score Diagnostic.}
The single-expert analysis in Section~\ref{sec:analysis} tests how a fixed threshold behaves when authentic and deepfake video distributions change. We use four FaceVid-trained experts with training seed 0, averaging video logits from uniform and half-interval-offset views of 16 frames each. Each threshold is the empirical 95th percentile of scores on all 2{,}001 held-out authentic calibration videos. We apply it unchanged to in-domain and OOD test clips, flagging scores that strictly exceed the threshold. The reported FAR and recall ranges span the four experts. Figure~\ref{fig:motivation} shows the corresponding fused-score behavior using the same experts, reference/threshold split seed 0, and $\alpha=5\%$.

\begin{table}[t]
\centering
\caption{\textbf{Sensitivity to the nominal false-alarm level.} Entries show fusion-only benchmark scores with FAR in parentheses, all in percentages. Results are averaged over five calibration splits within each of ten training seeds, then averaged across seeds. GenVideo retains source calibration; DVF uses authentic target videos.}
\label{tab:supp_alpha}
\resizebox{\linewidth}{!}{%
\begin{tabular}{lcccc}
\toprule
$\alpha$ & \textbf{FaceVid OOD} & \textbf{ViF} & \textbf{GenVideo} & \textbf{DVF} \\
\midrule
1\% & 64.30 (3.0) & 88.97 (2.7) & 97.14 (3.2) & 67.16 (1.0) \\
2\% & 67.82 (4.4) & 90.18 (5.1) & 96.48 (5.4) & 76.26 (3.4) \\
5\% & 71.77 (7.6) & 90.62 (10.4) & 93.09 (13.0) & 82.31 (8.4) \\
10\% & 75.04 (13.8) & 87.38 (20.9) & 87.30 (25.0) & 83.99 (14.3) \\
15\% & 75.45 (21.2) & 83.77 (30.2) & 81.74 (36.3) & 84.24 (18.2) \\
20\% & 74.49 (28.6) & 82.45 (34.0) & 77.13 (45.6) & 83.85 (22.2) \\
\bottomrule
\end{tabular}}
\end{table}

\paragraph{Held-Out Calibration Videos.}
Table~\ref{tab:ablation} examines whether OOF authentic training scores can replace held-out calibration videos. The OOF variant raises OOD BAcc but also increases in-domain FAR, reaching 59.6\% within the authentic FaceForensics++ subset. In this variant, calibration uses individual fold heads, whereas test prediction averages logits across folds. The scoring procedure therefore differs between calibration and testing, violating the common-score condition of Proposition~\ref{prop:main}. Held-out videos let both stages use the same fold-averaged scorer.

\begin{table}[t]
\centering
\caption{\textbf{Sensitivity to calibration set size.} The size $|\mathcal{C}|$ includes both reference and threshold subsets. Entries show fusion-only benchmark scores with FAR in parentheses, all in percentages. Full-pool sizes appear below the benchmark names; Full uses the entire pool. Reduced-pool results average ten random subsets and five splits per subset within each of ten training seeds. GenVideo retains source calibration; DVF uses authentic target videos.}
\label{tab:supp_csize}
\resizebox{\linewidth}{!}{%
\begin{tabular}{lcccc}
\toprule
$|\mathcal{C}|$ & \textbf{FaceVid OOD} & \textbf{ViF} & \textbf{GenVideo} & \textbf{DVF} \\
\textbf{Full Pool Size} & 2{,}001 & 602 & 602 & 300 \\
\midrule
50 & 64.99 (5.1) & 88.02 (8.3) & 93.68 (9.6) & 73.66 (5.1) \\
100 & 68.11 (5.7) & 89.15 (8.0) & 94.00 (9.7) & 75.44 (4.7) \\
200 & 71.55 (7.7) & 90.21 (10.0) & 93.38 (12.2) & 80.43 (7.1) \\
500 & 71.62 (7.6) & 90.62 (10.2) & 93.19 (12.8) & -- \\
Full & 71.77 (7.6) & 90.62 (10.4) & 93.09 (13.0) & 82.31 (8.4) \\
\bottomrule
\end{tabular}}
\end{table}

\paragraph{Nominal False-Alarm Level.}
Table~\ref{tab:supp_alpha} examines the operating points obtained by varying $\alpha$. FAR increases on all four benchmarks, whereas benchmark scores peak at different levels. The nominal level thus controls the balance between detecting deepfakes and flagging authentic videos. Its choice should reflect the intended tolerance for false alarms. GenVideo retains source calibration in this sweep, while DVF uses authentic target clips.

\begin{table*}[t]
\centering
\caption{\textbf{Investigator training recipes.} Detection scores use training seed 0, the uniform-view development setup, and five calibration splits. Validation reports family-balanced BAcc; testing reports ViF paired accuracy and FaceVid OOD BAcc. Judge acceptance rates assess explanations of held-out deepfakes (Qwen / Gemini). All scores are percentages. Bold values mark column maxima for detection; shading identifies the recipe selected by validation performance. CF denotes counterfactual evidence; dashes indicate unreported results.}
\label{tab:system2}
\begin{tabular}{lcccccc}
\toprule
& \multicolumn{2}{c}{\textbf{Validation BAcc}} & \multicolumn{2}{c}{\textbf{Test Score}} & \multicolumn{2}{c}{\textbf{Judge Acceptance}} \\
\cmidrule(lr){2-3}\cmidrule(lr){4-5}\cmidrule(lr){6-7}
\textbf{Recipe} & \textbf{ViF} & \textbf{FaceVid} & \textbf{ViF} & \textbf{FaceVid OOD} & \textbf{ViF} & \textbf{FaceVid} \\
\midrule
\textcolor{methodcolor}{\textsc{\textbf{Antiphony-F}}} & 94.94 & 89.06 & 90.26 & 71.86 & -- & -- \\
Privileged Teacher & -- & -- & -- & -- & 89 / 60 & 69 / 55 \\
\midrule
SFT (CF) & 94.52 & 88.34 & \textbf{90.81} & 71.78 & 62 / 25 & 45 / 30 \\
SFT (Clean) & 94.92 & 87.96 & 90.36 & 71.68 & -- & -- \\
SFT (CF) + RL (CF) & 95.06 & 88.23 & 90.12 & 71.51 & 54 / 23 & 37 / 26 \\
SFT (Clean) + RL (No CF) & 94.79 & 88.57 & 90.36 & 71.90 & -- & -- \\
\rowcolor[HTML]{EDF3EF}
\textbf{SFT (Clean) + RL (CF)} & \textbf{95.26} & \textbf{89.88} & 90.73 & \textbf{72.22} & 48 / 16 & 23 / 15 \\
\bottomrule
\end{tabular}
\end{table*}

\paragraph{Calibration Set Size.}
We next examine how much authentic footage is needed to estimate the operating point. For each reduced size in Table~\ref{tab:supp_csize}, we average ten random subsets per training seed and five reference/threshold splits per subset. At the tested sizes of 200 and 500 clips, scores on FaceVid, ViF-Bench, and GenVideo remain within half a point of the full-pool results. DVF benefits more from retaining its full pool. FAR also changes with pool size, so a stable benchmark score does not imply an unchanged false-alarm rate.

\subsection{Matched Calibration Access}
\label{sec:supp_equal_calibration}

To distinguish the effect of the fusion rule from access to authentic target footage, Table~\ref{tab:supp_equal_calibration} compares methods under a matched total calibration budget and, separately, a shared threshold subset. We retain the main evaluation sets and the existing authentic target pools of 2{,}000 GenVideo videos and 300 DVF videos. GenVideo uses budgets of 300 and 2{,}000 clips; DVF uses all 300 available clips. No score parameters are fitted on target deepfakes. Appendix~\ref{sec:supp_fresh_calibration} separately examines calibration on previously unused files.

\paragraph{Compared Scores and Thresholds.}
Fisher fusion, mean probability, and logistic stacking combine the same four fixed ViF-trained experts. Mean probability averages their sigmoid scores; logistic stacking is fitted only on source GHCF scores, excluding source calibration clips and linked videos. STIL and X-CLIP use ViF-trained checkpoints, and DRCT uses its released image-detector weights. For each scalar statistic, the threshold is the $\lceil(1-\alpha)(m+1)\rceil$-th ordered authentic calibration score, and scores strictly above it are flagged. The threshold is infinite if the rank exceeds $m$. Fisher first constructs reference $p$-values and then applies this threshold rule to the fused statistic.

\paragraph{Calibration Allocation.}
Under the same total budget, Fisher splits the selected videos into reference and threshold halves, while the other methods use all selected videos to set thresholds. Under the same threshold subset, every method uses Fisher's threshold half; Fisher additionally uses the reference half to transform expert scores. The first comparison matches the total amount of calibration data, while the second matches the videos used to estimate the threshold.

Each budget uses 20 prespecified selections and splits, with repetitions averaged within each of ten expert-training seeds. In the total-budget comparison, using the full 300-clip DVF pool or 2{,}000-clip GenVideo pool leaves the scalar thresholds unchanged across repetitions. Standalone detectors use one checkpoint each and are evaluated once per selection; we therefore do not report a training-seed standard deviation for them.

\paragraph{Effect of the Fusion Rule.}
Fisher outperforms mean probability and source-trained stacking on both targets under both calibration allocations. Its advantage among combinations of the same experts persists when calibration access is matched. Standalone detectors show a different ordering: on DVF, DRCT reaches 84.77 BAcc and 5.18\% FAR, compared with 81.04 and 7.03\% for Fisher. Their recalls are 74.73\% and 69.12\%, respectively, under the total-budget protocol. Matched calibration access does not equalize the achieved FAR, so BAcc should be considered alongside FAR and recall. These comparisons are descriptive; we do not test the statistical significance of the differences.

\begin{table*}[t]
\centering
\caption{\textbf{Investigation budget and fusion weight.} Entries give ten-seed mean gains over fusion alone in FaceVid OOD BAcc and ViF paired accuracy, in percentage points. Cost is the estimated amortized time per input video from the one-A100 ViF measurements.}
\label{tab:supp_rhow}
\begin{tabular}{lccccccccc}
\toprule
& & \multicolumn{4}{c}{\textbf{FaceVid OOD}, $w_I=$} & \multicolumn{4}{c}{\textbf{ViF-Bench}, $w_I=$} \\
\cmidrule(lr){3-6}\cmidrule(lr){7-10}
\textbf{\boldmath$\rho$} & \textbf{Cost (s/video)} & \textbf{0.5} & \textbf{1} & \textbf{2} & \textbf{4} & \textbf{0.5} & \textbf{1} & \textbf{2} & \textbf{4} \\
\midrule
0.05 & 0.36 & +0.26 & +0.33 & +0.33 & +0.26 & +0.16 & +0.26 & +0.23 & +0.23 \\
0.10 & 0.42 & +0.28 & +0.39 & +0.44 & +0.31 & +0.16 & +0.30 & +0.18 & +0.15 \\
0.20 & 0.54 & +0.28 & +0.39 & +0.47 & +0.20 & +0.16 & +0.29 & +0.13 & +0.06 \\
0.30 & 0.66 & +0.28 & +0.39 & +0.47 & +0.16 & +0.16 & +0.29 & +0.08 & -0.00 \\
0.50 & 0.91 & +0.28 & +0.39 & +0.47 & +0.14 & +0.16 & +0.29 & +0.07 & -0.03 \\
1.00 & 1.52 & +0.28 & +0.39 & +0.46 & +0.13 & +0.16 & +0.29 & +0.07 & -0.04 \\
\bottomrule
\end{tabular}
\end{table*}

\paragraph{Consistency of Recomputed Scores.}
We recompute calibration and test scores using each detector's original checkpoint and input pipeline. STIL and X-CLIP match their unmodified official evaluators exactly on a fixed 64-video GenVideo prefix; DRCT matches its saved scores on a fixed 16-video prefix. At the original threshold, recomputation changes five STIL decisions on GenVideo and none for X-CLIP relative to the saved main-table predictions. The matched-calibration study uses recomputed scores; the main table retains the original predictions. We retain calibration selections and prediction records for reproducibility.
\section{Selective Investigation Analysis}
\label{sec:supp_investigation_analysis}

This appendix examines how investigator training, routing, and tool access affect the final decision. We then relate investigation gains to computational cost and use recorded cases to illustrate how tool evidence enters re-fusion.

\subsection{Investigator Training and Selection}
\label{sec:supp_dev}

\paragraph{Comparison and Selection Protocol.}
We compare introducing counterfactual expert evidence during distillation, reinforcement learning, or both. Detection scores in this development comparison use uniform-view expert inputs and average five calibration splits. Validation BAcc averages authentic specificity and mean deepfake recall across source families, giving each family equal weight. It therefore differs from the video-level BAcc used on FaceVid OOD.

The validation data depend on the design choice. Frozen-expert selection uses authentic clips from part of $\mathcal{C}$ and GHCF out-of-fold deepfake scores. Input-view and investigator selection instead use deepfakes excluded from fitting but drawn from families represented in training. Investigator validation thus measures performance on new clips from known families. Encoder-adaptation test comparisons are exploratory. Appendix~\ref{sec:supp_theory} contrasts calibration reuse during selection with independent-calibration assumptions.

Table~\ref{tab:system2} presents validation scores alongside ViF paired test accuracy and FaceVid OOD test BAcc. The judge-acceptance columns address a separate question: whether explanations of held-out deepfakes are supported by the observed evidence according to the model judges. Appendix~\ref{sec:supp_judge} describes the assessment coverage and controls.

\paragraph{Effect of the Training Sequence.}
Clean distillation followed by counterfactual RL gives the highest validation BAcc on both benchmarks: 95.26 on ViF and 89.88 on FaceVid. It also improves both test scores over RL without counterfactuals. Introducing counterfactuals in both stages gives lower validation and test scores than the selected recipe. Within this development comparison, learning tool use from clean demonstrations before introducing misleading expert reports is the more effective sequence.

Selection uses validation performance; counterfactual SFT alone achieves the highest ViF test score. When expert fusion adopts two input views, we retain the investigators trained on uniform-view expert scores, and their visual input remains the uniform view. The main evaluation therefore uses the selected recipe without retraining the investigator for the second expert view.

\subsection{Routing, Tools, and Investigation Budget}
\label{sec:supp_inv}

\paragraph{Where Investigation Changes Decisions.}
We first examine which selected videos benefit from investigation. Table~\ref{tab:supp_strata} groups escalated videos by log-distance to the fusion threshold $\hat\tau$, fusion correctness, and video type. These groupings overlap. For each training seed and reference/threshold split, we recover the threshold from stored calibration scores and compute accuracy within each group. All entries use video-level accuracy, including on ViF, so they characterize the selected videos rather than the overall benchmark metrics.

On FaceVid and ViF, gains concentrate in the closest third of escalated videos. Investigation corrects about one in ten fusion errors within the selected set, but also reverses some correct decisions. On DVF, the transferred investigator produces a less favorable balance: overall BAcc falls by 0.67 points relative to fusion alone (Table~\ref{tab:main}). Threshold proximity identifies decisions that are open to revision; improving them also depends on the investigator's ability to interpret evidence in the target domain.

\clearpage
\begin{table*}[t]
\centering
\caption{\textbf{Where investigation changes decisions.} At $\alpha=5\%$, $\rho=0.3$, and $w_I=1$, results average ten training seeds after averaging calibration splits. Fusion, Full, and Inv.\ give within-stratum video accuracy (\%), including on ViF. Inv.\ uses a 0.5 cutoff on $p_I$. The paired change $\Delta=\mathrm{Full}-\mathrm{Fusion}$ is reported in percentage points as mean $\pm$ sample standard deviation across seeds. Videos give the mean stratum size; groupings by distance, correctness, and video type overlap.}
\label{tab:supp_strata}
\begin{minipage}[t]{0.49\textwidth}
\centering
\resizebox{\linewidth}{!}{%
\begin{tabular}[t]{lccccc}
\toprule
\textbf{Stratum} & \textbf{Videos} & \textbf{Fusion} & \textbf{Full} & \textbf{\boldmath$\Delta$} & \textbf{Inv.} \\
\midrule
\rowcolor[HTML]{EDF3EF}
\multicolumn{6}{c}{\textbf{FaceVid OOD}} \\
\midrule
All Escalated & 2291.0 & 65.7 & 66.4 & $+0.66${\scriptsize$\pm$1.12} & 65.7 \\
Closest Third & 764.0 & 55.0 & 57.2 & $+2.20${\scriptsize$\pm$3.46} & 60.6 \\
Middle Third & 763.0 & 67.1 & 66.9 & $-0.20${\scriptsize$\pm$0.28} & 65.8 \\
Farthest Third & 764.0 & 75.0 & 75.0 & $-0.01${\scriptsize$\pm$0.02} & 70.6 \\
Fusion Wrong & 785.5 & 0.0 & 8.4 & $+8.39${\scriptsize$\pm$1.43} & 39.9 \\
Fusion Correct & 1505.5 & 100.0 & 96.6 & $-3.37${\scriptsize$\pm$1.21} & 79.1 \\
Authentic & 1255.4 & 72.7 & 72.8 & $+0.04${\scriptsize$\pm$1.44} & 73.4 \\
Deepfake & 1035.6 & 57.2 & 58.6 & $+1.42${\scriptsize$\pm$2.05} & 56.2 \\
Open T2V & 282.5 & 60.5 & 62.5 & $+1.95${\scriptsize$\pm$2.34} & 58.7 \\
Open I2V & 227.2 & 38.8 & 40.3 & $+1.46${\scriptsize$\pm$2.13} & 42.2 \\
Commercial & 525.9 & 63.3 & 64.4 & $+1.12${\scriptsize$\pm$2.00} & 60.9 \\
\midrule
\rowcolor[HTML]{EDF3EF}
\multicolumn{6}{c}{\textbf{ViF-Bench}} \\
\midrule
All Escalated & 948.0 & 78.1 & 78.4 & $+0.26${\scriptsize$\pm$2.18} & 75.1 \\
Closest Third & 316.0 & 63.4 & 64.2 & $+0.82${\scriptsize$\pm$6.45} & 60.2 \\
Middle Third & 316.1 & 80.6 & 80.6 & $-0.04${\scriptsize$\pm$0.12} & 78.0 \\
Farthest Third & 315.9 & 90.4 & 90.4 & $+0.00${\scriptsize$\pm$0.00} & 87.0 \\
Fusion Wrong & 207.4 & 0.0 & 11.1 & $+11.10${\scriptsize$\pm$4.71} & 27.8 \\
Fusion Correct & 740.6 & 100.0 & 97.2 & $-2.77${\scriptsize$\pm$2.35} & 88.2 \\
Authentic & 56.9 & 71.6 & 73.2 & $+1.55${\scriptsize$\pm$3.40} & 64.2 \\
Deepfake & 891.1 & 78.5 & 78.7 & $+0.18${\scriptsize$\pm$2.48} & 75.8 \\
Open T2V & 163.3 & 82.1 & 82.8 & $+0.79${\scriptsize$\pm$2.94} & 79.1 \\
Open I2V & 459.9 & 74.0 & 74.1 & $+0.09${\scriptsize$\pm$2.53} & 71.8 \\
Commercial & 267.9 & 84.2 & 84.2 & $-0.03${\scriptsize$\pm$2.27} & 80.4 \\
\bottomrule
\end{tabular}}
\end{minipage}\hfill
\begin{minipage}[t]{0.49\textwidth}
\centering
\resizebox{\linewidth}{!}{%
\begin{tabular}[t]{lccccc}
\toprule
\textbf{Stratum} & \textbf{Videos} & \textbf{Fusion} & \textbf{Full} & \textbf{\boldmath$\Delta$} & \textbf{Inv.} \\
\midrule
\rowcolor[HTML]{EDF3EF}
\multicolumn{6}{c}{\textbf{DVF}} \\
\midrule
All Escalated & 1916.0 & 62.2 & 59.1 & $-3.12${\scriptsize$\pm$3.19} & 63.9 \\
Closest Third & 639.1 & 54.5 & 48.6 & $-5.95${\scriptsize$\pm$5.33} & 58.3 \\
Middle Third & 638.0 & 63.1 & 60.1 & $-2.96${\scriptsize$\pm$3.65} & 63.8 \\
Farthest Third & 638.9 & 69.1 & 68.6 & $-0.44${\scriptsize$\pm$1.06} & 69.6 \\
Fusion Wrong & 723.9 & 0.0 & 13.3 & $+13.27${\scriptsize$\pm$3.77} & 45.8 \\
Fusion Correct & 1192.1 & 100.0 & 86.9 & $-13.08${\scriptsize$\pm$5.57} & 74.6 \\
Authentic & 424.3 & 64.1 & 66.9 & $+2.82${\scriptsize$\pm$3.91} & 37.5 \\
Deepfake & 1491.7 & 61.7 & 56.9 & $-4.79${\scriptsize$\pm$4.57} & 70.8 \\
Open T2V & 580.2 & 62.6 & 56.9 & $-5.70${\scriptsize$\pm$5.37} & 68.8 \\
Open I2V & 838.2 & 61.1 & 56.7 & $-4.32${\scriptsize$\pm$4.39} & 71.9 \\
Commercial & 73.4 & 61.3 & 58.1 & $-3.21${\scriptsize$\pm$3.40} & 73.1 \\
\midrule
\rowcolor[HTML]{EDF3EF}
\multicolumn{6}{c}{\textbf{GenVideo}} \\
\midrule
All Escalated & 4891.0 & 78.1 & 75.5 & $-2.51${\scriptsize$\pm$1.65} & 68.5 \\
Closest Third & 1631.0 & 61.1 & 53.9 & $-7.25${\scriptsize$\pm$5.27} & 57.1 \\
Middle Third & 1630.1 & 79.8 & 79.6 & $-0.27${\scriptsize$\pm$0.57} & 67.0 \\
Farthest Third & 1629.9 & 93.2 & 93.2 & $-0.00${\scriptsize$\pm$0.01} & 81.4 \\
Fusion Wrong & 1073.5 & 0.0 & 7.1 & $+7.10${\scriptsize$\pm$6.23} & 27.3 \\
Fusion Correct & 3817.5 & 100.0 & 94.8 & $-5.18${\scriptsize$\pm$1.54} & 80.2 \\
Authentic & 4053.7 & 75.0 & 72.2 & $-2.76${\scriptsize$\pm$2.18} & 65.9 \\
Deepfake & 837.3 & 93.0 & 91.7 & $-1.27${\scriptsize$\pm$1.16} & 81.1 \\
Open T2V & 837.3 & 93.0 & 91.7 & $-1.27${\scriptsize$\pm$1.16} & 81.1 \\
\bottomrule
\end{tabular}}
\end{minipage}
\par
\vspace{1cm}
\begin{minipage}[t]{0.49\textwidth}
\centering
\resizebox{\linewidth}{!}{%
\begin{tabular}[t]{lccccc}
\toprule
\textbf{Stratum} & \textbf{Videos} & \textbf{Fusion} & \textbf{Full} & \textbf{\boldmath$\Delta$} & \textbf{Inv.} \\
\midrule
\rowcolor[HTML]{EDF3EF}
\multicolumn{6}{c}{\textbf{GenVideo (Target Calibration)}} \\
\midrule
All Escalated & 4891.0 & 90.5 & 90.2 & $-0.27${\scriptsize$\pm$0.35} & 75.6 \\
Closest Third & 1631.0 & 75.6 & 74.8 & $-0.81${\scriptsize$\pm$1.04} & 55.3 \\
Middle Third & 1630.1 & 96.6 & 96.6 & $+0.00${\scriptsize$\pm$0.00} & 82.2 \\
Farthest Third & 1629.9 & 99.2 & 99.2 & $+0.00${\scriptsize$\pm$0.00} & 89.2 \\
Fusion Wrong & 465.8 & 0.0 & 7.1 & $+7.08${\scriptsize$\pm$3.63} & 14.7 \\
Fusion Correct & 4425.2 & 100.0 & 99.0 & $-1.04${\scriptsize$\pm$0.60} & 82.1 \\
Authentic & 2904.7 & 86.9 & 86.9 & $-0.06${\scriptsize$\pm$0.53} & 67.5 \\
Deepfake & 1986.3 & 95.7 & 95.1 & $-0.57${\scriptsize$\pm$0.23} & 87.3 \\
Open T2V & 1986.3 & 95.7 & 95.1 & $-0.57${\scriptsize$\pm$0.23} & 87.3 \\
\bottomrule
\end{tabular}}
\end{minipage}
\end{table*}
\clearpage

\begin{table}[t]
\centering
\caption{\textbf{Tool-group ablations at inference.} The full-toolbox row gives absolute FaceVid OOD BAcc and ViF paired accuracy (\%). Other rows give paired changes in percentage points, reported as mean $\pm$ sample standard deviation over seeds 0--2. The investigator is not retrained; calls to removed tools are refused.}
\label{tab:supp_tools}
\begin{tabular}{lcc}
\toprule
\textbf{Removed} & \textbf{FaceVid OOD} & \textbf{ViF-Bench} \\
\midrule
None (Full Toolbox) & 71.80 & 91.10 \\
Expert Witnesses & $+0.09\pm0.53$ & $-1.11\pm0.53$ \\
Pixel Tools & $+0.18\pm0.29$ & $+0.18\pm0.44$ \\
Evidence Tools & $+0.04\pm0.50$ & $-0.13\pm0.22$ \\
Reliability Cards & $+0.23\pm0.22$ & $-0.24\pm0.16$ \\
\bottomrule
\end{tabular}
\end{table}

\paragraph{Contribution of Tool Groups.}
To assess the tools used by the trained investigator, we remove each group at inference without retraining (See Table~\ref{tab:supp_tools}). Expert witnesses provide detector reports, pixel tools expose frames and low-level differences, and evidence tools measure temporal, motion, and text consistency. Removing reliability cards leaves expert scores available but withholds their validation-based summaries, isolating the accompanying guidance.

Removing expert witnesses produces the largest observed decrease on ViF ($-1.11$ points), indicating that their reports remain useful alongside visual inspection. Other removals have smaller effects across the three seeds, with directions that vary by benchmark.

\paragraph{Routing Choices.}
We compare threshold proximity, random selection, and individual-expert suspicion in the uniform-view, seed-0 development configuration. At $\rho=0.3$, selecting videos nearest to the threshold gives 90.73 on ViF-Bench and 72.22 on FaceVid OOD, compared with 90.41 and 71.99 for random selection. Selecting unflagged videos with the highest individual-expert suspicion gives 90.91 and 72.39.

The individual-expert rule reaches 68\% of the deepfakes missed by fusion on FaceVid OOD, compared with 47\% for threshold proximity. Its accuracy advantage nevertheless remains within 0.2 points because the investigator rarely reverses a confident missed detection. Finding errors and correcting them are separate challenges: inspecting more missed deepfakes does not necessarily produce a proportional gain in final detections.

\paragraph{Investigation Budget and Fusion Weight.}
Table~\ref{tab:supp_rhow} varies the escalation rate $\rho$ and investigator weight $w_I$ in the main ten-seed setting. FaceVid and ViF show little additional gain beyond $\rho=0.10$, consistent with improvements concentrating near the threshold. In these experiments, inspecting more videos therefore increases computation with limited gains in detection. The two parameters control different parts of the procedure: $\rho$ sets how many videos are inspected, while $w_I$ sets how strongly their investigation scores contribute to re-fusion.

\subsection{Inference Cost and Investigation Examples}
\label{sec:supp_cost}

\paragraph{Amortized Batch Cost.}
Every video incurs expert-scoring cost; selected videos additionally require tool preparation and MLLM investigation. We measure these stages separately to estimate the effect of the investigation budget on total computation. Tool preparation includes frame statistics, OCR, point tracking, and temporal-expert features.

All stage measurements use one A100. We estimate marginal time per video as $(t_{n_1}-t_{n_0})/(n_1-n_0)$, using the difference between two batch sizes to remove fixed startup overhead. Expert fusion and baselines use nested ViF subsets of 10 and 100 videos. Investigation uses 12 and 120 escalated videos, with one vLLM replica allowing up to 64 concurrent episodes. Table~\ref{tab:supp_stage_costs} summarizes the stage costs.

The rounded subtotals give $0.299+\rho\,1.219$ seconds per input video. At $\rho=0.3$, this approximation gives about 0.665\,s. The more precise stage subtotals give 0.664405\,s, reported as 0.66\,s in the main text. The estimate counts two head-and-fusion passes as an upper bound for two-view scoring, although fusion runs once. It excludes fixed startup overhead and one-time calibration scoring. These measurements describe amortized batch cost, not individual response latency.

\begin{table}[t]
\centering
\caption{\textbf{Amortized Stage Costs on One A100.}}
\label{tab:supp_stage_costs}
\resizebox{\linewidth}{!}{%
\begin{tabular}{lr}
\toprule
\textbf{Stage} & \textbf{Seconds/Video} \\
\midrule
\rowcolor[HTML]{EDF3EF}
\multicolumn{2}{c}{\textbf{Expert Fusion: per Input Video}} \\
\midrule
Uniform-View Feature Extraction & 0.13773 \\
Offset-View Feature Extraction & 0.14870 \\
Head Scoring and Fusion (Counted Twice) & 0.01241 \\
\textbf{Fusion Subtotal} & \textbf{0.29884} \\
\midrule
\rowcolor[HTML]{EDF3EF}
\multicolumn{2}{c}{\textbf{Investigation: per Escalated Video}} \\
\midrule
Frame Statistics & 0.02456 \\
OCR & 0.37779 \\
Point Tracking & 0.19392 \\
Temporal-Expert Features & 0.03379 \\
Concurrent Investigator Episodes & 0.58850 \\
\textbf{Investigation Subtotal} & \textbf{1.21855} \\
\bottomrule
\end{tabular}}
\end{table}

\paragraph{Serving Configuration and Latency.}
The corresponding marginal costs are 61.9\,s for ARGUS, 4.8\,s for Skyra, and 0.06--0.19\,s for the video detectors and VideoVeritas. Each implementation retains its serving configuration on one A100: ARGUS uses its released single-process harness, while our investigator batches concurrent episodes. The cost ratios therefore reflect both the detection pipeline and its batching. Median latency for an individual investigator episode is 38.4\,s in the same run, distinguishing individual response time from batch throughput.

Separate seed-0, uniform-view development runs characterize the interaction pattern. The investigator averages 3.1 tool calls and 2.8 model calls on ViF, and 4.7 tool calls and 3.3 model calls on FaceVid. These runs use two vLLM replicas on two A100 GPUs with 64 concurrent episodes and have median episode latencies of 26\,s and 31\,s, respectively. The interaction statistics describe this development setup; the cost comparison above uses the one-A100 measurements.

\begin{figure*}[!tp]
\centering
\includegraphics[width=\textwidth,height=0.92\textheight,keepaspectratio]{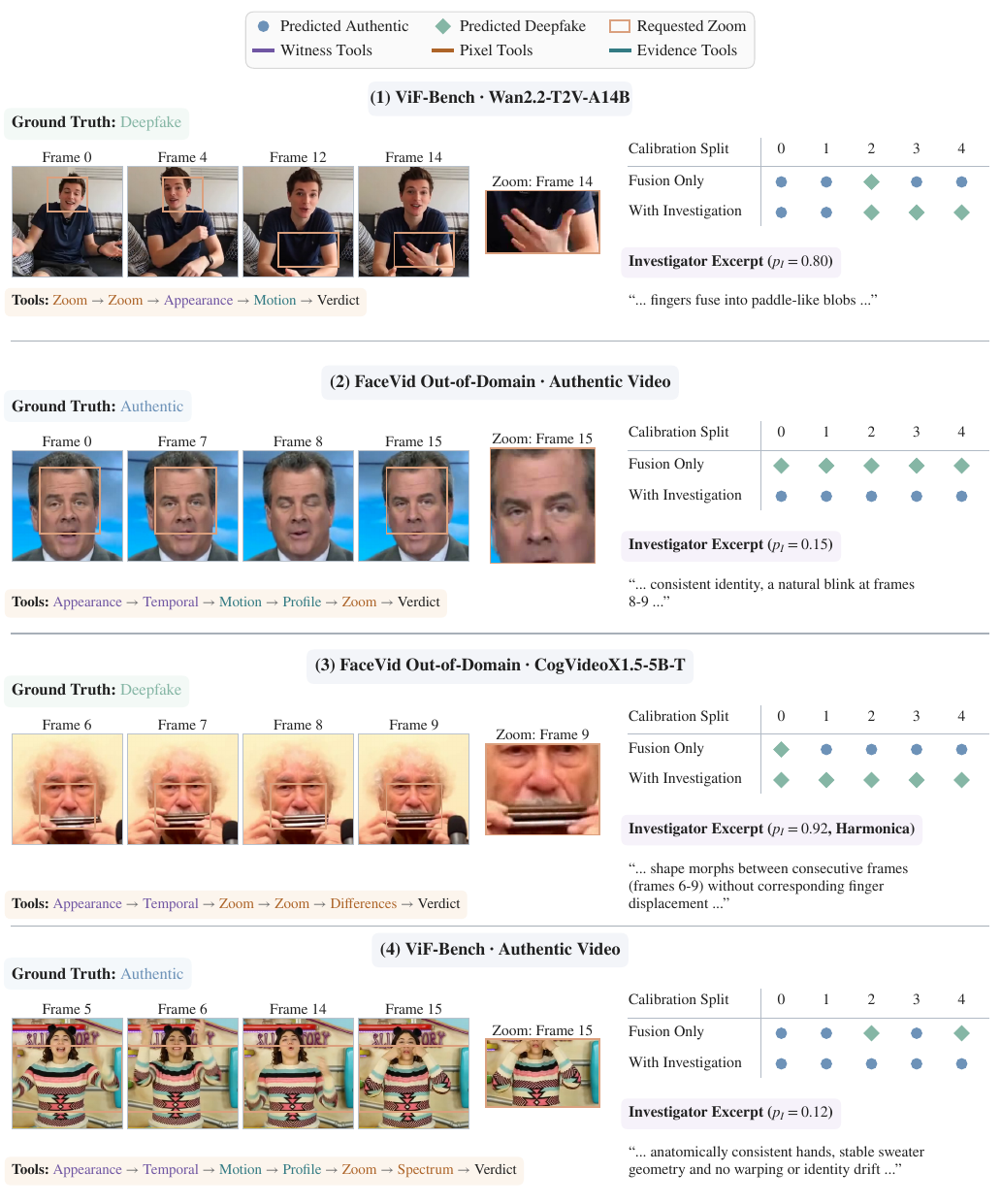}
\caption{\textbf{Recorded investigations: examples 1--4.} Four selected seed-0 episodes show sampled frames and requested zooms. Columns 0--4 compare fusion-only and final predictions across five calibration splits. Arrows follow the recorded tool-call order. Excerpts are investigator claims, not human-verified annotations.}
\label{fig:qualitative}
\end{figure*}

\begin{figure*}[!tp]
\centering
\includegraphics[width=\textwidth,height=0.92\textheight,keepaspectratio]{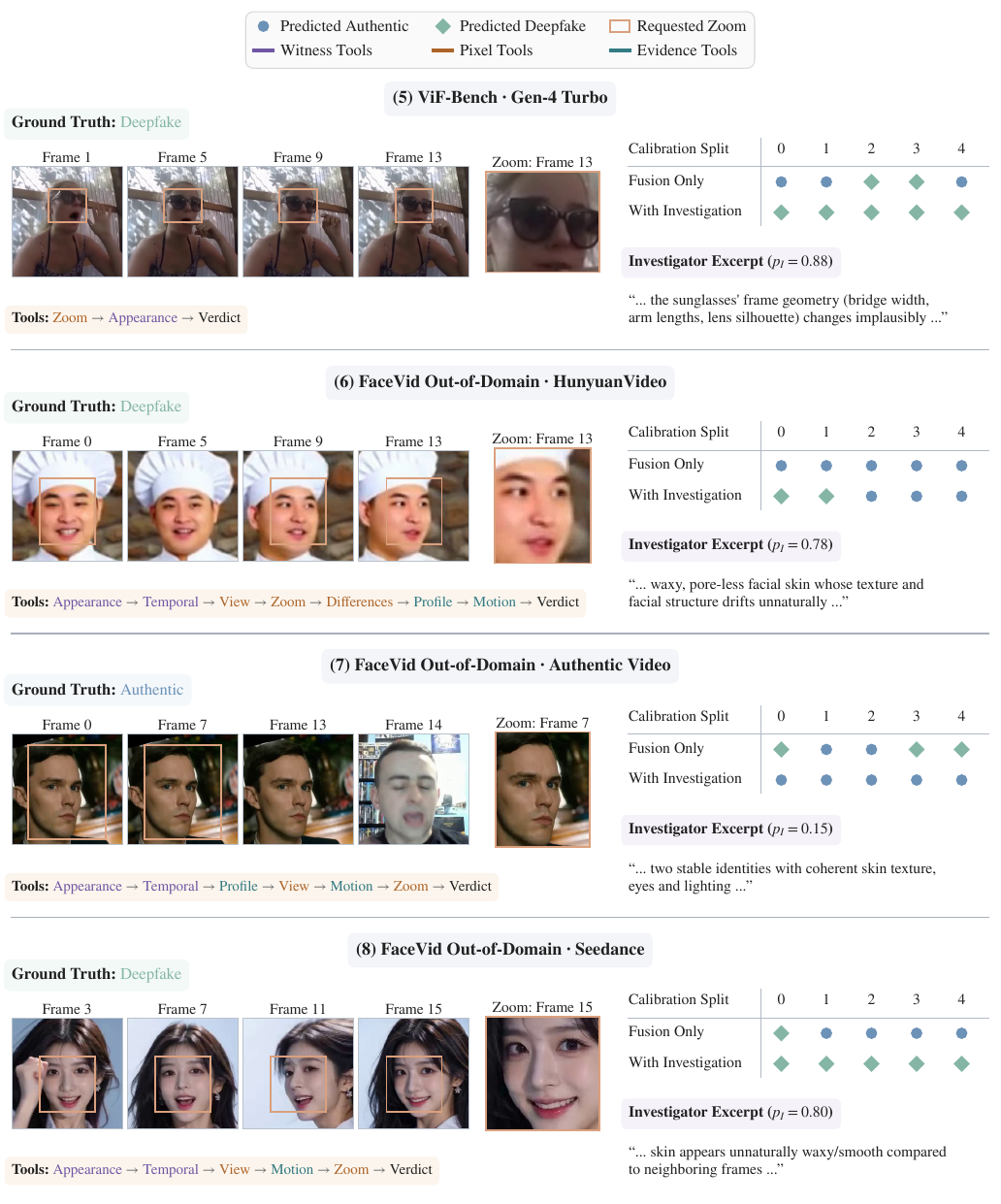}
\caption{\textbf{Recorded investigations: examples 5--8.} Four additional seed-0 episodes follow the frame, tool-call, prediction, and excerpt conventions in Figure.~\ref{fig:qualitative}.}
\label{fig:qualitative_2}
\end{figure*}

\begin{figure*}[!tp]
\centering
\includegraphics[width=\textwidth,height=0.92\textheight,keepaspectratio]{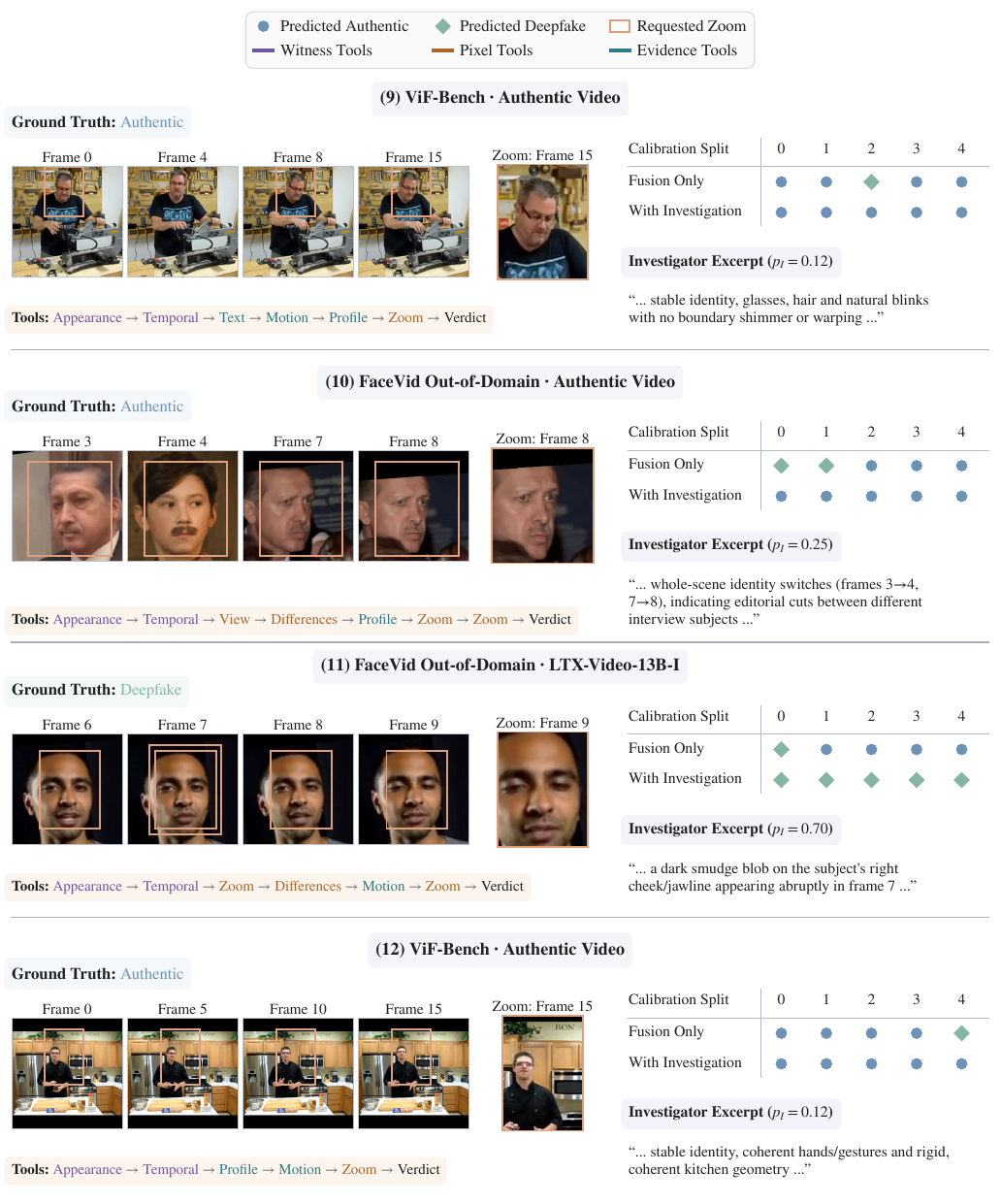}
\caption{\textbf{Recorded investigations: examples 9--12.} Four additional seed-0 episodes follow the frame, tool-call, prediction, and excerpt conventions in Figure.~\ref{fig:qualitative}.}
\label{fig:qualitative_3}
\end{figure*}

\begin{figure*}[!tp]
\centering
\includegraphics[width=\textwidth,height=0.92\textheight,keepaspectratio]{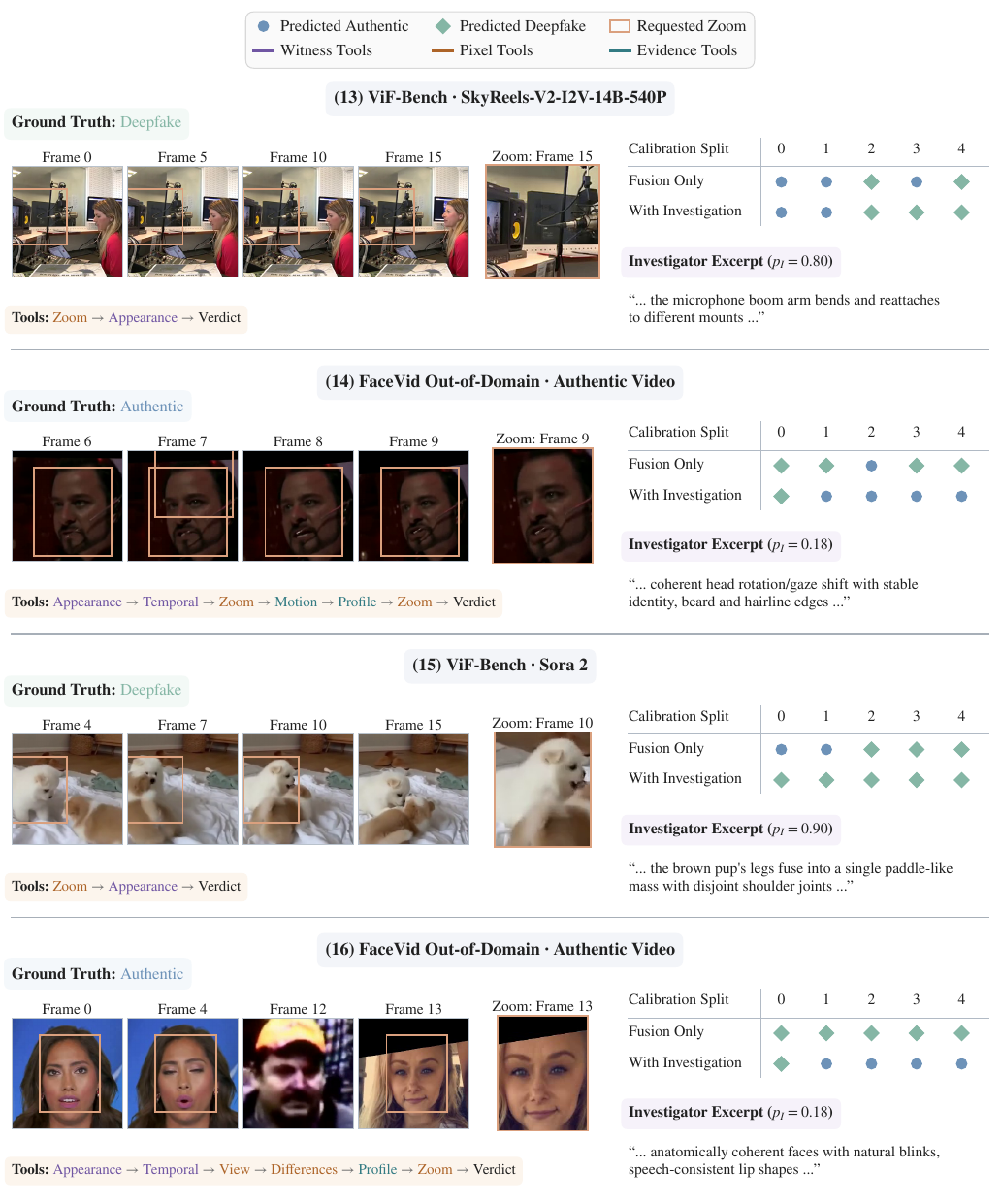}
\caption{\textbf{Recorded investigations: examples 13--16.} Four additional seed-0 episodes follow the frame, tool-call, prediction, and excerpt conventions in Figure.~\ref{fig:qualitative}.}
\label{fig:qualitative_4}
\end{figure*}

\begin{figure*}[!tp]
\centering
\includegraphics[width=\textwidth,height=0.92\textheight,keepaspectratio]{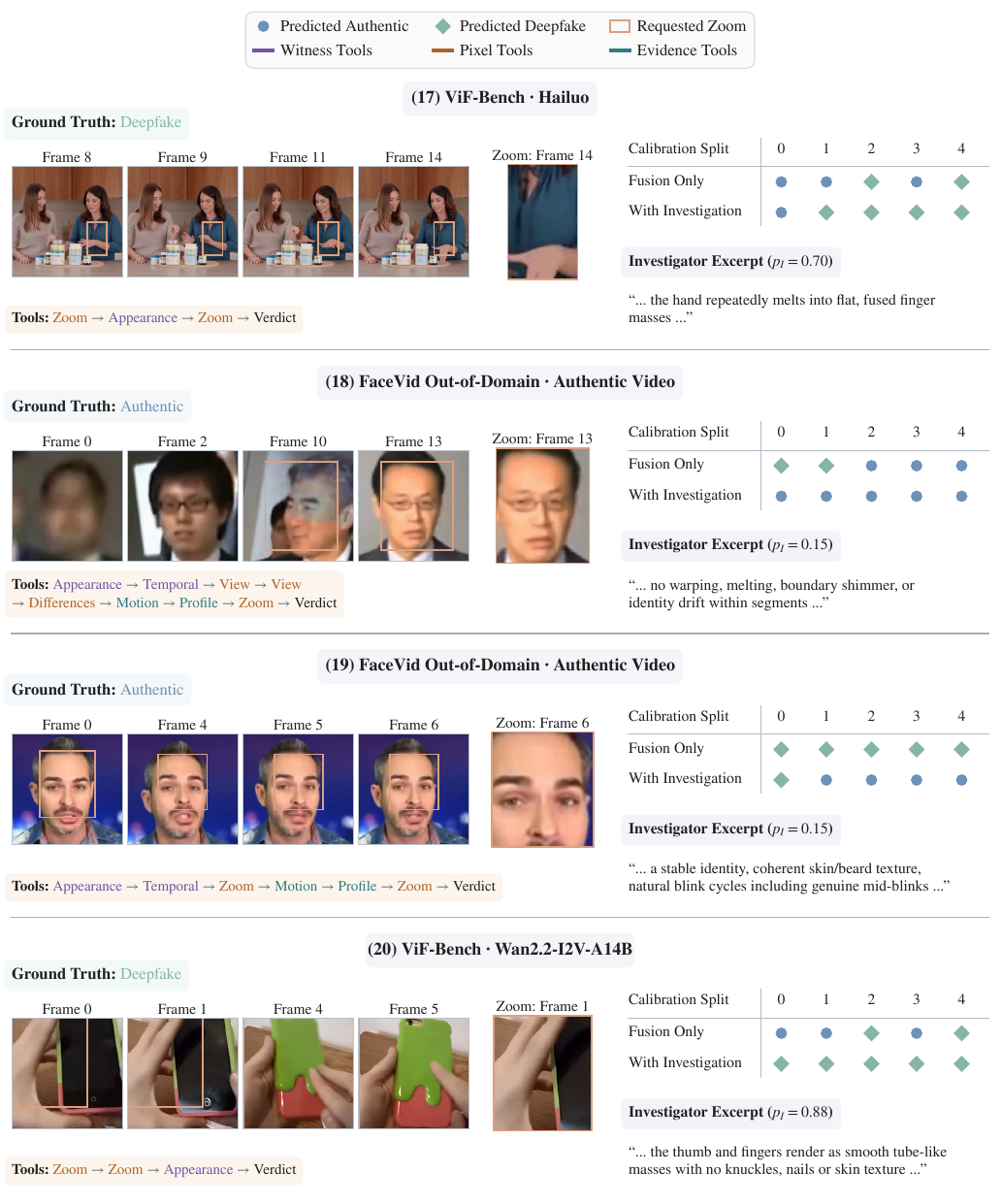}
\caption{\textbf{Recorded investigations: examples 17--20.} Four additional seed-0 episodes follow the frame, tool-call, prediction, and excerpt conventions in Figure.~\ref{fig:qualitative}.}
\label{fig:qualitative_5}
\end{figure*}

\paragraph{Reading the Investigation Examples.}
Figures~\ref{fig:qualitative}--\ref{fig:qualitative_5} present twenty selected investigations across five pages, with four examples per page. Each case shows sampled frames, tool calls, and a recorded rationale, alongside fusion-only and final predictions under five calibration splits. Arrows follow the recorded call order. The cases below illustrate how investigation can correct a decision or introduce an error.

\paragraph{From Tool Evidence to Re-Fusion.}
Example~(1) in Figure.~\ref{fig:qualitative} follows a Wan2.2-T2V-A14B video from its 16-frame overview to the final decision. The investigator first requests two zooms: frames 0, 2, and 4 around the face, then frames 12 and 14 around the hand. It next calls the appearance expert and the motion-consistency tool. Recorded outputs include an appearance score of 0.818, path tortuosity at the 92nd authentic percentile, and 87 vanish/reappear track events. The investigator submits $p_I(v)=0.80$ and the type \texttt{fully\_synthetic}. Its recorded rationale includes ``the wrist melts into the shirt''; this is an investigator claim, not a human annotation. After authentic-reference calibration, its contribution $-2w_I\log\pi_I(v)$ is added to the expert statistic before applying the re-fusion threshold. Across the five calibration splits, saved fusion decisions are $(0,0,1,0,0)$ and final decisions are $(0,0,1,1,1)$, where 1 flags a deepfake. The same investigation thus changes two decisions, while calibration still determines the outcome in each split.

\paragraph{Correcting and Introducing False Alarms.}
In example~(2) of Figure.~\ref{fig:qualitative}, the authentic news clip receives $p_I(v)=0.15$ and moves from five flags to none. In a separate failure case, an authentic clip has 16 identical sampled frames: fusion accepts it, but the investigator treats the lack of change as implausible and flags it. These cases illustrate why verdict accuracy and the visual support for a rationale should be assessed separately. Appendix~\ref{sec:supp_judge} describes the explanation-grounding evaluation.

\paragraph{Inspection Beyond a Low Expert Score.}
Example~(11) in Figure.~\ref{fig:qualitative_3} shows how local inspection can complement an expert report. The appearance tool assigns a deepfake score of 0.00, but the investigator requests face zooms, frame differences, and motion analysis. Its recorded rationale cites a changing cheek patch, and it returns $p_I(v)=0.70$. Fusion flags this video in one of five calibration splits; re-fusion flags it in all five.

\paragraph{Interpreting Abrupt Edits.}
Example~(18) in Figure.~\ref{fig:qualitative_5} illustrates the complementary behavior on authentic footage. The investigator inspects frames around scene changes, followed by frame-difference, motion, temporal-profile, and zoom tools. Its recorded rationale attributes the discontinuities to edits between subjects rather than continuous facial deformation. With $p_I(v)=0.15$, re-fusion removes the two false alarms and yields authentic predictions under all five calibration splits.
\subsection{Explanation Grounding}
\label{sec:supp_judge}

We assess whether the investigator's stated reasons are supported by the frames and tool outputs it observed. This complements verdict accuracy: a correct decision can still contain an unsupported visual claim. Two model judges assess teacher and student explanations under a common protocol, with acceptance rates summarized in Table~\ref{tab:supp_ground}. These rates measure support according to the judges rather than human-verified correctness.

\begin{table}[t]
\centering
\caption{\textbf{Judge acceptance of explanation support (\%).} Entries report Qwen / Gemini acceptance rates using the cohorts and valid-response denominators described in Appendix~\ref{sec:supp_judge}. The teacher saw privileged information; students did not. Shading marks the selected student.}
\label{tab:supp_ground}
\resizebox{\linewidth}{!}{%
\begin{tabular}{lcccc}
\toprule
\multirow{2}{*}{\textbf{Source}}
& \multicolumn{2}{c}{\textbf{ViF}}
& \multicolumn{2}{c}{\textbf{FaceVid}} \\
\cmidrule(lr){2-3}\cmidrule(lr){4-5}
& \textbf{Fake} & \textbf{Real}
& \textbf{Fake} & \textbf{Real} \\
\midrule
ViF-CoT Reference Reasoning
& 53 / 27 & -- & -- & -- \\
Privileged Teacher
& 89 / 60 & 93 / 99 & 69 / 55 & 95 / 98 \\
\midrule
SFT, CF Episodes
& 62 / 25 & 54 / 27 & 45 / 30 & 72 / 71 \\
SFT (CF) + RL (CF)
& 54 / 23 & 69 / 69 & 37 / 26 & 66 / 70 \\
\rowcolor[HTML]{EDF3EF}
SFT (Clean) + RL (CF)
& 48 / 16 & 60 / 58 & 23 / 15 & 62 / 69 \\
\bottomrule
\end{tabular}%
}
\end{table}

\paragraph{Assessment Protocol.}
Qwen3.8-Max via DashScope and Gemini-3.8-Flash each receive the expert-fusion suspicion, 16 overview frames, ordered tool calls and outputs, and the final verdict and rationale. The ground-truth label is withheld. A valid JSON response with \texttt{supported=true} counts as acceptance; specificity and confidence are collected separately and do not affect this criterion. Both judges use temperature 0 and a 4{,}096-token response budget. We exclude GPT-6.1 Sol because it accepted only 1\% of the reference reasoning in an earlier judge control.

We reconstruct the visual observations by replaying tool calls with the same implementation. Replayed numerical outputs are checked against the record to three decimal places. Judges receive the recorded output text, truncated to 1{,}500 characters per call, together with the replayed images. This keeps the assessment tied to the observations made during the investigation.

\paragraph{Evaluation Cohorts and Coverage.}
The teacher and all students are assessed on the same requested cohorts: 243 held-out ViF deepfakes and 100 authentic calibration videos, and 150 held-out FaceVid deepfakes and 100 authentic calibration videos. For the ViF deepfakes, the regions identified by human artifact annotations remain visible after center cropping, allowing a reference-reasoning comparison on the same clips. These cohorts are fixed independently of test-time escalation. Teacher episodes use unmodified expert evidence, and submitted explanations are included regardless of whether their verdicts are correct.

We compute rounded acceptance percentages from valid judge responses for available explanations. For ViF deepfakes, Qwen evaluates 239--243 cases per row and Gemini 240--243; the respective counts for authentic clips are 99--100 and 100. For FaceVid, Qwen evaluates 149 deepfakes and Gemini 150, each with 100 authentic clips.

\paragraph{Reference and Mismatched-Reasoning Controls.}
We first examine how the judges respond to reference evidence and to reasoning from a different video. The control uses 243 held-out ViF deepfakes with human-annotated artifact types, frame intervals, and bounding boxes whose regions remain visible after center cropping. Each judge receives overview frames and zoomed views of these regions, paired with one of three texts: the annotation stated as a claim (A), the video's ViF-CoT reference reasoning with timestamps mapped to frame indices (B), or another video's reasoning (C).

Qwen accepts 47\%, 53\% and 0\%, respectively; Gemini accepts 26\%, 27\% and 0\%. Both reject all mismatched reasoning in this control set, demonstrating sensitivity to the correspondence between the text and video. They also reject many claims derived from human annotations and many reference rationales. Acceptance can therefore compare explanations under a shared protocol, but a rejection alone does not establish that a claim lacks visual support.

\paragraph{Detection Performance and Explanation Support.}
The detection and explanation comparisons favor different training recipes. On ViF, the selected investigator receives 48\% acceptance from Qwen and 16\% from Gemini, compared with 62\% and 25\% for counterfactual SFT alone. The same ordering holds on FaceVid, although the selected recipe has the best validation detection scores. Improving detection through the current training objective therefore does not ensure stronger support for the resulting explanations.

The privileged teacher receives higher acceptance than the students while also benefiting from labels and annotations during demonstration generation. Its results provide context for the supervision available to the student, rather than a comparison under identical information. Together with the controls, these findings support assessing explanation grounding separately when selecting and training investigators.

\paragraph{Judge Instruction and Response Schema.}
The judge instruction in Figure.~\ref{fig:supp_judge_prompt} is shared across teacher, student, and control evaluations; only the supplied observations and rationale differ.

\begin{figure}[t]
\centering
\begin{prompt}{Judge Instruction and Response Schema}
You audit a video-forensics report. You are shown ALL observations the analyst collected about one video: the case message (with the fast detector's suspicion score), 16 uniformly sampled frames (indexed 0--15), then every tool call with its output and the images it returned, followed by the analyst's final verdict (probability that the video is fake, forgery type) and rationale. You do not know the true label.

1. Check every concrete visual claim in the rationale (e.g. "the ear melts in frames 0--1", "text changes between frames 10 and 12", "no warping on the face") against the images. A claim is supported only if a careful viewer could verify it from these images. Claims about tool numbers or the fast detector are supported if they match the tool outputs or the case message.

2. Decide whether the rationale is specific: it points to particular frames, regions or measurements rather than generic statements.

Reply with JSON only:

{
  "supported": true|false,
  "unsupported_claims": ["..."],
  "specific": true|false,
  "confidence": 0.0-1.0
}

"supported" is true only if all concrete visual claims that the verdict relies on are visible.
\end{prompt}
\caption{\textbf{Judge instruction and response schema.}}
\label{fig:supp_judge_prompt}
\end{figure}

\begin{table*}[t]
\centering
\caption{\textbf{Per-generator paired accuracy on ViF-Bench (\%).} Expert fusion alone is compared with \method using an investigation budget of $\rho=0.3$, under two-view and uniform-view expert inputs. Entries are the mean $\pm$ sample standard deviation across ten training seeds, after averaging calibration splits within each seed. The final row averages the 19 generators.}
\label{tab:supp_pergen}
\begin{tabular}{lcccc}
\toprule
\multirow{2}{*}{\textbf{Generator}}
& \multicolumn{2}{c}{\textbf{Two-View Expert Inputs}} & \multicolumn{2}{c}{\textbf{Uniform-View Expert Inputs}} \\
\cmidrule(lr){2-3}\cmidrule(lr){4-5}
& \textbf{Fusion Only} & \method & \textbf{Fusion Only} & \method \\
\midrule
Wan2.2-I2V-A14B & 73.58{\scriptsize$\pm$0.27} & 74.02{\scriptsize$\pm$0.79} & 72.56{\scriptsize$\pm$0.52} & 73.44{\scriptsize$\pm$0.68} \\
Wan2.2-TI2V-5B (I2V) & 76.93{\scriptsize$\pm$0.39} & 77.02{\scriptsize$\pm$1.01} & 76.48{\scriptsize$\pm$0.47} & 77.12{\scriptsize$\pm$0.84} \\
Gen-4 Turbo & 80.32{\scriptsize$\pm$0.43} & 80.91{\scriptsize$\pm$1.06} & 79.82{\scriptsize$\pm$0.54} & 80.86{\scriptsize$\pm$1.16} \\
Wan2.1-VACE-1.3B (T2V) & 86.05{\scriptsize$\pm$0.27} & 86.58{\scriptsize$\pm$0.58} & 85.88{\scriptsize$\pm$0.27} & 86.53{\scriptsize$\pm$0.55} \\
SkyReels-V2-I2V-14B-540P & 88.57{\scriptsize$\pm$0.23} & 88.93{\scriptsize$\pm$0.64} & 88.01{\scriptsize$\pm$0.22} & 88.46{\scriptsize$\pm$0.41} \\
MiniMax Hailuo 02 & 90.61{\scriptsize$\pm$0.31} & 90.62{\scriptsize$\pm$0.52} & 90.23{\scriptsize$\pm$0.25} & 90.28{\scriptsize$\pm$0.38} \\
LTX-Video (13B, I2V) & 93.12{\scriptsize$\pm$0.16} & 93.32{\scriptsize$\pm$0.52} & 92.99{\scriptsize$\pm$0.19} & 93.36{\scriptsize$\pm$0.48} \\
Wan2.2-T2V-A14B & 93.07{\scriptsize$\pm$0.17} & 93.48{\scriptsize$\pm$0.49} & 92.98{\scriptsize$\pm$0.20} & 93.44{\scriptsize$\pm$0.44} \\
Sora 2 & 93.49{\scriptsize$\pm$0.26} & 93.93{\scriptsize$\pm$0.59} & 93.43{\scriptsize$\pm$0.19} & 93.89{\scriptsize$\pm$0.68} \\
HunyuanVideo-I2V & 93.73{\scriptsize$\pm$0.16} & 94.04{\scriptsize$\pm$0.46} & 93.58{\scriptsize$\pm$0.18} & 93.91{\scriptsize$\pm$0.51} \\
PixVerse V4.5 & 94.31{\scriptsize$\pm$0.19} & 94.51{\scriptsize$\pm$0.57} & 94.15{\scriptsize$\pm$0.23} & 94.41{\scriptsize$\pm$0.47} \\
HunyuanVideo & 94.48{\scriptsize$\pm$0.14} & 94.82{\scriptsize$\pm$0.59} & 94.35{\scriptsize$\pm$0.18} & 94.64{\scriptsize$\pm$0.45} \\
Wan2.2-TI2V-5B (T2V) & 94.55{\scriptsize$\pm$0.16} & 94.86{\scriptsize$\pm$0.62} & 94.40{\scriptsize$\pm$0.19} & 94.73{\scriptsize$\pm$0.47} \\
CogVideoX1.5-5B (T2V) & 94.73{\scriptsize$\pm$0.15} & 94.92{\scriptsize$\pm$0.57} & 94.60{\scriptsize$\pm$0.21} & 94.83{\scriptsize$\pm$0.48} \\
LTX-Video (13B, T2V) & 94.74{\scriptsize$\pm$0.14} & 95.01{\scriptsize$\pm$0.62} & 94.63{\scriptsize$\pm$0.18} & 94.90{\scriptsize$\pm$0.46} \\
Kling 1.0 & 94.96{\scriptsize$\pm$0.12} & 95.06{\scriptsize$\pm$0.46} & 94.93{\scriptsize$\pm$0.15} & 95.02{\scriptsize$\pm$0.37} \\
SkyReels-V2-T2V-14B-540P & 94.81{\scriptsize$\pm$0.14} & 95.07{\scriptsize$\pm$0.61} & 94.69{\scriptsize$\pm$0.18} & 94.96{\scriptsize$\pm$0.45} \\
Wan2.1-T2V-1.3B & 94.81{\scriptsize$\pm$0.14} & 95.07{\scriptsize$\pm$0.61} & 94.69{\scriptsize$\pm$0.18} & 94.96{\scriptsize$\pm$0.45} \\
Pika 2.0 & 94.98{\scriptsize$\pm$0.16} & 95.19{\scriptsize$\pm$0.59} & 94.83{\scriptsize$\pm$0.20} & 95.03{\scriptsize$\pm$0.39} \\
\midrule
\rowcolor[HTML]{EDF3EF}
\textbf{Mean} & 90.62{\scriptsize$\pm$0.13} & 90.91{\scriptsize$\pm$0.41} & 90.38{\scriptsize$\pm$0.16} & 90.78{\scriptsize$\pm$0.41} \\
\bottomrule
\end{tabular}%
\end{table*}

\section{Generalization across Generators and Video Conditions}
\label{sec:supp_additional}

These studies examine how detection changes across generators, benchmarks, and video-processing conditions. We first compare performance by generator, then test transfer to recent talking-head generators and other benchmarks with fixed experts. The processing experiments examine whether matching authentic calibration footage to the evaluation pipeline helps preserve detection performance. Appendix~\ref{sec:supp_protocols} specifies the datasets and evaluation units.

\subsection{Variation across Generators}
\label{sec:supp_pergen}

To identify which generators remain difficult, Table~\ref{tab:supp_pergen} compares ViF-Bench paired accuracy for expert fusion and the full method under two-view and uniform-view expert inputs. Investigation increases the mean score for every generator in both settings, but substantial differences between generators remain. With two-view inputs, \method reaches 74.02\% on Wan2.2-I2V-A14B and 93.48\% on Wan2.2-T2V-A14B. These differences are larger than the gains from investigation: selective inspection refines the initial decisions, while performance remains closely tied to the evidence captured by the visual experts.

\subsection{Recent Talking-Head Generators}
\label{sec:supp_df26}

DF26~\cite{df26} tests transfer to recent talking-head generators. Its 271 authentic and 2{,}420 generated videos cover ten configurations: the commercial image-to-video models Grok Imagine, Kling~3.0, Veo~3.1, and Wan~2.6, and the open models HunyuanVideo~1.5, LTX~2.3, and Wan~2.2 in image-to-video and text-to-video modes. The primary protocol follows the dataset's evaluation-only license: all methods are evaluated zero-shot, and \method retains FaceVid- or ViF-trained experts with source calibration. With the dataset authors' permission, a separate experiment recalibrates on authentic DF26 videos using two-fold cross-fitting by source clip. Each video is evaluated using authentic calibration clips from the opposite fold; no DF26 deepfakes are used for calibration.

Table~\ref{tab:supp_df26} reports decision metrics alongside AUROC and equal error rate (EER), macro-averaged over the ten generator configurations. The ranking metrics distinguish score discrimination from performance at the selected threshold. Effort and TALL each exclude five videos without a verdict under their released scoring conventions. Decision metrics for the remaining methods include all 2{,}691 videos, with invalid API answers counted as errors.

\clearpage
\begin{table*}[!t]
\centering
\caption{\textbf{Transfer to DF26 (\%).} FV denotes FaceVid-trained heads; Recal. denotes two-fold recalibration on held-out authentic DF26 clips. Our results average ten seeds. AUROC and EER are macro means over ten generator configurations and use the expert-fusion statistic for both \method variants; verdict-only baselines omit these metrics. Effort and TALL each exclude five videos without a verdict. Bold and underline mark the best and second-best results, respectively; lower EER and higher values for all other metrics are better.}
\label{tab:supp_df26}
\resizebox{!}{0.46\textheight}{%
\begin{tabular}{lrrrrr}
\toprule
\textbf{Method} & \textbf{AUROC} & \textbf{EER} & \textbf{BAcc} & \textbf{Recall} & \textbf{Spec.} \\
\midrule
\rowcolor[HTML]{EDF3EF}
\multicolumn{6}{c}{\textbf{Image Detectors}} \\
\midrule
AIDE & 50.54 & 49.71 & 51.03 & 10.91 & 91.14 \\
DRCT & \underline{69.61} & 35.41 & 63.75 & 61.82 & 65.68 \\
NPR (Image) & 68.24 & 36.16 & 53.92 & 11.16 & 96.68 \\
SAFE & 35.70 & 60.03 & 50.00 & 0.00 & \textbf{100.00} \\
UnivFD & 43.75 & 54.12 & 49.92 & 0.21 & \underline{99.63} \\
\midrule
\rowcolor[HTML]{EDF3EF}
\multicolumn{6}{c}{\textbf{Face Specialists}} \\
\midrule
DFD-FCG (FV) & 65.03 & 38.14 & 50.10 & 0.21 & \textbf{100.00} \\
Effort (FV) & 51.35 & 50.69 & 50.52 & 1.41 & \underline{99.63} \\
TALL (FV) & 68.42 & 36.24 & 55.84 & 15.39 & 96.28 \\
\midrule
\rowcolor[HTML]{EDF3EF}
\multicolumn{6}{c}{\textbf{Video Detectors}} \\
\midrule
DeMamba (FV) & 56.69 & 45.96 & 50.39 & 0.79 & \textbf{100.00} \\
DeMamba (ViF) & 68.84 & 35.87 & 57.82 & 85.74 & 29.89 \\
NPR (Video) (FV) & 58.04 & 43.22 & 54.59 & 52.36 & 56.83 \\
NPR (Video) (ViF) & 55.79 & 47.07 & 53.61 & 68.47 & 38.75 \\
STIL (FV) & 56.79 & 45.66 & 53.89 & 55.74 & 52.03 \\
STIL (ViF) & 66.88 & 37.29 & 56.54 & 84.67 & 28.41 \\
X-CLIP (FV) & 61.19 & 43.23 & 50.56 & 1.12 & \textbf{100.00} \\
X-CLIP (ViF) & \textbf{76.02} & \textbf{28.84} & \underline{67.53} & 68.64 & 66.42 \\
\midrule
\rowcolor[HTML]{EDF3EF}
\multicolumn{6}{c}{\textbf{General MLLMs}} \\
\midrule
DeepSeek-V4.1-Flash & -- & -- & 57.03 & 31.40 & 82.66 \\
Gemini-3.8-Flash & -- & -- & 53.54 & 10.04 & 97.05 \\
GPT-6.1 Sol & -- & -- & 54.53 & 11.28 & 97.79 \\
InternVL3.5-8B & -- & -- & 49.94 & 1.36 & 98.52 \\
Kimi K3.1 Pro & -- & -- & 51.73 & 6.78 & 96.68 \\
Qwen2.5-VL-7B & -- & -- & 49.94 & 0.62 & 99.26 \\
Qwen3-VL-8B & -- & -- & 50.02 & 4.46 & 95.57 \\
Qwen3-VL-Plus & -- & -- & 50.67 & 2.81 & 98.52 \\
Qwen3.8-27B & -- & -- & 54.36 & 23.84 & 84.87 \\
Qwen3.8-Flash & -- & -- & 56.64 & 21.78 & 91.51 \\
Qwen3.8-Max & -- & -- & 57.15 & 19.09 & 95.20 \\
Qwen3.8-Omni-Flash & -- & -- & 56.33 & 20.79 & 91.88 \\
\midrule
\rowcolor[HTML]{EDF3EF}
\multicolumn{6}{c}{\textbf{Specialized MLLMs}} \\
\midrule
BusterX++ & -- & -- & 52.02 & 4.05 & \textbf{100.00} \\
Skyra & -- & -- & \textbf{88.83} & 85.41 & 92.25 \\
TUE-Detector & -- & -- & 46.53 & 51.36 & 41.70 \\
VideoVeritas & -- & -- & 66.92 & 43.43 & 90.41 \\
\midrule
\rowcolor[HTML]{EDF3EF}
\multicolumn{6}{c}{\textbf{Agent}} \\
\midrule
ARGUS & -- & -- & 51.21 & 16.45 & 85.98 \\
\midrule
\rowcolor[HTML]{EDF3EF}
\multicolumn{6}{c}{\textbf{Ours}} \\
\midrule
\method, FV & 66.58 & 36.89 & 59.83 & 41.96 & 77.70 \\
\method, FV + Recal. & 67.06 & 36.82 & 56.55 & 17.45 & 95.64 \\
\method, ViF & 67.91 & \underline{35.24} & 51.13 & \underline{99.68} & 2.59 \\
\textcolor{methodcolor}{\textsc{\textbf{Antiphony-F}}}, FV & 66.58 & 36.89 & 59.50 & 40.92 & 78.09 \\
\textcolor{methodcolor}{\textsc{\textbf{Antiphony-F}}}, FV + Recal. & 67.06 & 36.82 & 57.38 & 19.43 & 95.34 \\
\textcolor{methodcolor}{\textsc{\textbf{Antiphony-F}}}, ViF & 67.91 & \underline{35.24} & 51.04 & \textbf{99.79} & 2.28 \\
\bottomrule
\end{tabular}%
}
\end{table*}
\clearpage

\begin{table}[t]
\centering
\caption{\textbf{Transfer with authentic target calibration (\%).} Expert fusion uses fixed ViF-CoT-4K-trained heads, one uniform 16-frame view, training seed 0, and 300 held-out authentic target videos per benchmark at $\alpha=5\%$. BAcc and specificity average five calibration splits. CoCoVideo scores average 13 generators; its AUROC uses split-averaged fusion scores. Re-encoding applies H.264 CRF 23 to calibration and evaluation videos. These evaluation sets differ from those in Table~\ref{tab:main}.}
\label{tab:supp_zs}
\resizebox{\linewidth}{!}{%
\begin{tabular}{lrrr}
\toprule
\textbf{Benchmark} & \textbf{BAcc} & \textbf{Spec.} & \textbf{AUROC} \\
\midrule
GenBuster-Bench (All Fakes) & 52.0 & 98.2 & 77.1 \\
\quad Commercial (Closed) Fakes & 53.8 & 98.2 & 82.1 \\
\midrule
GenBuster++ & 58.8 & 95.2 & 75.5 \\
\midrule
CoCoVideo-26K (Mean over 13) & 57.9 & 96.6 & 72.1 \\
\quad Re-Encoded & 55.6 & 96.8 & 67.1 \\
\bottomrule
\end{tabular}%
}
\end{table}

\paragraph{Generator Coverage and Recalibration.}
The FaceVid experts distinguish commercial image-to-video deepfakes more effectively than the open image-to-video configurations, with per-generator AUROCs of 74--96\% and 42--49\%, respectively. The latter generate motion from an authentic first frame. Recalibration raises specificity to about 95\%, but \method's BAcc decreases from 59.83\% to 56.55\% as recall falls. Restoring authentic specificity therefore does not necessarily improve detection of unfamiliar deepfakes. Skyra's deepfake recall also varies across generators, ranging from 7\% on Veo~3.1 to 98--100\% on several other configurations. These differences illustrate why aggregate results should be read alongside generator-level performance.

\subsection{Cross-Benchmark Transfer with Fixed Experts}
\label{sec:supp_zeroshot}

To assess transfer without fitting new expert heads, we keep the ViF-trained experts fixed and update the authentic reference and fusion threshold using target videos. All results in this subsection use expert fusion alone. The comparisons use original videos unless re-encoding is explicitly specified.

\paragraph{Target Data and Calibration.}
The GenBuster-Bench transfer set~\cite{busterxv1} contains 1{,}000 authentic videos and 2{,}150 deepfakes: 1{,}000 in-domain, 1{,}000 commercial, and 150 in-the-wild examples. The additional GenBuster++ video subset~\cite{busterx} contains 1{,}000 authentic videos and 1{,}000 deepfakes. Each set reserves 300 authentic videos for calibration and evaluates the remaining 700 authentic videos and all deepfakes. These transfer sets differ from the main-table Test and Closed splits. For CoCoVideo-26K, we reserve 300 authentic calibration videos from the full collection and evaluate the remainder by generator group, rather than using the official test split in Table~\ref{tab:main}.

\begin{table}[t]
\centering
\caption{\textbf{Detection after re-encoding (\%).} Expert fusion uses a uniform view and seed 0. Heads are trained on the balanced 20{,}000-video GenBuster subset or CoCoVideo's official training split. CoCoVideo scores average 13 generators. Re-encoded rows apply H.264 CRF 23 to both evaluation and calibration videos.}
\label{tab:commercial}
\resizebox{\linewidth}{!}{%
\begin{tabular}{lcccc}
\toprule
\rowcolor[HTML]{EDF3EF}
\multicolumn{5}{c}{\textbf{GenBuster-200K}} \\
\midrule
& \multicolumn{2}{c}{\shortstack{\textbf{Test}\\\textbf{(4 Open Models)}}} & \multicolumn{2}{c}{\shortstack{\textbf{Closed}\\\textbf{(8 Commercial)}}} \\
\cmidrule(lr){2-3}\cmidrule(lr){4-5}
\textbf{Method} & \textbf{Acc} & \textbf{F1} & \textbf{Acc} & \textbf{F1} \\
\midrule
\textcolor{methodcolor}{\textsc{\textbf{Antiphony-F}}} & 94.9 & 95.2 & 95.0 & 95.2 \\
\quad Re-Encoded & 95.2 & 95.5 & 95.1 & 95.2 \\
\midrule
\rowcolor[HTML]{EDF3EF}
\multicolumn{5}{c}{\textbf{CoCoVideo-26K (13 Commercial Generators)}} \\
\midrule
\textbf{Method} & \textbf{Acc} & \textbf{F1} & \textbf{Recall} & \textbf{AUROC} \\
\midrule
\textcolor{methodcolor}{\textsc{\textbf{Antiphony-F}}} & 97.4 & 97.5 & 99.0 & 99.7 \\
\quad Re-Encoded & 93.5 & 93.0 & 91.6 & 99.0 \\
\bottomrule
\end{tabular}%
}
\end{table}

\paragraph{Processing Protocols.}
Additional transfer checks standardize video processing across classes. For GenVideo-Val, we use the first 2\,s at 3\,fps, the native rate of its authentic videos, with a short side of 224 pixels and H.264 CRF 18, yielding 14{,}358 videos. For DVF, we use the first 2\,s at 7\,fps, the native rate of SVD, with a short side of 360 pixels and H.264 CRF 18, yielding 6{,}275 evaluation videos and 287 calibration videos. Both protocols exclude videos that are too short or too small. GenBuster transfer uses the original videos, whose resolution, frame rate, and duration match across classes. CoCoVideo is evaluated on original files and, separately, on H.264 CRF 23 copies at the original resolution and frame rate.

\paragraph{What Authentic Target Data Change.}
With ViF calibration, expert fusion flags most target videos. Updating the authentic reference reduces false alarms: specificity reaches 98.2\% on GenBuster's commercial subset and 96.6\% on CoCoVideo (See Table~\ref{tab:supp_zs}). The corresponding BAcc values are 53.8\% and 57.9\%, reflecting limited deepfake recall despite the recovered specificity. This pattern is consistent with CoCoDetect's cross-dataset results~\cite{cocovideo}. The same frozen encoders achieve substantially higher scores with benchmark-specific heads in Table~\ref{tab:main}. Those comparisons also use different evaluation sets, however, and do not isolate the effect of head training.

\begin{table*}[t]
\centering
\caption{\textbf{Robustness to video processing (\%).} Scores are ViF-Bench paired accuracy and FaceVid OOD BAcc. All \textcolor{methodcolor}{\textsc{\textbf{Antiphony-F}}} rows use the fusion-only variant with two-view expert inputs and training seed 0. Clean $\mathcal{C}$ and Processed $\mathcal{C}$ denote calibration on original authentic videos and copies processed in the same way as the evaluation videos, respectively. Skyra uses video decoding here and released frames in Table~\ref{tab:main}. ARGUS and the corresponding fusion-only subset row use the same 1{,}000 stratified FaceVid OOD videos, including 251 deepfakes.}
\label{tab:robust}
\begin{tabular}{lccccc}
\toprule
\textbf{Method} & \textbf{Clean} & \textbf{Blur} & \textbf{CRF 28} & \textbf{CRF 35} & \textbf{Downscale $\boldsymbol{\times}2$} \\
\midrule
\rowcolor[HTML]{EDF3EF}
\multicolumn{6}{c}{\textbf{ViF-Bench}} \\
\midrule
Skyra (Video Input) & 77.33 & 74.13 & 74.73 & 72.97 & 76.28 \\
DeMamba (ViF) & 73.63 & 64.03 & 74.55 & 73.38 & 73.02 \\
X-CLIP (ViF) & 78.73 & 67.34 & 78.43 & 72.49 & 75.98 \\
DRCT & 61.38 & 57.92 & 58.19 & 56.84 & 56.58 \\
\textcolor{methodcolor}{\textsc{\textbf{Antiphony-F}}}, Clean $\mathcal{C}$ & 90.67 & 77.51 & 82.31 & 64.19 & 70.75 \\
\textcolor{methodcolor}{\textsc{\textbf{Antiphony-F}}}, Processed $\mathcal{C}$ & 90.67 & 84.38 & 86.39 & 79.36 & 82.85 \\
\midrule
\rowcolor[HTML]{EDF3EF}
\multicolumn{6}{c}{\textbf{FaceVid OOD}} \\
\midrule
Effort (FaceVid) & 62.51 & 65.44 & 55.06 & 52.17 & 63.08 \\
\textcolor{methodcolor}{\textsc{\textbf{Antiphony-F}}}, Clean $\mathcal{C}$ & 71.72 & 71.35 & 55.11 & 51.20 & 68.52 \\
\textcolor{methodcolor}{\textsc{\textbf{Antiphony-F}}}, Processed $\mathcal{C}$ & 71.72 & 65.99 & 63.16 & 58.10 & 70.99 \\
ARGUS (Subset) & 70.30 & 68.35 & 68.38 & 66.24 & 67.26 \\
\textcolor{methodcolor}{\textsc{\textbf{Antiphony-F}}}, Processed $\mathcal{C}$ (Subset) & 71.65 & 67.56 & 63.81 & 57.86 & 71.44 \\
\bottomrule
\end{tabular}%
\end{table*}

\paragraph{Calibration and Representation Transfer.}
\label{sec:supp_scope}
Authentic target footage can adapt the operating point without new deepfake examples, but its benefit depends on the scores produced by the fixed experts. Recalibration can reduce false alarms; detecting unfamiliar forgeries still requires evidence that separates authentic from generated content. The DF26 results and processing study examine these effects under different shifts. Appendix~\ref{sec:supp_theory} states the conditions for false-alarm control and distinguishes the expert-fusion guarantee from empirical evaluation of the full cascade.

\subsection{Robustness to Video Processing}
\label{sec:supp_robust}

Video processing alters the evidence available to a detector without changing a video's authenticity. We examine re-encoding with benchmark-trained heads, followed by controlled post-processing with either clean or similarly processed authentic calibration videos. Expert heads remain fixed within each comparison, so these experiments test how processing affects the scores and whether recalibration can recover detection performance.

\paragraph{Re-Encoding on Commercial Generators.}
\label{sec:supp_commercial}
Table~\ref{tab:commercial} compares original videos with H.264 CRF 23 copies on GenBuster and CoCoVideo, applying the same processing to evaluation and calibration videos. The study uses benchmark-trained heads, a uniform input view, and training seed 0. GenBuster accuracy changes by at most 0.3 points. CoCoVideo accuracy falls by 3.9 points, although AUROC remains high at 99.0\%. The score ranking thus remains strong on CoCoVideo, while decisions at the calibrated threshold are more sensitive to re-encoding.

\paragraph{Controlled Processing Conditions.}
Table~\ref{tab:robust} evaluates two-view expert fusion with training seed 0 under four processing conditions: H.264 compression at CRF 28 or 35, downscaling by a factor of two with area interpolation, and Gaussian blur with $\sigma=1.5$ pixels. Each condition is applied to all frames while preserving frame count and frame rate. Encoding uses libx264 with the medium preset and YUV420p output; downscaling and blur use CRF 23. For each condition, we compare calibration on clean authentic videos with calibration on copies processed in the same way as the evaluation videos. Baselines receive the same processed videos and retain their released input pipelines.

\paragraph{Effect of Matching the Calibration Pipeline.}
Compression and downscaling generally shift scores toward ``authentic'', reducing both recall and false alarms, whereas blur raises FAR. Matching calibration to the processing condition recovers much of the ViF accuracy loss. Under CRF 35 compression, paired accuracy rises from 64.19\% to 79.36\%, and recalibrated fusion exceeds the evaluated baselines in every ViF processing condition. On FaceVid OOD, recalibration improves BAcc under compression and downscaling but lowers it under blur. ARGUS also retains higher BAcc under heavy compression on the matched subset. Representative calibration footage can compensate for processing-induced score shifts, but recovery depends on how much evidence for distinguishing the two classes survives.
\section{Ethics and Research Transparency}
\label{sec:supp_llm_usage}

\subsection{Ethical Considerations}
\label{sec:supp_ethics_considerations}

\paragraph{Intended Use and User Impact.}
\method is designed to support video-authenticity assessment by combining complementary visual evidence with selective investigation. Reliable assessment requires attention to both missed deepfakes and false alarms: incorrectly labeling authentic footage can undermine trust in legitimate content and the people associated with it. Our evaluation considers detection performance and false alarms together, while authentic-reference calibration makes the intended operating level explicit. Updating this reference with authentic target footage offers a way to account for changing capture conditions without collecting new deepfakes.

\paragraph{Data Use and Privacy.}
Our experiments use the benchmark videos and annotations described in Appendix~\ref{sec:supp_protocols}. We follow the dataset authors' permission for the separate DF26 recalibration described in Appendix~\ref{sec:supp_df26}. Some teacher, judge, and baseline models are accessed through external APIs that process sampled benchmark frames and accompanying task text. External processing and the reproduction of illustrative examples remain subject to the source datasets' usage terms and the privacy of the individuals depicted.

\paragraph{Explanations and Human Review.}
The investigator's explanations make its stated reasons available for examination alongside the observed frames and tool outputs. We assess this visual support separately from verdict accuracy, since a correct prediction can still be accompanied by unsupported claims. Appendix~\ref{sec:supp_judge} describes the model judges and evaluation protocol used for this assessment. Their judgments characterize explanation quality under that protocol, rather than verify every claim. In applications where a decision may affect an individual or the credibility of a source, model outputs should support human review of the original footage and other available evidence.

\paragraph{Evaluation Scope and Deployment.}
Our benchmarks cover multiple generators and video sources. Performance may nevertheless vary across content types, generation methods, capture devices, post-processing conditions and demographic groups. Evaluation in a new setting should examine these differences, using authentic calibration footage representative of its capture and processing conditions. Aggregate benchmark scores alone do not establish reliability for every population or application. The nominal false-alarm level should be interpreted together with its calibration assumptions and the observed error rates.

\paragraph{Dual-Use Considerations.}
Analysis of detector failures can guide improvements in forensic methods, but may also inform attempts to evade detection. The intended use of our findings is to improve video-authenticity assessment. Decisions about sharing models, evaluation resources and detailed prediction outputs should consider this potential for misuse alongside their value for reproducible research.

\subsection{LLM Usage Statement}
\label{sec:supp_llm_disclosure}

Claude and OpenAI Codex helped draft and edit text, refine figures and the presentation of research claims, and generate and debug code. ChatGPT Deep Research supported literature search and related-work synthesis. Codex also supported experimental planning and analysis, executed and monitored experiments, and provided reviewer-style feedback and consistency checks.

As part of the research method, Qwen3.8-Max generated investigator training trajectories and revised rationales, while Qwen3.8-Max and Gemini-3.8-Flash assessed explanation grounding. Appendices~\ref{sec:supp_agent} and~\ref{sec:supp_judge} describe these training and assessment procedures.

The authors directed the work, reviewed and edited tool outputs, and take full responsibility for all submitted content.

\end{document}